\documentclass[11pt,a4paper]{article}
\usepackage[margin=2.35cm,headheight=22pt,headsep=13pt,footskip=30pt]{geometry}
\usepackage[T1]{fontenc}
\usepackage[utf8]{inputenc}
\usepackage{amsmath,amssymb,amsthm}
\usepackage{mathtools}
\IfFileExists{newtxtext.sty}{\usepackage{newtxtext,newtxmath}}{}
\usepackage{microtype}
\usepackage{booktabs}
\usepackage{tabularx}
\usepackage{array}
\usepackage{enumitem}
\usepackage{xcolor}
\usepackage{listings}
\usepackage[most]{tcolorbox}
\usepackage{needspace}
\usepackage{etoolbox}
\usepackage{fancyhdr}
\usepackage{titlesec}
\usepackage{titling}
\usepackage{xurl}

\usepackage{tikz}
\usepackage{pgfplots}
\usepackage{placeins}
\pgfplotsset{compat=1.18}
\usetikzlibrary{arrows.meta,positioning,calc,decorations.pathmorphing,fit,backgrounds}

\usepackage[colorlinks=true]{hyperref}

\definecolor{ink}{HTML}{202A35}
\definecolor{navy}{HTML}{173B63}
\definecolor{accent}{HTML}{2E75B6}
\definecolor{teal}{HTML}{287C78}
\definecolor{softblue}{HTML}{F2F7FC}
\definecolor{softgreen}{HTML}{F2F8F5}
\definecolor{softgray}{HTML}{F6F7F8}
\definecolor{rulegray}{HTML}{D7DEE6}
\definecolor{codebg}{HTML}{F7F8FA}

\definecolor{amber}{HTML}{B4651E}
\definecolor{deepred}{HTML}{9B2C2C}
\colorlet{srcA}{accent}
\colorlet{srcB}{amber}
\colorlet{pooled}{teal}
\colorlet{struct}{ink}
\colorlet{warn}{deepred}
\tikzset{
  figbase/.style={font=\footnotesize},
  ax/.style={rulegray!150,-{Stealth[length=1.7mm,width=1.3mm]},line width=0.45pt},
  vec/.style={-{Stealth[length=2.3mm,width=1.8mm]},line width=1.0pt},
  flow/.style={struct!70,-{Stealth[length=2.6mm,width=2.0mm]},line width=0.9pt},
  crv/.style={line width=1.1pt},
  thincrv/.style={line width=0.7pt},
  lbl/.style={font=\footnotesize,fill=white,inner sep=1.2pt},
  tick/.style={font=\scriptsize,struct!60},
  fcap/.style={font=\footnotesize,align=center,struct!85},
  bx/.style={draw,line width=0.9pt,rounded corners=2pt,align=center,
             minimum height=9mm,inner sep=3pt,font=\footnotesize},
}

\hypersetup{
  linkcolor=navy,
  citecolor=teal,
  urlcolor=accent,
  pdftitle={Source-Resolved Collective Cognition in Robotics: Mathematical Foundations for Multimodal and Collective Estimation, and Their Reach across the Cognition Loop},
  pdfauthor={Jürgen Graf},
  pdfsubject={Foundations of source-resolved collective estimation, generalized singular value decomposition, numerical linear algebra, sensor fusion, collective systems},
  pdfkeywords={GSVD, SVD, numerical linear algebra, Fisher information, sensor fusion, collective systems, canonical correlation analysis}
}
\newtheoremstyle{resultstyle}
  {8pt}{6pt}{\itshape}{}
  {\bfseries\color{navy}}{.}{0.55em}{}
\newtheoremstyle{definitionstyle}
  {8pt}{6pt}{\normalfont}{}
  {\bfseries\color{teal}}{.}{0.55em}{}
\newtheoremstyle{remarkstyle}
  {7pt}{5pt}{\normalfont}{}
  {\bfseries\color{ink!82}}{.}{0.55em}{}

\theoremstyle{resultstyle}
\newtheorem{theorem}{Theorem}[section]
\newtheorem{lemma}[theorem]{Lemma}
\newtheorem{corollary}[theorem]{Corollary}
\newtheorem{proposition}[theorem]{Proposition}
\theoremstyle{definitionstyle}
\newtheorem{definition}[theorem]{Definition}
\newtheorem{example}[theorem]{Example}
\theoremstyle{remarkstyle}
\newtheorem{remark}[theorem]{Remark}

\AtBeginEnvironment{theorem}{\Needspace{7\baselineskip}}
\AtBeginEnvironment{lemma}{\Needspace{6\baselineskip}}
\AtBeginEnvironment{corollary}{\Needspace{6\baselineskip}}
\AtBeginEnvironment{proposition}{\Needspace{7\baselineskip}}
\AtBeginEnvironment{definition}{\Needspace{5\baselineskip}}
\AtBeginEnvironment{example}{\Needspace{6\baselineskip}}
\AtBeginEnvironment{remark}{\Needspace{5\baselineskip}}

\newtcolorbox{practicebox}{
  enhanced,breakable,
  colback=softgreen,colframe=softgreen,
  boxrule=0pt,arc=1.2mm,
  borderline west={2.4pt}{0pt}{teal},
  left=0.95em,right=0.8em,top=0.55em,bottom=0.55em,
  before skip=5pt,after skip=8pt,
  fonttitle=\bfseries\sffamily\color{teal},
  title={Practical meaning}
}

\newtcolorbox{collective}[1][]{
  enhanced,breakable,
  colback=softblue,colframe=softblue,
  boxrule=0pt,arc=1.2mm,
  borderline west={2.4pt}{0pt}{accent},
  left=0.95em,right=0.8em,top=0.55em,bottom=0.55em,
  before skip=7pt,after skip=8pt,
  fonttitle=\bfseries\sffamily\color{accent},
  fontupper=\itshape,
  title={Collective reading#1}
}

\newtcolorbox{researchhypothesis}{
  enhanced,breakable,
  colback=softgray,colframe=softgray,
  boxrule=0pt,arc=1.2mm,
  borderline west={2.4pt}{0pt}{navy},
  left=0.95em,right=0.8em,top=0.55em,bottom=0.55em,
  before skip=7pt,after skip=8pt,
  fonttitle=\bfseries\sffamily\color{navy},
  title={Research hypothesis --- not a theorem}
}

\titleformat{\section}
  {\Large\bfseries\color{navy}}{\thesection}{0.65em}{}
  [\vspace{0.15em}\color{accent!65}\titlerule]
\titleformat{name=\section,numberless}
  {\Large\bfseries\color{navy}}{}{0pt}{}
  [\vspace{0.15em}\color{accent!65}\titlerule]
\titleformat{\subsection}
  {\large\bfseries\color{ink}}{\thesubsection}{0.6em}{}
\titleformat{\subsubsection}
  {\normalsize\bfseries\color{ink}}{\thesubsubsection}{0.55em}{}
\titleformat{\part}[display]
  {\centering\bfseries\color{navy}}
  {}{0pt}{\Huge}
\titlespacing*{\section}{0pt}{2.1ex plus .6ex minus .2ex}{1.2ex}
\titlespacing*{\subsection}{0pt}{1.8ex plus .5ex minus .2ex}{0.8ex}

\setlist[itemize]{topsep=3pt,itemsep=2pt,parsep=0pt}
\setlist[enumerate]{topsep=3pt,itemsep=2pt,parsep=0pt}
\allowdisplaybreaks[2]
\renewcommand{\headrulewidth}{0.4pt}
\renewcommand{\headrule}{\hbox to\headwidth{\color{rulegray}\leaders\hrule height \headrulewidth\hfill}}

\pretitle{\begin{flushleft}\vspace*{-1.2cm}\color{navy}\Huge\bfseries}
\posttitle{\par\vspace{0.45em}\color{accent}\rule{\textwidth}{1.15pt}\end{flushleft}}
\preauthor{\begin{flushleft}\small\color{ink!75}}
\postauthor{\par\end{flushleft}}
\predate{\begin{flushleft}\small\color{ink!65}}
\postdate{\par\vspace{0.4em}\end{flushleft}}

\renewenvironment{abstract}
  {\begin{tcolorbox}[enhanced,breakable,colback=softgray,colframe=rulegray,
     boxrule=0.5pt,arc=1.2mm,left=1em,right=1em,top=0.7em,bottom=0.7em,
     before skip=4pt,after skip=10pt,title={Abstract},
     fonttitle=\bfseries\sffamily\color{navy}]\small}
  {\end{tcolorbox}}

\usepackage{bm}
\newcommand{\mat}[1]{\bm{#1}}
\newcommand{\vekt}[1]{\bm{#1}}
\newcommand{\ten}[1]{\bm{#1}}
\newcommand{\FisherMat}{\bm{\mathcal{I}}}

\newcommand{\Tf}[3]{\mat{T}_{#1,#2}^{#3}}
\newcommand{\Cf}[2]{\mat{C}_{#1}^{#2}}

\newcommand{\R}{\mathbb{R}}
\newcommand{\rank}{\operatorname{rank}}
\newcommand{\trace}{\operatorname{trace}}
\newcommand{\diag}{\operatorname{diag}}
\newcommand{\range}{\operatorname{range}}
\newcommand{\nullsp}{\operatorname{null}}
\newcommand{\spn}{\operatorname{span}}
\newcommand{\Var}{\operatorname{Var}}
\newcommand{\Exp}{\operatorname{Exp}}

\newcommand{\Ad}{\operatorname{Ad}}
\newcommand{\T}{^{\mathsf{T}}}
\newcommand{\half}{\tfrac{1}{2}}

\newcommand{\FoundationsAuthor}{Prof.\ Dr.\ J\"urgen Graf}

\title{Source-Resolved Collective Cognition in Robotics\\[3pt]
\Large\mdseries Mathematical Foundations for Multimodal and Collective Estimation,\\
and Their Reach across the Cognition Loop}
\author{%
\FoundationsAuthor\\[2pt]
\normalsize Cognitive Systems and Robotics Laboratory (CSRL)\\
\normalsize Department of Computer Science\\
\normalsize Trier University of Applied Sciences, Trier, Germany}
\date{2 November 2026}

\begin{document}
\maketitle
\thispagestyle{empty}
\clearpage

\begin{abstract}
\noindent
The Generalized Singular Value Decomposition (GSVD) is a central tool for
comparing two linear operators that act on the same unknown. Once the output
metrics are fixed, it answers the basis-independent question ``along which
state directions does the first operator dominate the second, and by how
much?''.

This foundations volume builds the GSVD and its surrounding estimation theory from the ground up. Part~I re-derives every piece of linear algebra and statistical estimation machinery needed later --- inner products, the four subspaces, orthogonal matrices, the spectral theorem, quadratic forms, the ordinary SVD, Lie-group tangent perturbations, Jacobians, covariance, whitening, Fisher information, CCA, mutual information, discrimination criteria, and covariance intersection --- with self-contained derivations and worked numerical examples. Part~II states the GSVD in both a transparent and a dimensionally general form,
then proves it by two complementary routes. Part~III works one small matrix
pair through five different lenses, each computed by hand and verified
numerically. Part~IV separates information attribution, subspace geometry, statistical
dependence, and decision-theoretic discrimination for collective systems. Part~V evaluates
the role and limits of the GSVD across the cognition loop. Part~VI covers numerical
practice, including a tested reference implementation.

Every number printed in this document was verified in double precision. Where a
statement is an assumption rather than a theorem, it is labelled as such.
\end{abstract}

\clearpage
\section*{Notation and conventions}
\addcontentsline{toc}{section}{Notation and conventions}

This foundations volume follows the notational conventions of the Cognitive Systems and
Robotics Laboratory throughout.

\begin{center}
\begin{tabular}{@{}lll@{}}
\toprule
\textbf{Object} & \textbf{Typeface} & \textbf{Examples}\\
\midrule
Scalar & lowercase, not bold & $m$, $n$, $r$, $\alpha_i$, $\sigma_i$, $\lambda$, $\theta_i$\\
Vector & \textbf{bold} lowercase & $\vekt{x}$, $\vekt{u}_i$, $\vekt{b}$, $\vekt{\gamma}$\\
Matrix & \textbf{bold} uppercase & $\mat{A}$, $\mat{B}$, $\mat{X}$, $\mat{\Sigma}$\\
Tensor & \textbf{bold} uppercase & $\ten{T}$\\
Set, subspace & calligraphic / standard set symbol & $\mathcal{S}$, $\mathcal{T}$, $\R^n$\\
Agent, functional & calligraphic & $\mathcal{A}$, $\mathcal{B}$, $\mathcal{R}(\vekt{x})$\\
\bottomrule
\end{tabular}
\end{center}

Two consequences are worth stating explicitly, because they resolve collisions
that are otherwise easy to misread.

\textbf{Bold distinguishes a vector from a component.} $\vekt{x}_i$ is the
$i$-th generalized singular vector; $x_i$ is the $i$-th scalar component of $\vekt{x}$.
The same rule applies to Greek symbols: $\vekt{\eta}$ is a vector, whereas
$\eta_i$ is a scalar component. The zero vector is written $\vekt{0}$; the
scalar zero remains $0$. These distinctions are enforced throughout.

\textbf{Labels are not mathematical objects.} In $\mat{D}_A$, $\mat{H}_{\mathcal A}$,
$\mat{\Sigma}_{AB}$, or $\theta_i^{(A)}$, the letters $A$ and $B$ identify a
block, operator, or agent. They are labels, not scalar variables, and are therefore
not bold. The same rule applies to frame labels in coordinate transformations.

\paragraph{Coordinate frames.}
This foundations text uses the laboratory convention
\[
   \Tf{b}{c}{a}.
\]
Here $b$ is the \emph{reference frame}, $c$ the \emph{object frame}, and $a$ the
\emph{resolving frame}. Thus $\Tf{b}{c}{a}$ is read as ``the pose of $c$ relative
to $b$, resolved in $a$.'' More generally, $\vekt{q}_{b,c}^{a}$ means ``the quantity
of $c$ relative to $b$, resolved in $a$.'' For rotations, $\Cf{b}{a}$ maps vector
coordinates from frame $b$ to frame $a$, i.e.
$\vekt{x}^{a}=\Cf{b}{a}\vekt{x}^{b}$. The GSVD itself is frame-free, but every
frame-dependent residual or Jacobian must be expressed consistently before fusion.

\clearpage
\section*{Foreword}
\addcontentsline{toc}{section}{Foreword}

Two sensors look at the same thing. One is a LiDAR, the other a camera; or one
is a robot and the other is the vehicle beside it. Both are estimating the same
quantity, and both are wrong in their own way. Which of them is telling you
more, and about what?

Engineers answer this question constantly, and almost always empirically: turn
one sensor off, see what gets worse. That works, but it is expensive, it has to
be repeated for every configuration, and it produces a number without a
structure --- you learn \emph{that} accuracy dropped, not \emph{which
directions} of the state were carried by which sensor.

There is an exact linear-algebraic answer to the \emph{relative-response}
question. Once the residual metrics are fixed, the Generalized Singular Value
Decomposition finds shared state directions in which the two operators separate
cleanly and attaches to each direction a pair of nonnegative numbers whose
squares add to one. Under the additional assumptions used later --- a valid
linear model, known covariances, and independent Gaussian measurement errors ---
whitening turns those squared numbers into exact \emph{relative Fisher-information
shares}. In that setting the same decomposition also reveals the covariance
geometry of the fused estimator. The distinction matters: the decomposition is
always linear algebra; the information interpretation is a theorem only under
its stated statistical assumptions.

That is the reason for this foundations volume. The GSVD is not new --- it dates from 1976
and sits in every serious numerical linear algebra text --- but it is largely
absent from the perception, navigation and multi-robot literature, and where it
does appear it is easy to misread. Its angles are routinely confused with
principal angles between the sensors' observable subspaces, which they are not.
Its ratios are read as statements about redundancy, which they are not either.
Its information interpretation is exact, but only under assumptions that are
worth stating out loud, because one of them --- independence of the two error
sources --- fails routinely in exactly the collective settings where the
decomposition is most attractive.

So this text does three things. It builds the mathematics from the ground up,
assuming only that the reader once knew linear algebra well and would like it
back: inner products, the four subspaces, the spectral theorem, the ordinary
SVD, and then the GSVD itself, proved twice by independent routes. It works one
small pair of matrices through five different lenses by hand, with every
intermediate number shown and independently verified in double precision, on
the view that one example understood completely teaches more than ten examples
skimmed. And it is deliberately careful about the boundary between what the
decomposition establishes and what it merely suggests --- which angles are
which, when a ratio does and does not certify a single point of failure, what
the whitening step is actually doing, and where the algorithm stops being
numerically trustworthy.

The application chapters are written for collective systems: several agents,
heterogeneous sensors, a shared world, and the question of who knows what.
That framing is not decoration. A group of agents has a natural coordinate
system in which each member's contribution is separately visible, that
coordinate system is generally not the Euclidean one, and the GSVD is how you
find it. Section~\ref{sec:relatedwork} sets this against the current
literature and marks where the open problems lie.

The text can be read straight through, or used as a reference: Part~I is a
refresher, Part~II the theory, Part~III the arithmetic, Part~IV the
interpretation, and Part~V the numerical practice that keeps the rest honest.

\clearpage
\tableofcontents
\newpage

\clearpage
\part*{Part 0 --- Why This Decomposition Exists}
\section{The question the GSVD answers}

Start with a picture rather than a formula.

Two agents observe the same world. Agent~$\mathcal A$ carries a LiDAR; agent
$\mathcal B$ carries a stereo camera. Both are estimating the \emph{same} state
$\vekt{x}\in\R^n$ --- for example, the relative pose and velocity of another object.
Their raw linearised measurement models are
\[
   \vekt{z}_{\mathcal A}=\mat{H}_{\mathcal A}\vekt{x}+\vekt{\eta}_{\mathcal A},\qquad
   \vekt{z}_{\mathcal B}=\mat{H}_{\mathcal B}\vekt{x}+\vekt{\eta}_{\mathcal B}.
\]
The two matrices have the same number of columns, because they act on the same
unknown, but they need not have the same number of rows.

There is one important point to settle before any decomposition. A metre, a
pixel and a radian are not automatically comparable. Therefore the matrices
that enter the GSVD should be understood as \emph{metric-aware} operators
\begin{equation}\label{eq:scaledops}
   \mat{A}=\mat{W}_{\mathcal A}\mat{H}_{\mathcal A},\qquad
   \mat{B}=\mat{W}_{\mathcal B}\mat{H}_{\mathcal B},
\end{equation}
where $\mat{W}_{\mathcal A}$ and $\mat{W}_{\mathcal B}$ specify how measurement residuals
are measured. In a linear-Gaussian sensor model the natural choice is whitening,
$\mat{W}_{\mathcal A}=\mat{\Sigma}_{\mathcal A}^{-1/2}$ and
$\mat{W}_{\mathcal B}=\mat{\Sigma}_{\mathcal B}^{-1/2}$. Algebraically the GSVD exists
without this statistical interpretation; physically meaningful comparisons
between heterogeneous sensors require a stated metric.\label{sec:metric}

There is a second, separate issue, and it is numerical. The educational
stacking algorithm used later can lose the smaller block when the two
\emph{metric-aware} operators differ by many orders of magnitude in norm.
Whitening is still the right statistical metric when the noise covariances are
known, but it does \textbf{not} guarantee comparable Jacobian norms: it
standardises the residual noise, not the sensitivity of the measurement model.
A large post-whitening norm ratio may therefore be physically real. In that
case one should preserve the model and use a production GSVD or a carefully
justified equilibration, rather than separately rescaling the two agents merely
to make the numbers convenient. Section~\ref{sec:scale-precondition} measures
the scale range of the transparent stacking implementation.

Notice the asymmetry that drives everything that follows:
\begin{center}
\begin{tabular}{@{}ll@{}}
\toprule
\textbf{Shared} & the domain $\R^n$ --- the state, the unknown\\
\textbf{Not shared} & the codomains $\R^m$ and $\R^p$ --- the measurement spaces\\
\bottomrule
\end{tabular}
\end{center}

\begin{figure}[htbp]
\centering
\begin{tikzpicture}[figbase]
  \draw[struct!45,crv,fill=softgray] (0,0) ellipse (1.8 and 2.25);
  \node[struct!70] at (0,1.55) {shared};
  \node at (0,0.45) {$\R^n$};
  \node[fcap] at (0,-0.45) {state $\vekt x$};
  \node[struct!60] at (0,-2.75) {the world, the unknown};

  \draw[srcA,crv,fill=softblue] (6.6,2.5) ellipse (1.55 and 1.2);
  \node[srcA] at (6.6,2.78) {$\R^m$};
  \node[srcA] at (6.6,2.22) {agent $\mathcal A$};

  \draw[srcB,crv,fill=srcB!7] (6.6,-2.5) ellipse (1.55 and 1.2);
  \node[srcB] at (6.6,-2.22) {$\R^p$};
  \node[srcB] at (6.6,-2.78) {agent $\mathcal B$};

  \draw[vec,srcA] (1.9,0.85) to[out=25,in=192] (4.9,2.4);
  \draw[vec,srcB] (1.9,-0.85) to[out=-25,in=168] (4.9,-2.4);
  \node[text=srcA] at (3.20,2.75) {$\mat A\in\R^{m\times n}$};
  \node[text=srcB] at (3.20,-2.75) {$\mat B\in\R^{p\times n}$};

  \draw[struct!35,dashed,thincrv] (6.6,1.15) -- (6.6,-1.15);
  \node[lbl,text=struct!60,align=center] at (6.6,0)
    {different codomains,\\ no entrywise comparison};

  \node[fcap] at (3.2,-4.15)
    {same number of \emph{columns}, different number of \emph{rows}\\[2pt]
     after fixing the output metrics, a canonical directional comparison is\\[1pt]
     $\rho(\vekt x)=\|\mat A\vekt x\|\,/\,\|\mat B\vekt x\|$};
\end{tikzpicture}
\caption{\textbf{The two operators share their domain but not their codomain.} $\mat A$ and $\mat B$ act on the \emph{same} unknown state $\vekt x\in\R^n$, but their responses live in different measurement spaces and in general different dimensions. Before metric normalisation, the underlying sensor residuals may also carry different physical units. Nothing may therefore be compared entrywise across the dashed line; the output metrics must first be fixed. A canonical directional comparison that survives the crossing is the ratio of response lengths $\rho(\vekt x)=\|\mat A\vekt x\|/\|\mat B\vekt x\|$, and diagonalising that ratio over all directions at once is precisely what the GSVD does.}
\label{fig:domain-codomain-asymmetry}
\end{figure}
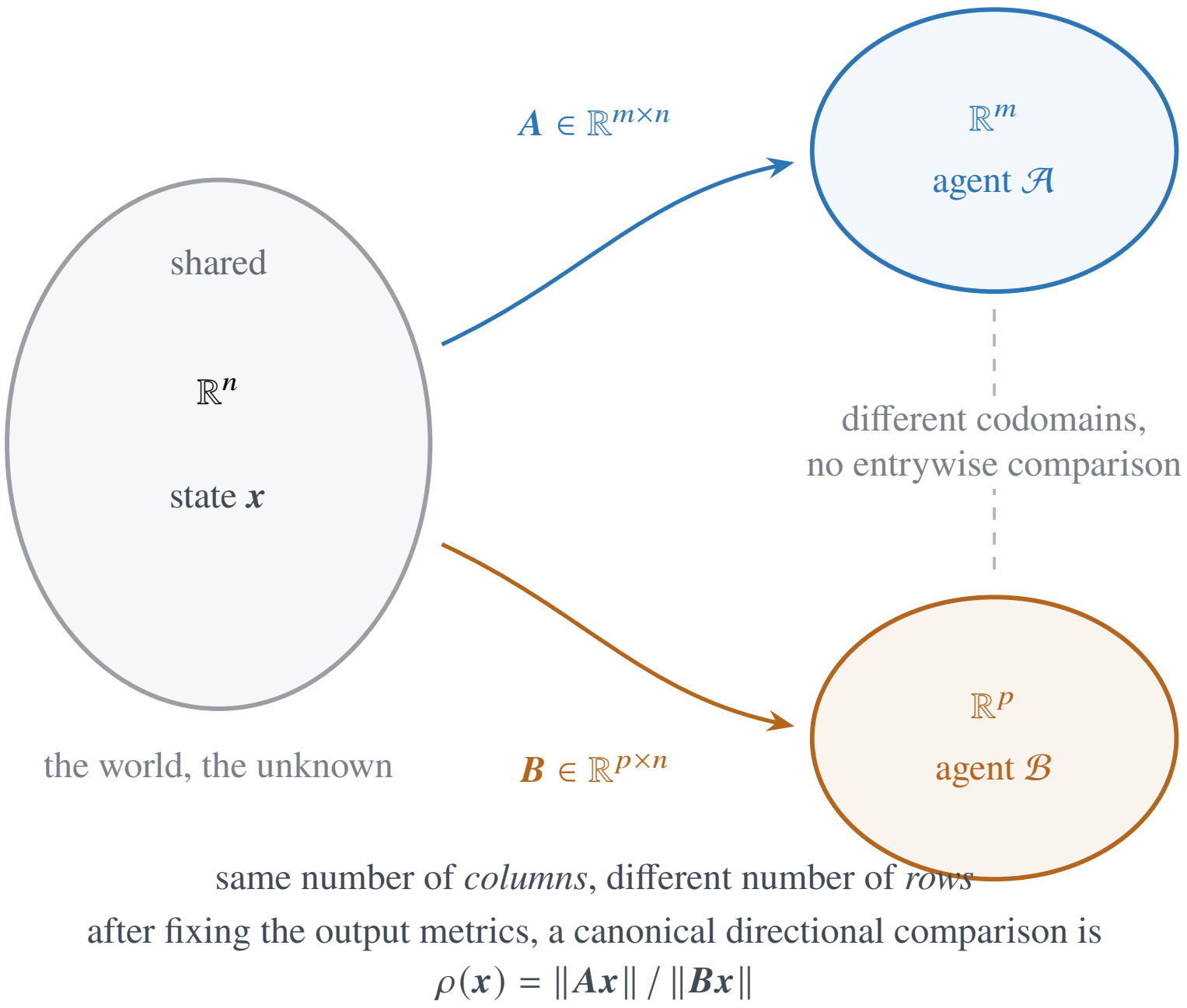

Figure~\ref{fig:domain-codomain-asymmetry} draws the same asymmetry, with the
dashed line marking where entrywise comparison stops being meaningful.

\FloatBarrier

The natural question is not whether ``$\mat{A}>\mat{B}$''. It is:
\begin{quote}
\emph{Along which state directions does $\mat{A}$ respond more strongly than $\mat{B}$,
and along which directions does $\mat{B}$ respond more strongly than $\mat{A}$?}
\end{quote}
For any $\vekt{x}$ with $\mat{B}\vekt{x}\neq\vekt{0}$, the relevant ratio is
\begin{equation}\label{eq:ratio}
   \rho(\vekt{x})=\frac{\|\mat{A}\vekt{x}\|}{\|\mat{B}\vekt{x}\|}.
\end{equation}
If $\mat{B}\vekt{x}=\vekt{0}$ but $\mat{A}\vekt{x}\neq\vekt{0}$, the ratio is $\infty$; if both vanish, the direction is
invisible to both and the ratio is undefined.

The GSVD finds special state directions $\vekt{x}_i$ on which this comparison decouples:
\[
   \mat{A}\vekt{x}_i=\alpha_i\vekt{u}_i,\qquad \mat{B}\vekt{x}_i=\beta_i\vekt{v}_i,
   \qquad \alpha_i^2+\beta_i^2=1,
\]
with orthonormal output directions $\vekt{u}_i$ and $\vekt{v}_i$. The generalized singular
value is
\[
   \sigma_i=\frac{\alpha_i}{\beta_i}.
\]
Thus $\sigma_i$ is a \emph{relative} quantity. Large finite $\sigma_i$ means
that $\mat{A}$ dominates $\mat{B}$ along $\vekt{x}_i$; only $\sigma_i=\infty$ means that $\mat{B}$ sees
nothing there. Likewise, only $\sigma_i=0$ means that $\mat{A}$ sees nothing.

\begin{collective}
The GSVD separates two questions that should never be confused.

\textbf{Relative attribution:} who contributes more along a direction? This is
what $\alpha_i^2$, $\beta_i^2$ and $\sigma_i$ answer.

\textbf{Absolute collective strength:} how strongly is that direction
constrained at all? This is governed by the pooled matrix
$\mat{J}=\mat{A}\T \mat{A}+\mat{B}\T \mat{B}$.

A balanced direction, $\sigma_i\approx1$, can be very well observed or almost
unobserved. A wide spectrum means heterogeneous relative strengths; it does
\emph{not} by itself mean that losing one agent makes a subspace unobservable.
That stronger conclusion requires $\sigma_i=0$ or $\infty$.
\end{collective}

\subsection{One warning that will save you a lot of confusion}

The generalized input directions are generally \textbf{not Euclidean-orthogonal}.
That is a genuine difference from the ordinary SVD. But ``oblique'' is only
half the story.

In the full-column-rank case, define
\[
   \mat{J}=\mat{A}\T \mat{A}+\mat{B}\T \mat{B}.
\]
The GSVD can be normalised so that
\begin{equation}\label{eq:Jorth-intro}
   \mat{X}\T \mat{J}\mat{X}=\mat{I}.
\end{equation}
Therefore the columns of $\mat{X}$ are orthonormal in the \emph{$\mat{J}$-metric}
\[
   \langle \vekt{u},\vekt{v}\rangle_{\mat{J}}=\vekt{u}\T \mat{J}\vekt{v}.
\]
They look oblique only when judged with the ordinary Euclidean inner product.
The pooled problem itself supplies the geometry in which the generalized
coordinates are orthonormal.

\begin{collective}
This is the useful geometric reading for a sensor collective. Heterogeneous
sensors need not admit one Euclidean frame that diagonalises every contribution,
but the pooled information $\mat{J}$ defines a natural metric. In that metric the
GSVD directions are orthonormal and each agent's contribution is diagonal. This
is a much stronger statement than saying merely that ``the basis is oblique''.
\end{collective}

\subsection{How to read this foundations text}

Part~I rebuilds the linear algebra and statistical foundations needed later, including Lie-group tangent perturbations, Jacobians, covariance, whitening, Fisher information, CCA, mutual information, discrimination criteria, and conservative fusion. Part~II states the GSVD first
in a transparent form and then in a dimensionally correct general form. Part~III
works one small matrix pair by hand. Part~IV separates four ideas that are often
mixed together: information attribution (GSVD), subspace geometry (principal
angles), statistical dependence (CCA/cross-covariance), and hypothesis
discrimination (information-theoretic distances). Part~V covers numerical
practice and a reference implementation.

\clearpage
\part*{Part I --- Prerequisites, Rebuilt}

\section{Vectors, inner products, length, angle}

A vector $\vekt{x}\in\R^n$ is a list of $n$ numbers. Two operations generate everything:

\paragraph{Inner product.} For $\vekt{x},\vekt{y}\in\R^n$,
\[
   \vekt{x}\T \vekt{y} \;=\; \sum_{i=1}^n x_i y_i .
\]
It is symmetric ($\vekt{x}\T \vekt{y}=\vekt{y}\T \vekt{x}$), linear in each argument, and positive:
$\vekt{x}\T \vekt{x}=\sum_i x_i^2 \ge 0$, with equality only if $\vekt{x}=\vekt{0}$.

\paragraph{Norm.} $\|\vekt{x}\| = \sqrt{\vekt{x}\T \vekt{x}}$. This is ordinary Euclidean length.

\paragraph{Angle.} For nonzero $\vekt{x},\vekt{y}$,
\[
   \cos\theta \;=\; \frac{\vekt{x}\T \vekt{y}}{\|\vekt{x}\|\,\|\vekt{y}\|}\,,\qquad \theta\in[0,\pi].
\]
That this is a legitimate cosine (i.e.\ that the quotient lies in $[-1,1]$) is
the Cauchy--Schwarz inequality $|\vekt{x}\T \vekt{y}|\le\|\vekt{x}\|\|\vekt{y}\|$.

\paragraph{Orthogonality.} $\vekt{x}\perp \vekt{y}$ means $\vekt{x}\T \vekt{y}=0$, i.e.\ $\theta=\pi/2$.

\begin{example}[Explicit numbers]
Let $\vekt{x}=(3,4)\T$, $\vekt{y}=(-4,3)\T$.
\begin{align*}
\vekt{x}\T \vekt{y} &= 3\cdot(-4) + 4\cdot 3 = -12+12 = 0 &&\text{so } \vekt{x}\perp \vekt{y},\\
\|\vekt{x}\| &= \sqrt{9+16}=\sqrt{25}=5, \qquad \|\vekt{y}\| = \sqrt{16+9}=5.
\end{align*}
Now take $\vekt{z}=(1,1)\T$. Then $\vekt{x}\T \vekt{z} = 3+4=7$, $\|\vekt{z}\|=\sqrt2$, so
$\cos\theta = 7/(5\sqrt2)= 7/7.0711 = 0.98995$, giving $\theta = 8.13^\circ$.
The vector $\vekt{z}$ points almost the same way as $\vekt{x}$.
\end{example}
\begin{practicebox}
This small calculation fixes the geometry used everywhere later: inner products measure alignment, norms measure response size, and orthogonality removes cross terms. The GSVD never leaves these basic operations; it only applies them to two operators at once.
\end{practicebox}

\paragraph{Pythagoras.} If $\vekt{x}\perp \vekt{y}$ then
$\|\vekt{x}+\vekt{y}\|^2 = (\vekt{x}+\vekt{y})\T(\vekt{x}+\vekt{y}) = \vekt{x}\T \vekt{x} + 2\vekt{x}\T \vekt{y} + \vekt{y}\T \vekt{y} = \|\vekt{x}\|^2+\|\vekt{y}\|^2$.
We will use this constantly, because $\alpha_i^2+\beta_i^2=1$ in the GSVD is
exactly a Pythagoras statement.

\section{Matrices as maps; the four fundamental subspaces}

A matrix $\mat{A}\in\R^{m\times n}$ is a linear map $\R^n\to\R^m$. Its columns
$\vekt{a}_1,\dots,\vekt{a}_n\in\R^m$ satisfy $\mat{A}\vekt{x} = \sum_j x_j \vekt{a}_j$: \emph{$\mat{A}\vekt{x}$ is a
combination of the columns of $\mat{A}$, with weights taken from $\vekt{x}$.} Hold on to that
sentence; it explains every subspace below.

\begin{definition}[The four subspaces]
For $\mat{A}\in\R^{m\times n}$:
\begin{align*}
\range(\mat{A})   &= \{\mat{A}\vekt{x} : \vekt{x}\in\R^n\}\subseteq\R^m &&\text{\emph{column space}, dimension } r,\\
\nullsp(\mat{A})  &= \{\vekt{x} : \mat{A}\vekt{x}=\vekt{0}\}\subseteq\R^n      &&\text{\emph{null space}, dimension } n-r,\\
\range(\mat{A}\T) &= \{\mat{A}\T \vekt{y} : \vekt{y}\in\R^m\}\subseteq\R^n &&\text{\emph{row space}, dimension } r,\\
\nullsp(\mat{A}\T)&= \{\vekt{y} : \mat{A}\T \vekt{y}=\vekt{0}\}\subseteq\R^m   &&\text{\emph{left null space}, dimension } m-r,
\end{align*}
where $r=\rank(\mat{A})$.
\end{definition}
\begin{practicebox}
For sensing, $\nullsp(\mat{A})$ is exactly the set of state changes that produce no response, while $\range(\mat{A}\T)$ is the state subspace that the measurements can constrain. Any observability or sensor-suite audit begins with this split.
\end{practicebox}

Two orthogonality facts, both one-liners, both essential:

\begin{lemma}\label{lem:orth}
$\nullsp(\mat{A})\perp\range(\mat{A}\T)$ and $\nullsp(\mat{A}\T)\perp\range(\mat{A})$, and in each case
the two subspaces together span the whole space.
\end{lemma}
\begin{practicebox}
This lemma gives a clean decomposition into visible and invisible directions. In practice it lets us reason about missing information geometrically instead of reading significance from individual matrix entries.
\end{practicebox}

\begin{proof}
If $\mat{A}\vekt{x}=\vekt{0}$ then for any $\vekt{y}$, $\vekt{x}\T(\mat{A}\T \vekt{y}) = (\mat{A}\vekt{x})\T \vekt{y} = \vekt{0}\T \vekt{y} = 0$. So every
null-space vector is orthogonal to every row-space vector. Dimensions add up:
$\dim\nullsp(\mat{A})+\dim\range(\mat{A}\T) = (n-r)+r = n$. Two orthogonal subspaces whose
dimensions sum to $n$ span $\R^n$. Apply the same argument to $\mat{A}\T$ for the
second pair.
\end{proof}

\begin{collective}
For an observation matrix $\mat{A}$, $\nullsp(\mat{A})$ is precisely the \textbf{unobservable
subspace} of that agent: state changes along $\nullsp(\mat{A})$ produce no measurement
whatsoever. $\range(\mat{A}\T)$ is the observable subspace. The GSVD will compare
$\nullsp(\mat{A})$ and $\nullsp(\mat{B})$ for two agents, and the interesting object is
$\nullsp(\mat{A})\cap\nullsp(\mat{B})$ --- the \textbf{collectively unobservable subspace},
which no amount of fusion can recover.
\end{collective}

\begin{example}[Subspaces by hand]
Let
\[
\mat{A}=\begin{pmatrix}1&2&3\\2&4&6\end{pmatrix}.
\]
Row 2 is twice row 1, so $\rank(\mat{A})=1$.
\begin{itemize}[leftmargin=1.4em]
\item $\range(\mat{A})=\spn\{(1,2)\T\}$: every column is a multiple of $(1,2)\T$.
\item $\range(\mat{A}\T)=\spn\{(1,2,3)\T\}$.
\item $\nullsp(\mat{A})$: solve $x_1+2x_2+3x_3=0$, a plane. A basis:
      $(-2,1,0)\T$ and $(-3,0,1)\T$. Check: $(1,2,3)\cdot(-2,1,0)=-2+2+0=0$
      \checkmark; $(1,2,3)\cdot(-3,0,1)=-3+0+3=0$ \checkmark. Dimension
      $3-1=2$ \checkmark.
\item $\nullsp(\mat{A}\T)$: solve $y_1+2y_2=0$ and $2y_1+4y_2=0$ --- the same
      equation twice --- giving $\spn\{(-2,1)\T\}$, dimension $2-1=1$ \checkmark.
\end{itemize}
Every null-space basis vector is orthogonal to the row space, as
Lemma~\ref{lem:orth} promised.
\end{example}
\begin{practicebox}
The example shows that rank loss is a geometric fact, not an algorithmic failure. A sensor Jacobian can contain many nonzero entries and still leave an entire plane of state changes unconstrained.
\end{practicebox}

\section{Orthogonal matrices: rigid motions}

\begin{definition}
$\mat{Q}\in\R^{n\times n}$ is \emph{orthogonal} if $\mat{Q}\T \mat{Q} = \mat{I}$, equivalently
$\mat{Q}^{-1}=\mat{Q}\T$, equivalently the columns of $\mat{Q}$ form an orthonormal basis of $\R^n$.
\end{definition}
\begin{practicebox}
Orthogonal transformations preserve lengths, angles, and numerical scale. That is why QR and SVD algorithms prefer them: changing coordinates does not amplify the data before the actual scaling step is exposed.
\end{practicebox}

The defining property is that orthogonal matrices \emph{change nothing that
matters geometrically}:
\begin{equation}\label{eq:isometry}
   \|\mat{Q}\vekt{x}\|^2 = (\mat{Q}\vekt{x})\T(\mat{Q}\vekt{x}) = \vekt{x}\T \mat{Q}\T \mat{Q} \vekt{x} = \vekt{x}\T \vekt{x} = \|\vekt{x}\|^2 ,
\end{equation}
and likewise $(\mat{Q}\vekt{x})\T(\mat{Q}\vekt{y})=\vekt{x}\T \vekt{y}$: lengths and angles are preserved. Geometrically
$\mat{Q}$ is a rotation, a reflection, or a composition of the two.

We will also use \emph{rectangular} matrices with orthonormal columns:
$\mat{Q}\in\R^{m\times n}$ with $m\ge n$ and $\mat{Q}\T \mat{Q}=\mat{I}_n$. Then \eqref{eq:isometry}
still holds (an isometry into a bigger space), but $\mat{Q}\mat{Q}\T \ne \mat{I}_m$ --- it is the
orthogonal projector onto $\range(\mat{Q})$.

\begin{example}
$\mat{Q}=\frac{1}{\sqrt2}\begin{pmatrix}1&-1\\1&1\end{pmatrix}$ is a $45^\circ$
rotation. Check: $\mat{Q}\T \mat{Q} = \half\begin{pmatrix}1&1\\-1&1\end{pmatrix}
\begin{pmatrix}1&-1\\1&1\end{pmatrix}
=\half\begin{pmatrix}2&0\\0&2\end{pmatrix}=\mat{I}$ \checkmark. It maps $(1,0)\T$ to
$\frac{1}{\sqrt2}(1,1)\T$, which indeed has length $1$.
\end{example}
\begin{practicebox}
This is the simplest concrete reminder that coordinates may change while geometry does not. Later the matrices $\mat{U}$ and $\mat{V}$ play exactly this harmless role in the two measurement spaces.
\end{practicebox}

\begin{remark}[Why this matters for the GSVD]
In the ordinary SVD, \emph{all three} factors involve orthogonal matrices, so
the decomposition never distorts lengths --- it only rotates and stretches along
axes. In the GSVD, the two outer factors $\mat{U},\mat{V}$ are orthogonal but the inner
factor $\mat{X}$ is merely \emph{invertible}. Precisely this relaxation buys us
simultaneity.
\end{remark}
\begin{practicebox}
The GSVD keeps the output changes $\mat{U}$ and $\mat{V}$ orthogonal but allows the shared state frame $\mat{X}$ to be oblique. That extra freedom is precisely what makes simultaneous simplification of two different operators possible.
\end{practicebox}

\section{Symmetric matrices, the spectral theorem, definiteness}

\begin{definition}
$\mat{S}\in\R^{n\times n}$ is \emph{symmetric} if $\mat{S}=\mat{S}\T$.
\end{definition}
\begin{practicebox}
Gram matrices, covariance matrices, and Fisher information matrices are symmetric. Symmetry is what gives them real eigenvalues and geometrically meaningful orthogonal eigendirections.
\end{practicebox}

\begin{theorem}[Spectral theorem]\label{thm:spectral}
Every real symmetric $\mat{S}$ can be written $\mat{S} = \mat{Q}\mat{\Lambda} \mat{Q}\T$ with $\mat{Q}$ orthogonal
and $\mat{\Lambda}=\diag(\lambda_1,\dots,\lambda_n)$ real. The columns $\vekt{q}_i$ of $\mat{Q}$
are eigenvectors, $\mat{S}\vekt{q}_i=\lambda_i \vekt{q}_i$, and they are orthonormal.
\end{theorem}
\begin{practicebox}
A symmetric uncertainty or information matrix can therefore be read as independent principal axes with scalar strengths. In practice this converts a coupled quadratic form into directions that can be ranked and interpreted.
\end{practicebox}

\begin{proof}[Proof of the two key pieces]
\emph{Eigenvalues are real.} Suppose $\mat{S}\vekt{q}=\lambda\vekt{q}$ with possibly complex
$\lambda,\vekt{q}$, $\vekt{q}\ne\vekt{0}$. Multiply by the conjugate transpose $\bar{\vekt{q}}\T$:
$\bar{\vekt{q}}\T \mat{S}\vekt{q} = \lambda\, \bar{\vekt{q}}\T \vekt{q}$. The left side is real, because
$\overline{\bar{\vekt{q}}\T \mat{S}\vekt{q}} = \vekt{q}\T \mat{S} \bar{\vekt{q}} = (\bar{\vekt{q}}\T \mat{S}\T \vekt{q})\T = \bar{\vekt{q}}\T \mat{S}\vekt{q}$
using $\mat{S}=\mat{S}\T$ real. The right side is $\lambda\|\vekt{q}\|^2$ with $\|\vekt{q}\|^2>0$ real.
Hence $\lambda$ is real.

\emph{Eigenvectors for distinct eigenvalues are orthogonal.} Let $\mat{S}\vekt{q}_1=\lambda_1\vekt{q}_1$, $\mat{S}\vekt{q}_2=\lambda_2 \vekt{q}_2$, $\lambda_1\ne\lambda_2$. Then
\[
\lambda_1 \vekt{q}_2\T \vekt{q}_1 = \vekt{q}_2\T \mat{S} \vekt{q}_1 = (\mat{S} \vekt{q}_2)\T \vekt{q}_1 = \lambda_2 \vekt{q}_2\T \vekt{q}_1 ,
\]
so $(\lambda_1-\lambda_2)\,\vekt{q}_2\T \vekt{q}_1 = 0$, forcing $\vekt{q}_2\T \vekt{q}_1=0$.
(The remaining work --- handling repeated eigenvalues --- is a standard
induction on a deflated subspace and is omitted.)
\end{proof}

\begin{definition}[Definiteness]
A symmetric $\mat{S}$ is \emph{positive semidefinite} (psd) if $\vekt{x}\T \mat{S}\vekt{x}\ge0$ for all
$\vekt{x}$, and \emph{positive definite} (pd) if $\vekt{x}\T \mat{S}\vekt{x}>0$ for all $\vekt{x}\ne\vekt{0}$.
Equivalently: psd iff all $\lambda_i\ge0$; pd iff all $\lambda_i>0$.
\end{definition}
\begin{practicebox}
Positive definiteness means that every nonzero state direction carries positive energy or information; semidefiniteness allows blind directions. This distinction decides whether an ordinary inverse covariance exists or whether a pseudoinverse/subspace treatment is required.
\end{practicebox}

\begin{lemma}\label{lem:ATA}
For any $\mat{A}\in\R^{m\times n}$, the matrix $\mat{A}\T \mat{A}$ is symmetric psd, and
$\nullsp(\mat{A}\T \mat{A})=\nullsp(\mat{A})$, hence $\rank(\mat{A}\T \mat{A})=\rank(\mat{A})$.
\end{lemma}
\begin{practicebox}
Passing from $\mat{A}$ to $\mat{A}\T \mat{A}$ preserves the exact null space while converting response lengths into a quadratic form. Thus the Gram matrix changes the representation, not which state directions are observable.
\end{practicebox}

\begin{proof}
Symmetry: $(\mat{A}\T \mat{A})\T = \mat{A}\T \mat{A}$. Semidefiniteness:
$\vekt{x}\T \mat{A}\T \mat{A} \vekt{x} = (\mat{A}\vekt{x})\T(\mat{A}\vekt{x})=\|\mat{A}\vekt{x}\|^2\ge0$. For the null spaces: if $\mat{A}\vekt{x}=\vekt{0}$ then
clearly $\mat{A}\T \mat{A}\vekt{x}=\vekt{0}$. Conversely if $\mat{A}\T \mat{A}\vekt{x}=\vekt{0}$ then
$0 = \vekt{x}\T \mat{A}\T \mat{A} \vekt{x} = \|\mat{A}\vekt{x}\|^2$, so $\mat{A}\vekt{x}=\vekt{0}$.
\end{proof}

Lemma~\ref{lem:ATA} is the engine of the SVD. Note also the identity we will
reuse many times:
\begin{equation}\label{eq:quadnorm}
  \boxed{\;\|\mat{A}\vekt{x}\|^2 = \vekt{x}\T (\mat{A}\T \mat{A})\, \vekt{x}\;}
\end{equation}
\emph{Every question about the lengths $\|\mat{A}\vekt{x}\|$ is a question about the symmetric
matrix $\mat{A}\T \mat{A}$.}

\begin{collective}
If $\mat{A}$ is a whitened measurement Jacobian (Section~\ref{sec:fisher}), then
$\mat{A}\T \mat{A}$ is exactly the \textbf{Fisher information matrix} that agent contributes
about $\vekt{x}$. Equation~\eqref{eq:quadnorm} therefore says: the squared response
length \emph{is} the information content along direction $\vekt{x}$. Everything the
GSVD does with norms, it is simultaneously doing with information.
\end{collective}

\section{Quadratic forms and the Rayleigh quotient}

For symmetric $\mat{S}$ and $\vekt{x}\ne\vekt{0}$, define
\[
   \mathcal{R}(\vekt{x}) \;=\; \frac{\vekt{x}\T \mat{S} \vekt{x}}{\vekt{x}\T \vekt{x}}\,.
\]

\begin{theorem}[Rayleigh]\label{thm:rayleigh}
$\displaystyle\max_{\vekt{x}\ne\vekt{0}} \mathcal{R}(\vekt{x}) = \lambda_{\max}(\mat{S})$ and
$\displaystyle\min_{\vekt{x}\ne\vekt{0}} \mathcal{R}(\vekt{x}) = \lambda_{\min}(\mat{S})$, attained at the
corresponding eigenvectors.
\end{theorem}
\begin{practicebox}
To find the strongest or weakest direction of a symmetric quadratic form, one need not search over all vectors: solve an eigenproblem. The generalized version of this idea is the variational core of the GSVD.
\end{practicebox}

\begin{proof}
Write $\mat{S}=\mat{Q}\mat{\Lambda} \mat{Q}\T$ and substitute $\vekt{y}=\mat{Q}\T \vekt{x}$; since $\mat{Q}$ is orthogonal,
$\vekt{x}\T \vekt{x}=\vekt{y}\T \vekt{y}$ and $\vekt{x}\T \mat{S} \vekt{x} = \vekt{y}\T\mat{\Lambda} \vekt{y}=\sum_i\lambda_i y_i^2$. So
\[
   \mathcal{R} = \frac{\sum_i \lambda_i y_i^2}{\sum_i y_i^2},
\]
a weighted average of the $\lambda_i$ with nonnegative weights $y_i^2$ summing
(after normalisation) to $1$. An average of numbers lies between the smallest and
the largest, with equality when all the weight sits on one of them --- i.e.\ when
$\vekt{y}=\vekt{e}_i$, i.e.\ $\vekt{x}=\vekt{q}_i$.
\end{proof}

This little theorem is the whole variational story. Everything about extremal
singular values follows from it.

\section{The ordinary SVD}

\begin{theorem}[SVD]\label{thm:svd}
Every $\mat{A}\in\R^{m\times n}$ factors as
\[
   \mat{A} = \mat{U}\mat{\Sigma} \mat{V}\T,
\]
with $\mat{U}\in\R^{m\times m}$ orthogonal, $\mat{V}\in\R^{n\times n}$ orthogonal, and
$\mat{\Sigma}\in\R^{m\times n}$ ``diagonal'' with entries
$\sigma_1\ge\sigma_2\ge\cdots\ge\sigma_{\min(m,n)}\ge0$.
\end{theorem}
\begin{practicebox}
The SVD is the canonical audit of a single linear sensor or operator: it separates input directions, gains, and output directions. The GSVD will repeat the same logic for two operators that share the same state.
\end{practicebox}

\begin{proof}
By Lemma~\ref{lem:ATA}, $\mat{A}\T \mat{A}$ is symmetric psd, so by the spectral theorem
$\mat{A}\T \mat{A} = \mat{V}\mat{\Lambda} \mat{V}\T$ with $\mat{V}$ orthogonal and $\lambda_1\ge\cdots\ge\lambda_n\ge0$.
Put $\sigma_i=\sqrt{\lambda_i}$ and let $r$ be the number of positive $\sigma_i$;
by Lemma~\ref{lem:ATA}, $r=\rank(\mat{A})$.

For $i\le r$ define
\[
  \vekt{u}_i \;=\; \frac{1}{\sigma_i}\,\mat{A} \vekt{v}_i .
\]
These are orthonormal:
\[
  \vekt{u}_i\T \vekt{u}_j \;=\; \frac{1}{\sigma_i\sigma_j}\, \vekt{v}_i\T \mat{A}\T \mat{A}\, \vekt{v}_j
             \;=\; \frac{\lambda_j}{\sigma_i\sigma_j}\, \vekt{v}_i\T \vekt{v}_j
             \;=\; \frac{\lambda_j}{\sigma_i\sigma_j}\,\delta_{ij}
             \;=\;\delta_{ij},
\]
using $\mat{A}\T \mat{A} \vekt{v}_j=\lambda_j \vekt{v}_j$ and, for $i=j$, $\lambda_i/\sigma_i^2=1$.
Extend $\vekt{u}_1,\dots,\vekt{u}_r$ to an orthonormal basis $\vekt{u}_1,\dots,\vekt{u}_m$ of $\R^m$
(always possible: complete a basis and run Gram--Schmidt).

Now verify $\mat{A}\mat{V}=\mat{U}\mat{\Sigma}$ column by column. For $i\le r$,
$\mat{A}\vekt{v}_i=\sigma_i \vekt{u}_i$ by construction. For every $i>r$, $\lambda_i=0$, hence
$\|\mat{A}\vekt{v}_i\|^2=\vekt{v}_i\T \mat{A}\T \mat{A} \vekt{v}_i=0$ and therefore $\mat{A}\vekt{v}_i=\vekt{0}$. The $i$-th column of
$\mat{U}\mat{\Sigma}$ is also zero because the $i$-th column of $\mat{\Sigma}$ is zero. In the
wide case $i>m$, there is no left singular vector $\vekt{u}_i$ to index --- nor is one
needed. Thus every column agrees, $\mat{A}\mat{V}=\mat{U}\mat{\Sigma}$, and multiplying on the right by
$\mat{V}\T$ gives $\mat{A}=\mat{U}\mat{\Sigma} \mat{V}\T$.
\end{proof}

\paragraph{What the SVD says geometrically.}
Read $\mat{A}=\mat{U}\mat{\Sigma} \mat{V}\T$ right to left, acting on $\vekt{x}$:
\begin{enumerate}[leftmargin=1.6em,itemsep=1pt]
\item $\mat{V}\T \vekt{x}$ --- \emph{rotate} into the input frame $\vekt{v}_1,\dots,\vekt{v}_n$.
\item $\mat{\Sigma}(\cdot)$ --- \emph{stretch} axis $i$ by $\sigma_i$ (and change dimension).
\item $\mat{U}(\cdot)$ --- \emph{rotate} into the output frame $\vekt{u}_1,\dots,\vekt{u}_m$.
\end{enumerate}
Every linear map is a rotation, then a pure axis-aligned stretch, then a
rotation. The unit sphere maps to an ellipsoid with semi-axes $\sigma_i \vekt{u}_i$.

\begin{figure}[htbp]
\centering
\begin{tikzpicture}[figbase,scale=0.80]
 \begin{scope}
  \draw[ax] (-1.9,0)--(1.9,0); \draw[ax] (0,-1.9)--(0,1.9);
  \draw[struct!45,crv] (0,0) circle (1);
  \draw[vec,srcA] (0,0)--(0.7071,0.7071);
  \draw[vec,srcB] (0,0)--(0.7071,-0.7071);
  \node[lbl,text=srcA] at (1.05,0.86) {$\vekt v_1$};
  \node[lbl,text=srcB] at (1.05,-0.86) {$\vekt v_2$};
  \node[fcap] at (0,-2.55) {unit circle};
 \end{scope}
 \draw[flow,pooled] (2.5,0)--(3.9,0);
 \node[pooled,above] at (3.2,0.10) {$\mat M$};
 \begin{scope}[xshift=8.0cm]
  \draw[ax] (-3.2,0)--(3.6,0); \draw[ax] (0,-2.9)--(0,3.2);
  \draw[struct!45,crv,rotate=45] (0,0) ellipse (3 and 1);
  \draw[vec,srcA] (0,0)--(2.1213,2.1213);
  \draw[vec,srcB] (0,0)--(-0.7071,0.7071);
  \node[lbl,text=srcA] at (2.05,2.62) {$\sigma_1\vekt u_1,\ \sigma_1=3$};
  \node[lbl,text=srcB] at (-1.55,1.02) {$\sigma_2\vekt u_2$};
  \node[fcap] at (0,-2.55) {image ellipse};
 \end{scope}
\end{tikzpicture}
\caption{\textbf{Every linear map is a rotation, an axis-aligned stretch, and a second rotation.} The unit circle maps to an ellipse whose semi-axes are the singular values, laid out along the left singular vectors; the right singular vectors mark the directions in the domain that are sent to them. The GSVD generalizes this picture by giving two operators a \emph{shared} input frame while each keeps its own orthonormal output frame.}
\label{fig:svd-geometry}
\end{figure}
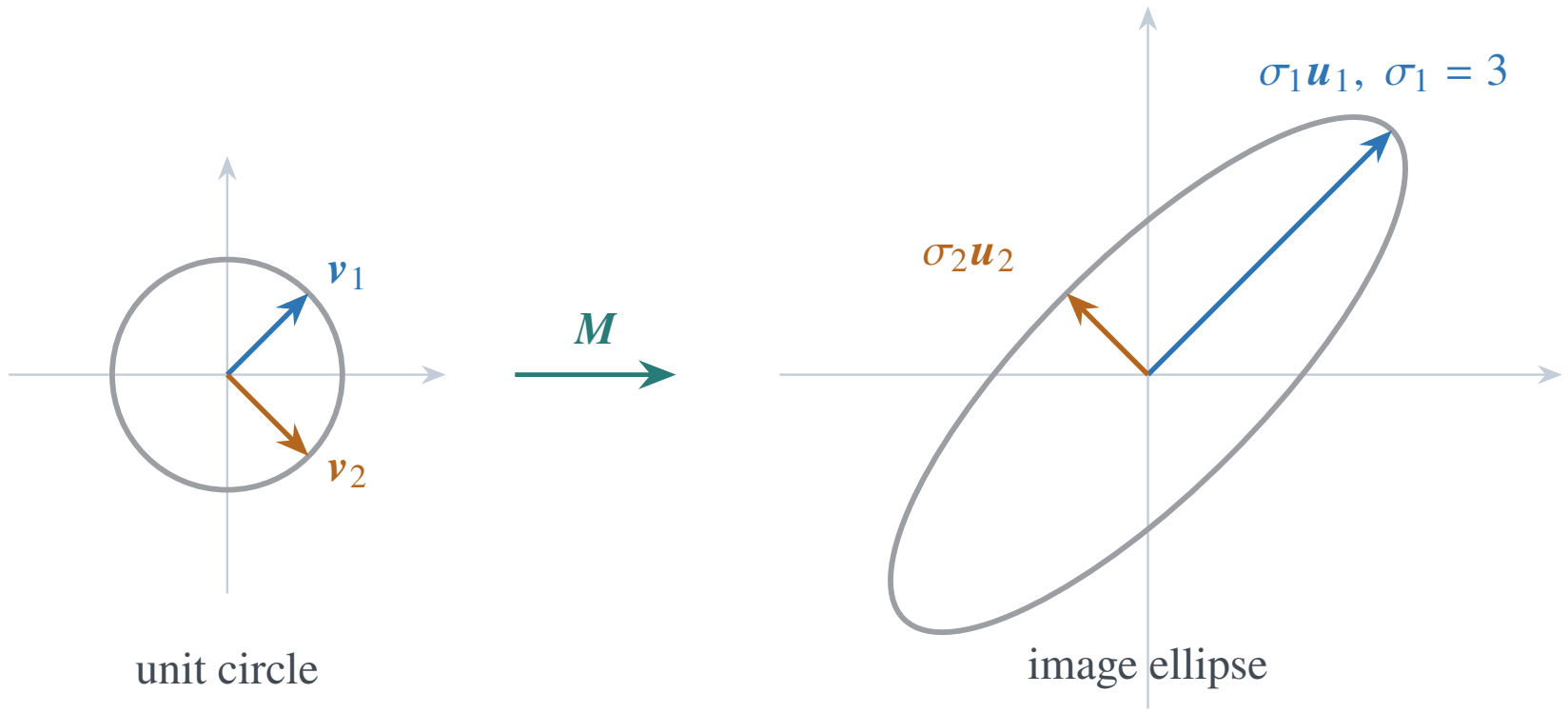

Figure~\ref{fig:svd-geometry} shows the composite map for a two-dimensional
case: the unit circle with its right singular vectors on the left, the image
ellipse with semi-axes $\sigma_i\vekt{u}_i$ on the right.

\FloatBarrier

\paragraph{The subspaces, read off.}
$\{\vekt{u}_1,\dots,\vekt{u}_r\}$ is an orthonormal basis of $\range(\mat{A})$;
$\{\vekt{u}_{r+1},\dots,\vekt{u}_m\}$ of $\nullsp(\mat{A}\T)$;
$\{\vekt{v}_1,\dots,\vekt{v}_r\}$ of $\range(\mat{A}\T)$;
$\{\vekt{v}_{r+1},\dots,\vekt{v}_n\}$ of $\nullsp(\mat{A})$.

\paragraph{Extremal property.} By Theorem~\ref{thm:rayleigh} applied to
$\mat{S}=\mat{A}\T \mat{A}$ with \eqref{eq:quadnorm},
\begin{equation}\label{eq:sigmax}
   \sigma_{\max}(\mat{A})=\max_{\vekt{x}\ne\vekt{0}}\frac{\|\mat{A}\vekt{x}\|}{\|\vekt{x}\|},
   \qquad
   \sigma_{\min}(\mat{A})=\min_{\vekt{x}\ne\vekt{0}}\frac{\|\mat{A}\vekt{x}\|}{\|\vekt{x}\|}\quad(m\ge n).
\end{equation}

\subsection{A fully worked SVD}\label{sec:svdex}

Take the symmetric matrix we will reuse throughout:
\[
   \mat{M} = \begin{pmatrix}1&2\\2&1\end{pmatrix}.
\]

\textbf{Step 1: form $\mat{M}\T \mat{M}$.}
\[
\mat{M}\T \mat{M} = \begin{pmatrix}1&2\\2&1\end{pmatrix}\begin{pmatrix}1&2\\2&1\end{pmatrix}
      = \begin{pmatrix}1+4 & 2+2\\ 2+2 & 4+1\end{pmatrix}
      = \begin{pmatrix}5&4\\4&5\end{pmatrix}.
\]

\textbf{Step 2: eigenvalues.} $\det(\mat{M}\T \mat{M}-\lambda \mat{I}) = (5-\lambda)^2-16=0$, so
$5-\lambda=\pm4$, giving $\lambda_1=9$, $\lambda_2=1$. Hence
$\sigma_1=3$, $\sigma_2=1$.

\textbf{Step 3: eigenvectors.} Write $\vekt{v}=(v_1,v_2)\T$ for the eigenvector sought. For $\lambda=9$:
$(5-9)v_1+4v_2=0 \Rightarrow -4v_1+4v_2=0\Rightarrow v_1=v_2$, so
$\vekt{v}_1=\frac{1}{\sqrt2}(1,1)\T$.
For $\lambda=1$: $4v_1+4v_2=0\Rightarrow v_1=-v_2$, so
$\vekt{v}_2=\frac{1}{\sqrt2}(1,-1)\T$. Orthogonal \checkmark.

\textbf{Step 4: left vectors.}
\[
\vekt{u}_1=\frac{\mat{M}\vekt{v}_1}{\sigma_1}=\frac{1}{3}\cdot\frac{1}{\sqrt2}\binom{1+2}{2+1}
   =\frac{1}{3}\cdot\frac{1}{\sqrt2}\binom{3}{3}=\frac{1}{\sqrt2}\binom{1}{1},
\]
\[
\vekt{u}_2=\frac{\mat{M}\vekt{v}_2}{\sigma_2}=\frac{1}{1}\cdot\frac{1}{\sqrt2}\binom{1-2}{2-1}
   =\frac{1}{\sqrt2}\binom{-1}{1}.
\]

\textbf{Step 5: assemble and check.}
\[
\mat{U}=\frac{1}{\sqrt2}\begin{pmatrix}1&-1\\1&1\end{pmatrix},\quad
\mat{\Sigma}=\begin{pmatrix}3&0\\0&1\end{pmatrix},\quad
\mat{V}=\frac{1}{\sqrt2}\begin{pmatrix}1&1\\1&-1\end{pmatrix}.
\]
Verify:
\[
\mat{U}\mat{\Sigma} \mat{V}\T=\frac{1}{\sqrt2}\begin{pmatrix}1&-1\\1&1\end{pmatrix}
\begin{pmatrix}3&0\\0&1\end{pmatrix}\frac{1}{\sqrt2}\begin{pmatrix}1&1\\1&-1\end{pmatrix}
=\half\begin{pmatrix}3&-1\\3&1\end{pmatrix}\begin{pmatrix}1&1\\1&-1\end{pmatrix}
=\half\begin{pmatrix}2&4\\4&2\end{pmatrix}=\begin{pmatrix}1&2\\2&1\end{pmatrix}\;\checkmark
\]

\subsection{Two consequences we will need}

\paragraph{Pseudoinverse.} With $\mat{A}=\mat{U}\mat{\Sigma} \mat{V}\T$ of rank $r$,
$\mat{A}^{+}=\mat{V}\mat{\Sigma}^{+}\mat{U}\T$ where $\mat{\Sigma}^{+}$ inverts the nonzero $\sigma_i$ and
transposes. Then $\vekt{x}=\mat{A}^{+}\vekt{b}$ is the minimum-norm least-squares solution of
$\mat{A}\vekt{x}\approx\vekt{b}$.

\paragraph{Condition number.} If $\mat{A}$ has full column rank, then its smallest
singular value is positive and
\[
   \kappa_2(\mat{A}\T \mat{A})=\frac{\lambda_{\max}}{\lambda_{\min}}
                  =\frac{\sigma_{\max}^2}{\sigma_{\min}^2}
                  =\kappa_2(\mat{A})^2 .
\]
For a rank-deficient or wide matrix, $\mat{A}\T \mat{A}$ is singular. On the constrained
subspace the corresponding statement remains true if
$\kappa_{+}$ denotes the ratio of the largest to the smallest \emph{positive}
singular value:
\[
   \kappa_{+}(\mat{A}\T \mat{A})=\kappa_{+}(\mat{A})^2.
\]

\begin{remark}[The single most important numerical rule in this text]
For a full-column-rank matrix, forming $\mat{A}\T \mat{A}$ \textbf{squares} the condition
number. If $\kappa_2(\mat{A})=10^{8}$ ---
entirely ordinary for a GNSS geometry matrix near a bad satellite constellation,
or a photogrammetric normal-equation system --- then $\kappa_2(\mat{A}\T \mat{A})=10^{16}$,
which is the reciprocal of double-precision machine epsilon. You have lost
\emph{all} accuracy. The SVD works on $\mat{A}$ directly and never forms $\mat{A}\T \mat{A}$; the
proof above uses $\mat{A}\T \mat{A}$ only as a theoretical device. The same discipline
governs the GSVD: we will \emph{characterise} it through $\mat{A}\T \mat{A}$ and $\mat{B}\T \mat{B}$ and
\emph{compute} it through $\mat{A}$ and $\mat{B}$.
\end{remark}
\begin{practicebox}
Normal equations can destroy useful digits even when the original matrix problem is still accurately solvable. For implementation, use QR/SVD/GSVD algorithms on the original factors and reserve Gram matrices for analysis and interpretation.
\end{practicebox}


\section{Lie groups, tangent perturbations, and robotic state geometry}\label{sec:foundation-lie}

The previous sections treated state directions as vectors in a fixed Euclidean
space. That is sufficient for many calibration parameters, biases, velocities,
and landmark coordinates, but it is not globally correct for the quantities at
the heart of robotics: rotations and rigid-body poses. A rotation belongs to
$\mathrm{SO}(3)$ and a pose to $\mathrm{SE}(3)$; neither set is a vector space.
Modern estimation therefore works with small perturbations in a local tangent
space and maps them back to the group through the exponential map. This
``micro Lie theory'' has become standard machinery in SLAM, visual odometry,
inertial navigation, and geometric filtering \cite{SolaDerayAtchuthan2018,Barfoot2024}.

For the present volume the role of Lie theory is sharply delimited:
\begin{center}
\fbox{\parbox{0.91\textwidth}{
\textbf{Lie geometry specifies a geometrically consistent local coordinate system for robotic states; the GSVD then decomposes source information in that local linearised space.}
}}
\end{center}
The group structure does \emph{not} by itself guarantee that a first-order
linearisation is accurate. Accuracy still depends on the perturbation being
small enough, the sensor model being sufficiently smooth, and the chosen
operating point being representative.

\begin{figure}[htbp]
\centering
\begin{tikzpicture}[figbase,scale=1.15]
  \draw[struct!40,crv,fill=softgray]
     plot[smooth cycle,tension=0.75]
     coordinates {(-4.3,-1.15)(-1.2,-2.25)(2.9,-1.75)(4.5,0.3)(1.2,1.6)(-3.2,1.0)};
  \node[struct!60] at (-2.70,-1.48) {$\mathcal M$};
  \draw[srcA!60,crv,fill=softblue]
     (-2.15,-0.55) -- (1.25,-1.05) -- (2.15,0.55) -- (-1.25,1.05) -- cycle;
  \node[text=srcA,anchor=west] at (2.35,-0.75) {$T_{\bar{\mat T}}\mathcal M\simeq\mathfrak{se}(3)$};
  \fill[struct] (-0.15,-0.10) circle (1.6pt);
  \node[text=struct,anchor=north] at (-0.15,-0.28) {$\bar{\mat T}$};
  \draw[vec,pooled] (-0.15,-0.10)--(1.55,0.12);
  \node[text=pooled,anchor=west] at (2.08,0.14) {$\bm\xi_R$};
  \draw[vec,warn] (-0.15,-0.10)--(-0.55,0.88);
  \node[text=warn,anchor=south] at (-0.55,0.96) {$\bm\xi_L$};
  \draw[struct!55,thincrv,-{Stealth[length=1.9mm,width=1.5mm]}]
     (1.20,0.55) to[out=140,in=5] (-0.30,1.02);
  \node[text=struct!75,anchor=south west] at (0.55,1.02) {$\operatorname{Ad}_{\bar{\mat T}}$};
\end{tikzpicture}

\vspace{0.5em}
\begin{tikzpicture}[figbase]
  \node[text=struct!85] at (0,0)
    {$\mathcal M\ \xrightarrow{\ \text{operating point}\ }\ T_{\bar{\mat T}}\mathcal M
      \ \xrightarrow{\ \text{linearise}\ }\ \{\mat H_k\}
      \ \xrightarrow{\ \text{whiten}\ }\ \{\mat A_k\}
      \ \xrightarrow{\ \text{GSVD}\ }\ \text{source-resolved directions}$};
\end{tikzpicture}
\caption{\textbf{Where the local linear analysis is represented.} A robotic state is a point on a manifold, so the Jacobians $\mat H_k$ are defined only once a tangent space and a perturbation convention have been fixed. Left and right conventions differ by the adjoint $\operatorname{Ad}_{\bar{\mat T}}$: the coordinate vector of a direction changes, while every scalar source-attribution quantity does not. This is the geometric precondition for everything the text does with $\mat A$ and $\mat B$, and the reason the chapter sits before the linearisation section rather than after it.}
\label{fig:lie-to-gsvd}
\end{figure}
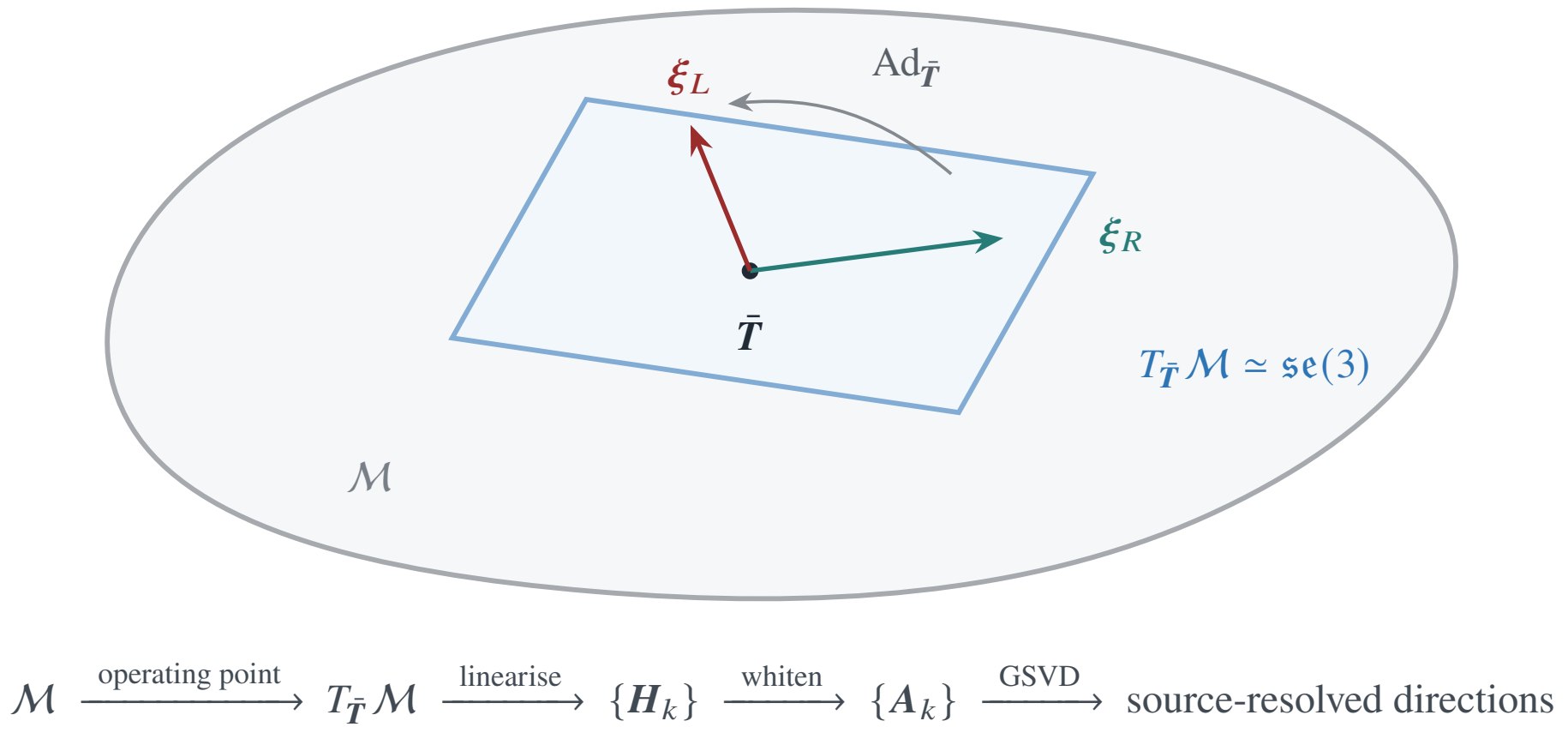

Figure~\ref{fig:lie-to-gsvd} shows the resulting chain in full --- manifold,
tangent space, Jacobians, whitening, source-resolved directions --- and records
which quantities survive a change of perturbation convention and which do not.

\FloatBarrier

\subsection{From \texorpdfstring{$\mathrm{SO}(3)$ and $\mathrm{SE}(3)$}{SO(3) and SE(3)} to vectors again}

A rotation matrix satisfies
\[
  \mat R\in\mathrm{SO}(3)
  :=\{\mat R\in\R^{3\times3}:\mat R\T\mat R=\mat I,\ \det\mat R=1\}.
\]
Its Lie algebra $\mathfrak{so}(3)$ consists of skew-symmetric matrices.  For
$\vekt\phi=(\phi_x,\phi_y,\phi_z)\T$ define the hat map
\[
 \vekt\phi^\wedge=
 \begin{bmatrix}
 0&-\phi_z&\phi_y\\
 \phi_z&0&-\phi_x\\
 -\phi_y&\phi_x&0
 \end{bmatrix},
 \qquad
 \vekt\phi^\wedge\vekt a=\vekt\phi\times\vekt a,
\]
and let $(\cdot)^\vee$ denote its inverse on $\mathfrak{so}(3)$.  Locally,
\[
  \mat R=\bar{\mat R}\,\Exp(\vekt\phi^\wedge)
\]
represents a rotation near $\bar{\mat R}$ by a three-vector
$\vekt\phi\in\R^3$.  The matrix exponential maps tangent perturbations to the
group; the local logarithm map performs the reverse operation.

A pose is represented by
\[
 \mat T=
 \begin{bmatrix}
   \mat R&\vekt p\\
   \vekt 0\T&1
 \end{bmatrix}\in\mathrm{SE}(3).
\]
With the ordering translation--rotation, write
\[
 \vekt\xi=
 \begin{bmatrix}\vekt\rho\\\vekt\phi\end{bmatrix}\in\R^6,
 \qquad
 \vekt\xi^\wedge=
 \begin{bmatrix}
  \vekt\phi^\wedge&\vekt\rho\\
  \vekt 0\T&0
 \end{bmatrix}\in\mathfrak{se}(3).
\]
A \emph{right} perturbation and a \emph{left} perturbation are, respectively,
\[
 \mat T=\bar{\mat T}\Exp(\vekt\xi_R^\wedge),
 \qquad
 \mat T=\Exp(\vekt\xi_L^\wedge)\bar{\mat T}.
\]
They describe the same nearby pose with different tangent coordinates. To
first order -- and exactly at the level of tangent vectors -- they are related
by the adjoint
\[
 \vekt\xi_L=\Ad_{\bar{\mat T}}\vekt\xi_R,
 \qquad
 \Ad_{\mat T}=
 \begin{bmatrix}
   \mat R&\vekt p^\wedge\mat R\\
   \mat 0&\mat R
 \end{bmatrix}.
\]
The distinction is therefore a coordinate convention, not a difference in the
underlying physical perturbation. Sol\`a et al. give a robotics-oriented
catalogue of these operators and Jacobians, while Barfoot develops the same
machinery systematically for state estimation \cite{SolaDerayAtchuthan2018,Barfoot2024}.

\subsection{Jacobians on the tangent space}

Let a sensor have model $\vekt z=\vekt h(\mat T)+\vekt n$. Under a right
perturbation, the local Jacobian is defined by
\[
 \vekt h\!\left(\bar{\mat T}\Exp(\vekt\xi_R^\wedge)\right)
 \approx
 \vekt h(\bar{\mat T})+\mat H_R\vekt\xi_R.
\]
Under the left convention,
\[
 \vekt h\!\left(\Exp(\vekt\xi_L^\wedge)\bar{\mat T}\right)
 \approx
 \vekt h(\bar{\mat T})+\mat H_L\vekt\xi_L.
\]
Since $\vekt\xi_L=\Ad_{\bar{\mat T}}\vekt\xi_R$ for the same physical tangent
vector,
\[
 \boxed{\mat H_L=\mat H_R\Ad_{\bar{\mat T}}^{-1}.}
\]
This identity matters directly for the GSVD: a Jacobian is not fully specified
until the state perturbation convention and frame are specified.

With covariance $\mat\Sigma_k$ and whitened Jacobian
\[
 \mat A_k=\mat\Sigma_k^{-1/2}\mat H_k,
\]
the local Fisher information is
\[
 \mat J_k=\mat H_k\T\mat\Sigma_k^{-1}\mat H_k=\mat A_k\T\mat A_k.
\]
Changing from right to left coordinates applies the same congruence transform
to every source,
\[
 \boxed{
 \mat J_{k,L}=\Ad_{\bar{\mat T}}^{-\mathsf T}
               \mat J_{k,R}
               \Ad_{\bar{\mat T}}^{-1}.}
\]

\begin{proposition}[Coordinate invariance of pairwise GSVD source shares]\label{prop:lie-gsvd-invariance}
Let two whitened source Jacobians $\mat A_R,\mat B_R$ be expressed in one
set of tangent coordinates and let $\mat G$ be any nonsingular common change
of state coordinates. Define
\[
  \mat A_L=\mat A_R\mat G^{-1},\qquad
  \mat B_L=\mat B_R\mat G^{-1}.
\]
Then the generalized singular values, and hence the normalized pairwise shares
$\alpha_i^2$ and $\beta_i^2$, are unchanged. If $\vekt x_{i,R}$ is a
physical generalized direction in the first coordinates, the corresponding
direction is $\vekt x_{i,L}=\mat G\vekt x_{i,R}$, and for each source
\[
  \vekt x_{i,L}\T\mat J_{k,L}\vekt x_{i,L}
  =\vekt x_{i,R}\T\mat J_{k,R}\vekt x_{i,R}.
\]
In particular, left/right $\mathrm{SE}(3)$ perturbation conventions with
$\mat G=\Ad_{\bar{\mat T}}$ cannot change any scalar source-attribution
quantity when the same physical direction is compared.
\end{proposition}

\begin{proof}
The generalized eigenproblem associated with the full-column-rank GSVD is
\[
 \mat J_A\vekt x=\lambda(\mat J_A+\mat J_B)\vekt x,
 \qquad \lambda=\alpha^2.
\]
Under the common coordinate change,
$\mat J_{k,L}=\mat G^{-\mathsf T}\mat J_{k,R}\mat G^{-1}$.
Substituting $\vekt x_L=\mat G\vekt x_R$ cancels the two factors
$\mat G^{-1}\mat G$ and gives exactly the original generalized eigenproblem.
Thus the generalized eigenvalues are identical. The quadratic-form identity
follows directly:
\[
 (\mat G\vekt x)\T\mat G^{-\mathsf T}\mat J_k\mat G^{-1}
 (\mat G\vekt x)=\vekt x\T\mat J_k\vekt x.
\]
Since $\beta_i^2=1-\alpha_i^2$ under the pooled normalization, both shares and
$\sigma_i=\alpha_i/\beta_i$ are invariant.
\end{proof}

\begin{practicebox}
Left versus right perturbations can make the \emph{coordinate entries} of a
GSVD direction look very different, yet they do not create a new physical
source-attribution result. The scalar attribution is invariant under a common
nonsingular tangent-coordinate change. The correct operation is to transport
the direction, not to compare its raw coordinate vector component by component.
A recent analysis of left- and right-IEKFs likewise shows equivalence when the
state distribution and reset are transformed consistently \cite{GeEtAl2026}.
\end{practicebox}

\subsection{A worked \texorpdfstring{$\mathrm{SE}(3)$}{SE(3)} example: GNSS versus body-frame landmark sensing}\label{sec:lie-worked}

Consider a platform at
\[
 \vekt p=(2,-1,0.5)\T\ \mathrm{m},
\]
with orientation
\[
 \mat R=\mat R_z(30^\circ)\mat R_y(10^\circ)\mat R_x(-5^\circ),
\]
and a GNSS antenna lever arm, expressed in the body frame,
\[
 \vekt\ell=(0.5,0.2,0.1)\T\ \mathrm{m}.
\]
Use the right perturbation $\vekt\xi_R=[\vekt\rho\T,\vekt\phi\T]\T$.
The GNSS antenna position is
\[
 \vekt y_G=\vekt p+\mat R\vekt\ell.
\]
Its first-order perturbation is
\[
 \delta\vekt y_G
 =\mat R\vekt\rho-\mat R\vekt\ell^\wedge\vekt\phi,
\]
so
\[
 \boxed{\mat H_G=
 \begin{bmatrix}\mat R&-\mat R\vekt\ell^\wedge\end{bmatrix}.}
\]
Assume isotropic GNSS standard deviation $\sigma_G=0.25$ m.

A body-frame 3-D landmark sensor observes two fixed world points
\[
 \vekt q_1=(5,2,1.2)\T\ \mathrm{m},\qquad
 \vekt q_2=(1,4,2)\T\ \mathrm{m},
\]
through
\[
 \vekt z_j=\mat R\T(\vekt q_j-\vekt p).
\]
Under the same right perturbation,
\[
 \delta\vekt z_j=-\vekt\rho+\vekt z_j^\wedge\vekt\phi,
 \qquad
 \mat H_j=\begin{bmatrix}-\mat I&\vekt z_j^\wedge\end{bmatrix}.
\]
Stacking the two landmarks gives
\[
 \mat H_M=\begin{bmatrix}\mat H_1\\\mat H_2\end{bmatrix},
\]
and take isotropic standard deviation $\sigma_M=0.12$ m. The whitened pair is
\[
 \mat A=\sigma_G^{-1}\mat H_G,\qquad
 \mat B=\sigma_M^{-1}\mat H_M.
\]
The pooled information matrix
$\mat J=\mat A\T\mat A+\mat B\T\mat B$ is positive definite; its eigenvalues
are, to four significant figures,
\[
 11.45,\ 42.50,\ 154.89,\ 661.57,\ 2732.69,\ 3362.83.
\]
Thus the two sensors jointly constrain all six local pose directions in this
configuration.

Define $\mat J_G=\mat A\T\mat A$ and $\mat J_M=\mat B\T\mat B$.
Solving the generalized eigenproblem
\[
 \mat J_G\vekt x_i=\alpha_i^2\mat J\vekt x_i,
 \qquad \vekt x_i\T\mat J\vekt x_i=1,
\]
gives the nontrivial finite source splits
\[
 \begin{array}{c|ccc}
 i&\alpha_i^2\ \text{(GNSS)}&\beta_i^2\ \text{(landmarks)}&
 \sigma_i=\alpha_i/\beta_i\\ \hline
 2&0.3038&0.6962&0.6606\\
 3&0.1033&0.8967&0.3394
 \end{array}
\]
plus one GNSS-exclusive generalized direction ($\alpha^2=1$) and three
directions with zero GNSS contribution ($\alpha^2=0$), as expected from the
rank-three GNSS position measurement. For mode 2, one $\mat J$-normalized
right-coordinate vector is
\[
 \vekt x_{2,R}\approx
 \begin{bmatrix}
 0.1129&-0.0992&-0.0045&-0.0075&-0.0096&0.0238
 \end{bmatrix}\T.
\]
Direct evaluation reproduces the GSVD shares:
\[
 \vekt x_{2,R}\T\mat J_G\vekt x_{2,R}=0.3038,
 \qquad
 \vekt x_{2,R}\T\mat J_M\vekt x_{2,R}=0.6962.
\]

Now express the \emph{same physical tangent direction} with a left perturbation.
The coordinate vector becomes
\[
 \vekt x_{2,L}=\Ad_{\bar{\mat T}}\vekt x_{2,R}
 \approx
 \begin{bmatrix}
 0.1251&-0.0804&-0.0307&0.0010&-0.0081&0.0255
 \end{bmatrix}\T.
\]
The entries have visibly changed. Nevertheless, transforming both Fisher
matrices by the common adjoint congruence gives, to machine precision,
\[
 \alpha_{2,L}^2=0.3038,\qquad
 \beta_{2,L}^2=0.6962,\qquad
 \sigma_{2,L}=0.6606.
\]
This is Proposition~\ref{prop:lie-gsvd-invariance} in numbers: coordinate vectors
move; source attribution does not.

\subsection{Uncertainty on a Lie group: useful, but local}\label{sec:lie-uncertainty}

A common robotic approximation writes a random pose as a concentrated
Gaussian in tangent coordinates,
\[
 \mat T=\bar{\mat T}\Exp(\vekt\xi^\wedge),
 \qquad
 \vekt\xi\sim\mathcal N(\vekt 0,\mat P).
\]
This construction is standard in Lie-group filtering and error propagation
\cite{Chirikjian2012,HartleyEtAl2020}. It is exactly the regime in which the
covariance, whitening, Fisher-information, and GSVD machinery developed in the
next sections can be applied most transparently.

The qualification \emph{concentrated} is essential. A large or multimodal
orientation uncertainty cannot generally be represented faithfully by one
Gaussian in one tangent space. The exponential map is nonlinear and not
one-to-one globally; on $\mathrm{SO}(3)$ the topology itself prevents one
Euclidean Gaussian chart from being a global model. Consequently, a local
Fisher matrix and a GSVD of local Jacobians should not be interpreted as a
global uncertainty decomposition when the posterior spreads far outside the
linearisation neighbourhood. Lie-group probability and information theory go
well beyond the concentrated-Gaussian approximation \cite{Chirikjian2009,Chirikjian2012};
that broader theory is outside the scope of this foundations volume.

\subsection{Invariant filtering: what transfers to INS, and what does not}

Lie geometry is more than notation when the system dynamics possess a compatible
symmetry. Barrau and Bonnabel identified the group-affine class for which the
invariant estimation error has the log-linear property and used it to establish
local stability results for the IEKF \cite{BarrauBonnabel2017,BarrauBonnabel2018}.
For bias-free inertial-navigation states, this structure can make the invariant
error dynamics independent of the estimated trajectory in a way unavailable to
a conventional additive EKF.

The caveat matters for this volume's intended applications: simply appending
IMU biases as Euclidean states destroys the exact group-affine property of the
standard inertial-navigation construction. Modern work therefore develops
larger equivariant or semidirect-product symmetries to recover more of the
favourable geometry for biased INS \cite{FornasierEtAl2022}. Contact-aided
invariant filtering provides a concrete robotics example and also notes that
with sensor noise and IMU bias the ideal log-linear structure is only
approximate in the implemented system \cite{HartleyEtAl2020}.

\begin{practicebox}
For an INS/GNSS/camera/LiDAR estimator, state the group, the perturbation side,
the frame in which the tangent coordinates live, and the bias convention before
publishing any Jacobian or GSVD result. Do not claim trajectory-independent
error dynamics merely because $\mathrm{SE}(3)$ or an invariant error is used;
that property requires the appropriate group-affine or equivariant system
structure.
\end{practicebox}

\subsection{Current research frontier: geometry beyond the classical local EKF}

The contemporary literature is moving in two complementary directions. First,
rather than accepting a fixed state group and then adding nuisance states
ad hoc, recent geometric-navigation work constructs larger symmetry groups for
practical navigation states, including body-frame vectors, biases, and lever-arm
parameters. Barrau and Bonnabel's ``geometry of navigation problems'' is an
important example: it systematizes the group construction and explicitly treats
extensions that are problematic for the classical bias-free IEKF
\cite{BarrauBonnabel2023}. Equivariant filtering develops the same broader idea
from symmetry actions rather than requiring the state itself to be a simple
matrix Lie group \cite{FornasierEtAl2022}.

Second, there is active work on reducing the limitations of first-order tangent
Gaussian filtering itself. A 2026 preprint by Zhang et al. proposes a
natural-gradient Gaussian approximation on Lie groups and uses the Fisher
information of the increment variable directly in the optimization
\cite{ZhangEtAl2026NANOL}. This is not yet established literature on the level of
the IEKF results above, and it is cited here explicitly as a \emph{research
frontier}, not as settled doctrine. Its relevance to this volume is conceptual:
Lie-group state geometry and Fisher information are already being combined in
modern robotic estimation, while the present GSVD layer asks an orthogonal
question --- how that local information is divided between sources.

The resulting research boundary is therefore precise. The present volume does
not propose a manifold-valued GSVD, nor does it claim that a tangent-space GSVD
solves global nonlinear uncertainty. It uses Lie geometry to define the local
state perturbations correctly, and then applies source-resolved linear
information analysis in that local model.

\subsection{The actual open issue: comparing directions at different operating points}\label{sec:lie-transport}

Proposition~\ref{prop:lie-gsvd-invariance} closes the left-versus-right question
at a \emph{fixed} physical state: it is ordinary coordinate covariance. The
harder problem appears when the operating point itself changes. A GSVD direction
at time $t_1$ belongs to
\[
 T_{\bar x(t_1)}\mathcal M,
\]
whereas a direction at time $t_2$ belongs to
\[
 T_{\bar x(t_2)}\mathcal M.
\]
These are different tangent spaces. On a Lie group one can identify tangent
spaces by left translation, right translation, or another connection/transport
rule, but there is no reason that an application-level comparison must be
independent of that choice. The same issue arises when a planner recomputes a
GSVD at each candidate trajectory.

It is useful to separate two effects:
\[
\begin{array}{ll}
\textbf{algebraic basis drift:}&
  \text{the generalized directions change because the matrix pair changes},\\[2pt]
\textbf{geometric tangent drift:}&
  \text{the directions live at different points of the state manifold}.
\end{array}
\]
The first is the moving-direction issue discussed later in
Section~\ref{sec:moving-direction}; the second precedes it geometrically.
A time-, subset-, or action-dependent source-attribution study must therefore
state how directions are anchored or transported before comparing their
components or assigning persistence to ``mode $i$''.

\begin{researchhypothesis}
A useful research direction for source-resolved collective robotics is not a
new ``Lie-GSVD'' by name. The sharper question is which directional information
and source-attribution statements remain invariant when physically equivalent
tangent directions are transported across time, agents, frames, and planning
hypotheses, and which statements depend on the chosen transport convention.
That question is distinct from the fixed-state left/right invariance proved
above and should be tested against task-level semantics, not only algebraic
convenience.
\end{researchhypothesis}

\begin{collective}
The resulting stack is deliberately modular:
\[
\begin{aligned}
 \text{Lie geometry}
 &\longrightarrow \text{local Jacobians and covariance}
 \longrightarrow \text{Fisher information / whitening},\\
 &\longrightarrow \text{GSVD source attribution}
 \longrightarrow \text{CCA / dependence analysis},\\
 &\longrightarrow \text{planning, decision, and control}.
\end{aligned}
\]
Lie theory tells us how to parameterise local robotic perturbations consistently;
it does not replace statistical modelling. Fisher information quantifies local
information strength; the GSVD resolves its pairwise source balance; CCA addresses
statistical dependence. None of those steps should be asked to do the job of the
next one.
\end{collective}

\section{Jacobians and local linearisation}\label{sec:foundation-jacobians}

The GSVD is a linear-algebraic decomposition, but most robotic sensors are not
linear. A camera projects by division, an IMU contains rotations, and a GNSS
pseudorange contains a Euclidean distance. The bridge from these nonlinear
models to the matrices used later is the \emph{Jacobian}. Section~\ref{sec:foundation-lie}
showed that, when the state contains rotations or poses, this Jacobian is taken
with respect to a specified tangent perturbation rather than by pretending that
the group state itself is a Euclidean vector.

For a purely Euclidean measurement model
\[
   \vekt z = \vekt h(\vekt x)+\vekt n,
\]
where \(\vekt x\in\R^n\) is the state, \(\vekt z\in\R^m\) the measurement,
and \(\vekt n\) measurement error, write around a reference state
\(\bar{\vekt x}\) the perturbation
\(\vekt x=\bar{\vekt x}+\delta\vekt x\). A first-order Taylor expansion gives
\[
   \vekt h(\bar{\vekt x}+\delta\vekt x)
   \approx \vekt h(\bar{\vekt x})+\mat H\,\delta\vekt x,
   \qquad
   \mat H=\left.\frac{\partial\vekt h}{\partial\vekt x}\right|_{\bar{\vekt x}}.
\]
The matrix \(\mat H\in\R^{m\times n}\) is the Jacobian. Its entry
\[
   H_{ij}=\frac{\partial h_i}{\partial x_j}
\]
answers a concrete sensitivity question: \emph{how much does measurement
component \(i\) change when state component \(j\) changes slightly?}

\begin{practicebox}
A Jacobian is not merely a table of derivatives. Its rows are local measurement
constraints and its null space contains state perturbations that are invisible
to the sensor at the current operating point. Consequently, the geometry of
\(\mat H\) changes with viewpoint, trajectory, scene geometry, and linearisation
point. A GSVD of Jacobians is therefore a \emph{local} statement unless the
underlying model is genuinely linear.
\end{practicebox}

\begin{example}[Range sensor: a Jacobian computed by hand]
A robot at \(\vekt x=(x,y)\T\) measures its distance to a beacon at the origin,
\[
   h(x,y)=\sqrt{x^2+y^2}.
\]
The Jacobian is
\[
   \mat H=\begin{bmatrix}
   \dfrac{x}{\sqrt{x^2+y^2}} & \dfrac{y}{\sqrt{x^2+y^2}}
   \end{bmatrix}.
\]
At \((x,y)=(3,4)\), the predicted range is \(5\) m and
\[
   \mat H=\begin{bmatrix}0.6&0.8\end{bmatrix}.
\]
For a small displacement \(\delta\vekt x=(0.10,-0.05)\T\) m, the linear model predicts
\[
   \delta z\approx \mat H\delta\vekt x
   =0.6(0.10)+0.8(-0.05)=0.020\ \text{m}.
\]
The exact new range is
\[
 \sqrt{3.1^2+3.95^2}=5.02021\ \text{m},
\]
so the exact change is \(0.02021\) m. The first-order error is only
\(0.00021\) m for this perturbation.

The null space is the line orthogonal to \((0.6,0.8)\T\), for example
\(\delta\vekt x=(0.8,-0.6)\T\). Indeed,
\[
  \mat H\begin{bmatrix}0.8\\-0.6\end{bmatrix}=0.
\]
Moving tangentially around the beacon therefore produces no first-order range
change. One scalar range observes the radial direction but not the tangential
one.
\end{example}

\begin{example}[Why geometry changes observability]
Add a second range sensor whose local Jacobian is
\[
 \mat H_2=\begin{bmatrix}0&1\end{bmatrix}.
\]
Together,
\[
 \mat H_{12}=\begin{bmatrix}0.6&0.8\\0&1\end{bmatrix},
 \qquad \det(\mat H_{12})=0.6\neq0.
\]
The two measurements now observe both planar directions locally. This is the
simplest example of \emph{collective rank complementarity}: neither scalar
measurement alone determines a two-dimensional displacement, but their stacked
Jacobian does.
\end{example}

\section{Random vectors, covariance, and Mahalanobis distance}\label{sec:foundation-covariance}

A deterministic Jacobian tells us how a state perturbation changes a
measurement. It does not tell us how trustworthy that measurement is. For that
we need a stochastic model.

For a random vector \(\vekt n\) with mean \(\vekt\mu\), the covariance is
\[
 \mat\Sigma=\mathbb E\!\left[(\vekt n-\vekt\mu)(\vekt n-\vekt\mu)\T\right].
\]
Diagonal entries \(\Sigma_{ii}\) are variances; off-diagonal entries
\(\Sigma_{ij}\) are covariances. The correlation coefficient is
\[
 \rho_{ij}=\frac{\Sigma_{ij}}{\sqrt{\Sigma_{ii}\Sigma_{jj}}}.
\]
Covariance therefore contains both scale and linear dependence.

For \(\mat\Sigma\succ0\), the statistically natural squared residual is the
Mahalanobis distance
\[
 d_M^2=\vekt r\T\mat\Sigma^{-1}\vekt r.
\]
A residual of one metre is not intrinsically large or small; it is large or
small relative to its expected uncertainty and correlation structure.

\begin{example}[The same numerical residual can mean something very different]
Suppose \(\vekt r=(0.20,0.20)\T\) m. First take independent errors with
standard deviations \(0.10\) m and \(0.50\) m:
\[
 \mat\Sigma_1=\begin{bmatrix}0.01&0\\0&0.25\end{bmatrix}.
\]
Then
\[
 d_{M,1}^2=\frac{0.20^2}{0.10^2}+\frac{0.20^2}{0.50^2}=4+0.16=4.16.
\]
Now suppose both components have standard deviation \(0.20\) m and correlation
\(\rho=0.9\):
\[
 \mat\Sigma_2=0.04\begin{bmatrix}1&0.9\\0.9&1\end{bmatrix}.
\]
Using
\[
 \mat\Sigma_2^{-1}=\frac{1}{0.04(1-0.9^2)}
 \begin{bmatrix}1&-0.9\\-0.9&1\end{bmatrix},
\]
we obtain \(d_{M,2}^2\approx1.053\). The raw residual is identical, but in the
second model a common positive displacement is expected because the two errors
move together.
\end{example}

\begin{practicebox}
This is why a sensor comparison must not begin with raw Jacobian norms. A row
with large physical units can be noisy and carry little information; a small
row can be precise and carry much more. Whitening converts the residual metric
from physical units to noise-normalised units.
\end{practicebox}

\section{Whitening: putting residuals into noise units}\label{sec:foundation-whitening}

Whitening is one of the most important operations in this text. It is also
one of the easiest to misunderstand.

Let
\[
 \vekt z=\mat H\vekt x+\vekt n,
 \qquad \vekt n\sim\mathcal N(\vekt0,\mat\Sigma),
 \qquad \mat\Sigma\succ0.
\]
Choose a matrix \(\mat L\) satisfying
\[
 \mat L\mat\Sigma\mat L\T=\mat I.
\]
For example, if \(\mat\Sigma=\mat R\mat R\T\) is a Cholesky factorisation,
then \(\mat L=\mat R^{-1}\). Multiplying the complete measurement equation by
\(\mat L\) gives
\[
 \underbrace{\mat L\vekt z}_{\tilde{\vekt z}}
 =\underbrace{\mat L\mat H}_{\mat A}\vekt x
 +\underbrace{\mat L\vekt n}_{\tilde{\vekt n}},
 \qquad
 \operatorname{Cov}(\tilde{\vekt n})=\mat I.
\]
The matrix \(\mat A=\mat L\mat H\) is the \emph{whitened Jacobian}.

\begin{theorem}[Whitening converts Mahalanobis length to Euclidean length]
For \(\tilde{\vekt r}=\mat L\vekt r\) and
\(\mat L\T\mat L=\mat\Sigma^{-1}\),
\[
 \|\tilde{\vekt r}\|^2
 =\vekt r\T\mat\Sigma^{-1}\vekt r.
\]
\end{theorem}
\begin{proof}
Direct substitution gives
\[
 \|\mat L\vekt r\|^2
 =(\mat L\vekt r)\T(\mat L\vekt r)
 =\vekt r\T\mat L\T\mat L\vekt r
 =\vekt r\T\mat\Sigma^{-1}\vekt r.
\]
Thus whitening does not invent a new statistical criterion; it rewrites the
existing Mahalanobis metric as ordinary Euclidean geometry.
\end{proof}
\begin{practicebox}
After whitening, ``one unit'' means one standard-deviation unit in the proper
correlated noise metric. This is why ordinary SVD/GSVD geometry becomes
statistically meaningful after whitening.
\end{practicebox}

\begin{example}[Diagonal whitening: centimetres versus decimetres]
Two measurements of the same scalar state have sensitivities
\[
 \mat H=\begin{bmatrix}1\\1\end{bmatrix},
\]
but standard deviations \(\sigma_1=0.02\) m and \(\sigma_2=0.10\) m. Hence
\[
 \mat\Sigma=\diag(0.0004,0.01),
 \qquad
 \mat L=\diag(50,10),
\]
and
\[
 \mat A=\mat L\mat H=\begin{bmatrix}50\\10\end{bmatrix}.
\]
The first measurement contributes \(50^2=2500\) units of scalar information;
the second contributes \(10^2=100\). Their relative shares are
\[
 \frac{2500}{2600}=96.15\%,\qquad
 \frac{100}{2600}=3.85\%.
\]
The two raw Jacobian rows were identical. Their information was not.
\end{example}

\begin{example}[Correlated whitening rotates as well as scales]
Take
\[
 \mat\Sigma=\begin{bmatrix}4&1.2\\1.2&1\end{bmatrix}.
\]
A Cholesky factor is approximately
\[
 \mat R=\begin{bmatrix}2&0\\0.6&0.8\end{bmatrix},
 \qquad
 \mat L=\mat R^{-1}
 =\begin{bmatrix}0.5&0\\-0.375&1.25\end{bmatrix}.
\]
For \(\mat H=\mat I\), whitening gives
\[
 \mat A=\begin{bmatrix}0.5&0\\-0.375&1.25\end{bmatrix}.
\]
Whitening has not merely divided each row by a standard deviation: the second
whitened residual subtracts part of the first because the original errors were
correlated. This is why diagonal ``sigma scaling'' is wrong when off-diagonal
covariances matter.
\end{example}

\begin{figure}[htbp]
\centering
\begin{tikzpicture}[figbase,scale=1.0]
 \begin{scope}
  \draw[ax] (-2.95,0)--(3.05,0) node[right,tick]{$r_1$};
  \draw[ax] (0,-2.35)--(0,2.45) node[above,tick]{$r_2$};
  \draw[rulegray!170,thincrv] (0,0) circle (1);
  \draw[srcA,crv,rotate=19.33] (0,0) ellipse (2.1026 and 0.7610);
  \foreach \x/\y in {0.2519/-0.7181,-1.9841/-0.6960,-0.2519/0.7181,1.9841/0.6960}
     \fill[srcA] (\x,\y) circle (1.7pt);
  \node[srcA] at (0,-3.05) {$\{\vekt r:\vekt r\T\bm\Sigma^{-1}\vekt r=1\}$};
  \node[struct!60] at (0,-3.6) {before whitening};
 \end{scope}
 \draw[flow,pooled] (3.55,0)--(5.35,0);
 \node[pooled,above] at (4.45,0.12) {$\mat L=\mat R^{-1}$};
 \node[pooled,below,font=\scriptsize] at (4.45,-0.12) {$\bm\Sigma=\mat R\mat R\T$};
 \begin{scope}[xshift=8.9cm]
  \draw[ax] (-2.95,0)--(3.05,0) node[right,tick]{$\tilde r_1$};
  \draw[ax] (0,-2.35)--(0,2.45) node[above,tick]{$\tilde r_2$};
  \draw[pooled,crv] (0,0) circle (1);
  \foreach \x/\y in {0.1259/-0.9920,-0.9920/-0.1259,-0.1259/0.9920,0.9920/0.1259}
     \fill[pooled] (\x,\y) circle (1.7pt);
  \node[pooled] at (0,-3.05) {$\{\tilde{\vekt r}:\|\tilde{\vekt r}\|=1\}$};
  \node[struct!60] at (0,-3.6) {after whitening};
 \end{scope}
 \node[fcap] at (4.45,-4.5)
   {the four marked points are the \emph{same} residuals in both frames;\\
    correlation makes $\mat L$ rotate, not only rescale};
\end{tikzpicture}
\caption{\textbf{Whitening rewrites the Mahalanobis metric as ordinary Euclidean geometry.} Left: the $1\sigma$ ellipse of a correlated noise covariance. Right: after multiplication by $\mat L=\mat R^{-1}$ the \emph{same} four residuals lie on the unit circle, so that ``one unit'' now means one standard deviation in the correct correlated metric. The rotation between the two panels is the essential point: with correlated noise, whitening cannot be reduced to dividing each row by its own standard deviation. Every information share reported later inherits the accuracy of the $\bm\Sigma$ used here.}
\label{fig:whitening-geometry}
\end{figure}
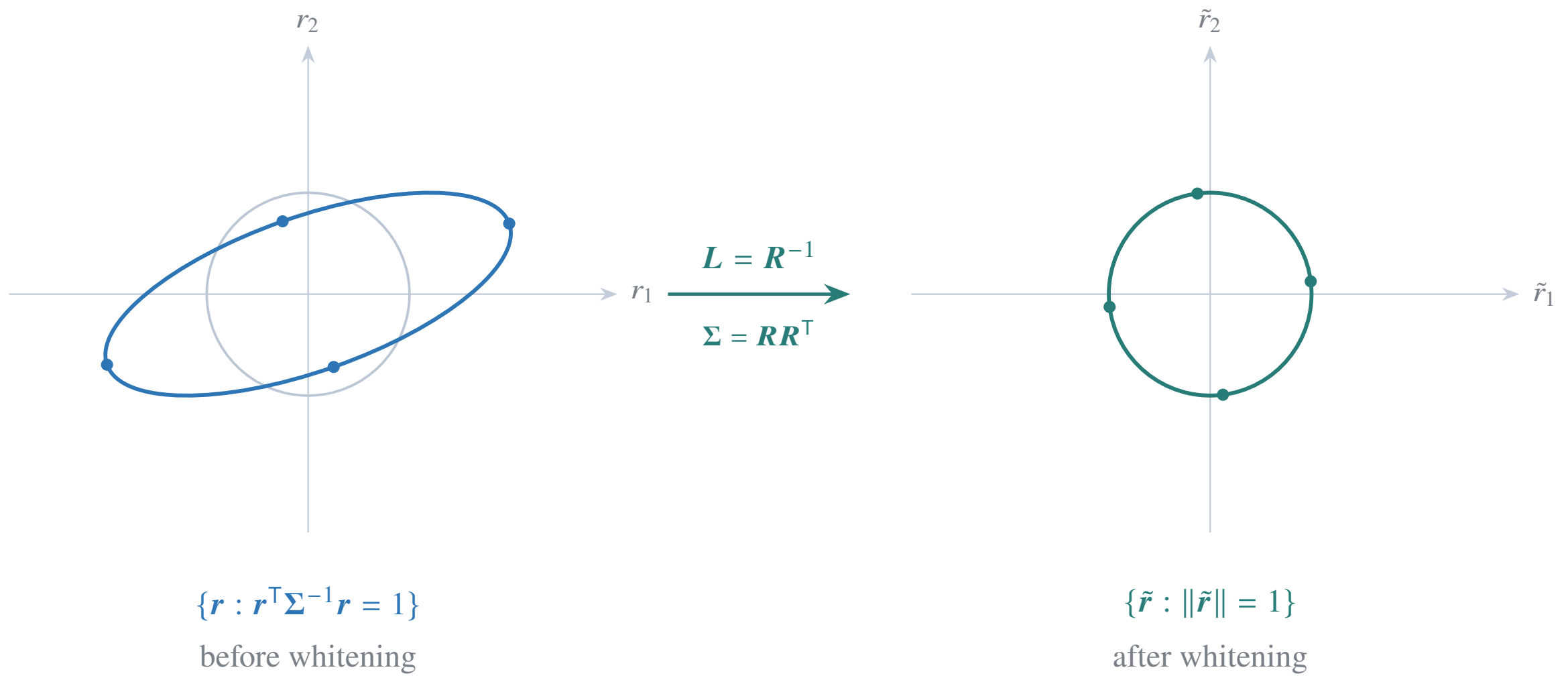

Figure~\ref{fig:whitening-geometry} shows this geometrically: four residuals on
the $1\sigma$ ellipse of the correlated covariance, and the same four on the unit
circle after whitening. The rotation between the two panels is precisely what
row-wise sigma scaling cannot produce.

\FloatBarrier

\begin{remark}[What whitening does not do]
Whitening makes the \emph{noise covariance} equal to \(\mat I\). It does not
make the columns of \(\mat A\) orthogonal, it does not equalise sensor
sensitivities, and it does not guarantee that two sensor blocks have comparable
norms. Geometry remains geometry; whitening only supplies the statistically
correct output metric.
\end{remark}

\section{Fisher information from a whitened Jacobian}\label{sec:foundation-fisher}

For the linear Gaussian model above, the negative log likelihood, up to a
constant, is
\[
 \ell(\vekt x)=\half\|\tilde{\vekt z}-\mat A\vekt x\|^2.
\]
Its Hessian is
\[
 \nabla^2\ell(\vekt x)=\mat A\T\mat A
 =\mat H\T\mat\Sigma^{-1}\mat H.
\]
This matrix is the Fisher information matrix,
\[
 \FisherMat=\mat H\T\mat\Sigma^{-1}\mat H=\mat A\T\mat A.
\]
For any state direction \(\vekt d\),
\[
 \vekt d\T\FisherMat\vekt d=\|\mat A\vekt d\|^2.
\]
Thus Fisher information has a direct geometric meaning: it is the squared
noise-normalised measurement response generated by a unit state perturbation.

\begin{example}[Directional information]
Let
\[
 \mat A=\begin{bmatrix}2&0\\0&0.5\end{bmatrix},
 \qquad
 \FisherMat=\begin{bmatrix}4&0\\0&0.25\end{bmatrix}.
\]
Along \(\vekt e_1\), the information is \(4\); along \(\vekt e_2\), it is
\(0.25\). The first direction is therefore sixteen times stronger in Fisher
information. Since standard deviation scales as the inverse square root of
information, the corresponding idealised standard-deviation ratio is
\(\sqrt{4/0.25}=4\), not sixteen.
\end{example}

\begin{collective}
For independent whitened agents with Jacobians \(\mat A\) and \(\mat B\),
information adds:
\[
 \mat J=\mat A\T\mat A+\mat B\T\mat B.
\]
This additive pooled matrix answers ``how strongly does the collective constrain
each state direction?'' The GSVD answers the different question ``how is that
pooled directional strength divided between the two sources?''
\end{collective}

\section{Linear Gaussian estimation and posterior covariance}\label{sec:foundation-gaussian-estimation}

The same matrices reappear in estimation. With a Gaussian prior
\[
 \vekt x\sim\mathcal N(\vekt\mu_0,\mat P_0)
\]
and a linear Gaussian measurement, the posterior precision is
\[
 \mat P_+^{-1}=\mat P_0^{-1}+\mat H\T\mat\Sigma^{-1}\mat H.
\]
Without a prior, and assuming full column rank, the efficient least-squares/ML
covariance is
\[
 \mat P=(\mat H\T\mat\Sigma^{-1}\mat H)^{-1}.
\]
This explains why information matrices are convenient: independent evidence
adds in precision space, while uncertainty itself is represented by the inverse.

\begin{example}[One prior, one measurement]
Suppose a scalar position has prior standard deviation \(2\) m, so
\(p_0=4\) and prior information \(1/p_0=0.25\). A GNSS measurement has standard
deviation \(1\) m and sensitivity one, so its information is \(1\). Therefore
\[
 p_+^{-1}=0.25+1=1.25,
 \qquad
 p_+=0.8,
 \qquad
 \sigma_+=\sqrt{0.8}=0.894\ \text{m}.
\]
Information added linearly; standard deviations did not.
\end{example}

\section{Canonical correlation analysis: dependence between two random vectors}\label{sec:foundation-cca}

GSVD and CCA both use whitening and an SVD, which makes them easy to confuse.
Their questions are different. GSVD compares two \emph{operators acting on the
same state}. CCA compares two \emph{random vectors observed in paired samples}
and asks which linear combinations co-vary most strongly.

Let zero-mean random vectors \(\vekt a\in\R^p\) and \(\vekt b\in\R^q\) have
covariance blocks
\[
 \mat\Sigma_{aa}=\mathbb E[\vekt a\vekt a\T],\quad
 \mat\Sigma_{bb}=\mathbb E[\vekt b\vekt b\T],\quad
 \mat\Sigma_{ab}=\mathbb E[\vekt a\vekt b\T].
\]
CCA searches for vectors \(\vekt u,\vekt v\) maximising
\[
 \rho=\operatorname{corr}(\vekt u\T\vekt a,\vekt v\T\vekt b).
\]
After whitening both sides, define
\[
 \mat K=\mat\Sigma_{aa}^{-1/2}\mat\Sigma_{ab}
        \mat\Sigma_{bb}^{-1/2}.
\]
If
\[
 \mat K=\mat U\diag(\rho_1,\ldots)\mat V\T,
\]
then the singular values \(\rho_i\in[0,1]\) are the canonical correlations.

\begin{practicebox}
CCA is a dependence diagnostic. A canonical correlation near one means that a
linear combination of the first random vector is strongly predictable from a
linear combination of the second. It says nothing by itself about which sensor
provides more Fisher information about a state. Conversely, a GSVD dominance
ratio says nothing by itself about whether the realised errors of two agents
are correlated.
\end{practicebox}

\begin{example}[CCA by hand: hidden dependence that components obscure]
Let
\[
 \mat\Sigma_{aa}=\mat I,\qquad
 \mat\Sigma_{bb}=\mat I,\qquad
 \mat\Sigma_{ab}=\begin{bmatrix}0.6&0.6\\0.6&0.6\end{bmatrix}.
\]
The individual cross-correlations are only \(0.6\), but
\[
 \mat K=\mat\Sigma_{ab}.
\]
Its singular values are \(1.2\) and \(0\), which cannot be a valid correlation
matrix because a canonical correlation may not exceed one. This immediately
reveals that the proposed covariance blocks are not jointly positive
semidefinite. CCA therefore also provides a useful consistency check.

Now replace \(0.6\) by \(0.4\):
\[
 \mat K=\begin{bmatrix}0.4&0.4\\0.4&0.4\end{bmatrix}.
\]
Its singular values are \(0.8\) and \(0\). The first canonical directions are
\[
 \vekt u_1=\vekt v_1=\frac{1}{\sqrt2}(1,1)\T.
\]
Hence the sums \((a_1+a_2)/\sqrt2\) and \((b_1+b_2)/\sqrt2\) correlate at
\(0.8\), while the orthogonal difference directions are uncorrelated. CCA has
found the coupled mode that componentwise inspection hides.
\end{example}

\begin{example}[Nontrivial whitening in CCA]
Take
\[
 \mat\Sigma_{aa}=\diag(4,1),\qquad
 \mat\Sigma_{bb}=\diag(1,9),\qquad
 \mat\Sigma_{ab}=\begin{bmatrix}1&1.2\\0&1.5\end{bmatrix}.
\]
The inverse square roots are
\[
 \mat\Sigma_{aa}^{-1/2}=\diag(1/2,1),\qquad
 \mat\Sigma_{bb}^{-1/2}=\diag(1,1/3),
\]
so
\[
 \mat K=
 \begin{bmatrix}0.5&0.2\\0&0.5\end{bmatrix}.
\]
The eigenvalues of \(\mat K\T\mat K\) are approximately \(0.3719\) and
\(0.1681\), giving canonical correlations
\[
 \rho_1\approx0.610,
 \qquad
 \rho_2\approx0.410.
\]
The raw covariance entry \(1.5\) was the largest number in
\(\mat\Sigma_{ab}\), but after accounting for marginal scales the strongest
canonical correlation is only about \(0.61\). This is exactly why CCA whitens
before comparing cross-covariance.
\end{example}

\section{Mutual information and Gaussian log-determinants}\label{sec:foundation-mi}

Fisher information is local sensitivity of a likelihood. Mutual information
answers another question: how much uncertainty about one random variable is
removed by observing another?
\[
 I(\vekt x;\vekt z)=h(\vekt x)-h(\vekt x\mid\vekt z).
\]
For a Gaussian \(n\)-vector with covariance \(\mat P\),
\[
 h(\vekt x)=\half\log\!\big((2\pi e)^n\det\mat P\big).
\]
Therefore, for a linear Gaussian measurement with prior \(\mat P_0\),
\[
 I(\vekt x;\vekt z)
 =\half\log\det\!\left(\mat I+\mat P_0\mat H\T
 \mat\Sigma^{-1}\mat H\right).
\]
This log-determinant formula will later explain why information gain is not the
same object as a GSVD ratio.

\begin{example}[Scalar mutual information]
Let \(x\sim\mathcal N(0,4)\) and \(z=x+n\), with
\(n\sim\mathcal N(0,1)\). Then
\[
 I(x;z)=\half\log(1+4)=\half\log5=0.8047\ \text{nats}.
\]
If the measurement noise variance is reduced to \(0.25\), the information is
\[
 \half\log(1+4/0.25)=\half\log17=1.4166\ \text{nats}.
\]
Four times the Fisher information does not produce four times the mutual
information because the latter grows logarithmically.
\end{example}

\section{KL, Bhattacharyya, and Chernoff: discrimination is a different task}\label{sec:foundation-discrimination}

Later we will ask whether an observation can distinguish hypotheses. That is a
decision problem, not a source-attribution problem. For two distributions
\(p\) and \(q\), the Kullback--Leibler divergence is
\[
 d_{\mathrm{KL}}(p\|q)=\mathbb E_p\!\left[\log\frac{p}{q}\right].
\]
For equal-covariance Gaussians
\(\mathcal N(\vekt\mu_0,\mat\Sigma)\) and
\(\mathcal N(\vekt\mu_1,\mat\Sigma)\),
\[
 d_{\mathrm{KL}}
 =\half(\vekt\mu_1-\vekt\mu_0)\T\mat\Sigma^{-1}
       (\vekt\mu_1-\vekt\mu_0).
\]
Thus discrimination depends on the hypothesis separation \emph{and} the noise
metric. Bhattacharyya distance and Chernoff information provide related
symmetric/error-exponent viewpoints and become important when covariances
also differ.

\begin{example}[A large sensor response need not discriminate the hypotheses]
Suppose two hypotheses differ by
\(\Delta\vekt\mu=(1,0)\T\). With
\(\mat\Sigma=\diag(100,1)\),
\[
 d_{\mathrm{KL}}=\half\frac{1^2}{100}=0.005.
\]
A one-unit mean separation sounds large, but it lies in a direction with
standard deviation \(10\), so the hypotheses are barely distinguishable. If
instead \(\Delta\vekt\mu=(0,1)\T\), then \(d_{\mathrm{KL}}=0.5\): the same
Euclidean separation is one hundred times stronger in KL divergence.
\end{example}

\section{Covariance intersection: conservative fusion when dependence is unknown}\label{sec:foundation-ci}

If two estimates \((\hat{\vekt x}_A,\mat P_A)\) and
\((\hat{\vekt x}_B,\mat P_B)\) have unknown cross-correlation, adding their
information matrices as if they were independent can become overconfident.
Covariance intersection (CI) avoids that assumption by choosing
\(\omega\in[0,1]\) and forming
\[
 \mat P_{\mathrm{CI}}^{-1}
 =\omega\mat P_A^{-1}+(1-\omega)\mat P_B^{-1},
\]
\[
 \hat{\vekt x}_{\mathrm{CI}}
 =\mat P_{\mathrm{CI}}
 \left[\omega\mat P_A^{-1}\hat{\vekt x}_A
 +(1-\omega)\mat P_B^{-1}\hat{\vekt x}_B\right].
\]
The weight is selected by a criterion such as minimising
\(\trace(\mat P_{\mathrm{CI}})\) or \(\log\det\mat P_{\mathrm{CI}}\).

\begin{example}[Why unknown dependence matters]
Two scalar estimates both report variance \(1\). If they were independent,
ordinary information fusion would give variance
\[
 (1+1)^{-1}=0.5.
\]
If they were in fact perfectly correlated copies of the same error, variance
\(0.5\) would be unjustified: no independent information was added. CI with
\(\omega=0.5\) gives
\[
 p_{\mathrm{CI}}^{-1}=0.5(1)+0.5(1)=1,
 \qquad p_{\mathrm{CI}}=1,
\]
which correctly refuses to claim a precision gain without knowledge of the
cross-correlation.
\end{example}

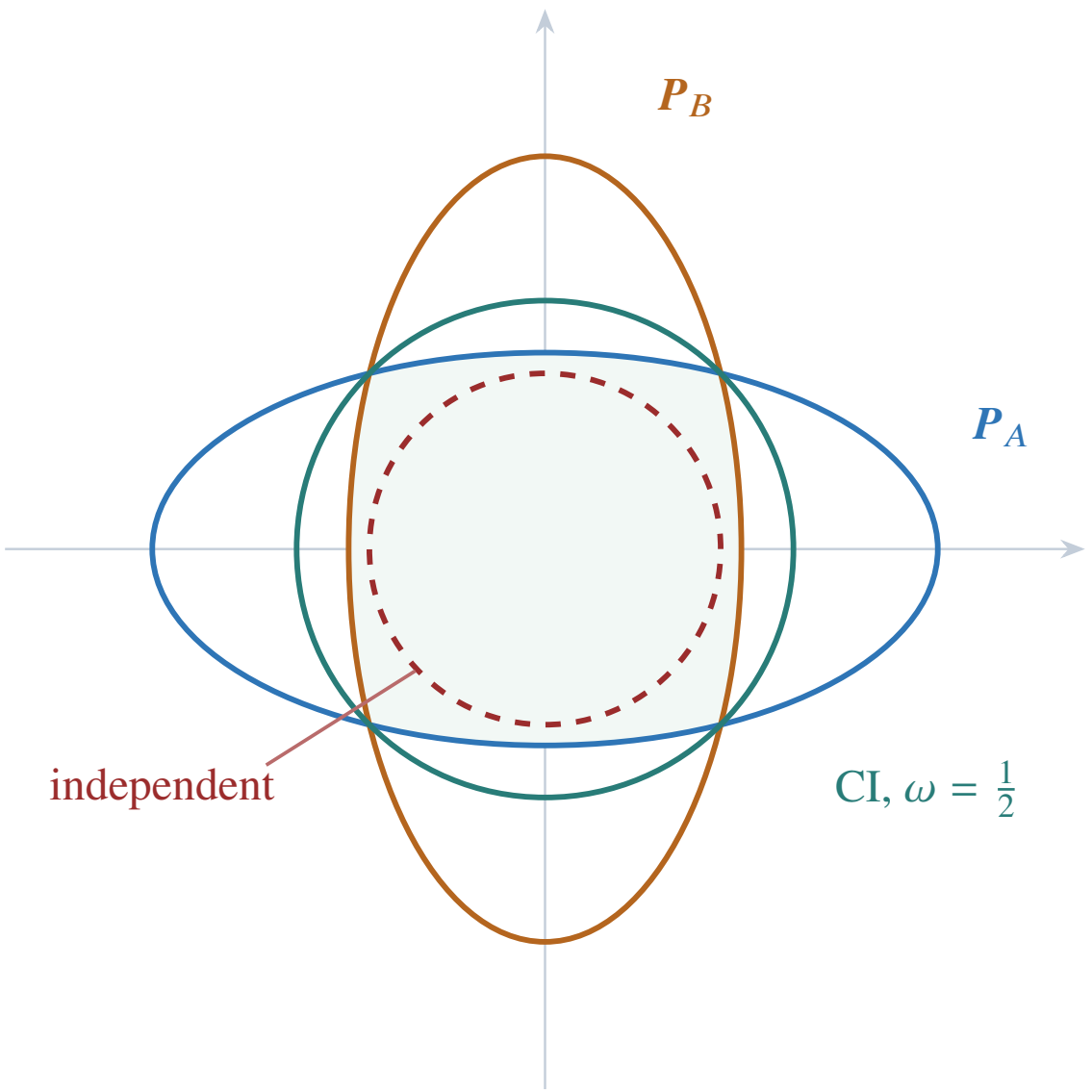
\begin{figure}[htbp]
\centering
\begin{tikzpicture}[figbase,scale=1.35]
  \draw[ax] (-2.75,0)--(2.75,0); \draw[ax] (0,-2.75)--(0,2.75);
  \begin{scope}
    \clip (0,0) ellipse (2 and 1);
    \fill[softgreen] (0,0) ellipse (1 and 2);
  \end{scope}
  \draw[srcA,crv] (0,0) ellipse (2 and 1);
  \draw[srcB,crv] (0,0) ellipse (1 and 2);
  \draw[pooled,crv] (0,0) circle (1.2649);
  \draw[warn,crv,dashed] (0,0) circle (0.8944);
  \node[lbl,text=srcA] at (2.32,0.62) {$\mat P_A$};
  \node[lbl,text=srcB] at (0.72,2.30) {$\mat P_B$};
  \node[lbl,text=pooled] at (1.95,-1.22) {CI, $\omega=\tfrac12$};
  \node[lbl,text=warn] at (-1.95,-1.22) {independent};
  \draw[warn!70,thincrv] (-1.42,-1.10)--(-0.66,-0.62);
  \node[fcap,text width=10.6cm] at (0,-3.35)
    {for this symmetric example, CI remains deliberately broader than the naive
     independence fusion.\\
     Its consistency comes from the covariance-intersection construction, not from
     treating the geometric overlap of the two ellipses as a fused covariance.};
\end{tikzpicture}
\caption{\textbf{Covariance intersection buys consistency with conservatism.} For the symmetric two-dimensional example shown here, the CI covariance is deliberately broader than the covariance obtained by assuming independence. CI is designed to remain consistent without knowing the cross-correlation; that guarantee comes from the covariance-intersection construction, not from identifying the fused covariance with the geometric intersection of the two marginal ellipses. Assuming independence instead yields the smaller dashed circle, which is justified only when the required independence model is valid. The gap between the two circles is the price of not knowing the correlation --- and the reason that measuring it, rather than bounding it, is worth something.}
\label{fig:covariance-intersection}
\end{figure}

Figure~\ref{fig:covariance-intersection} shows the two-dimensional form of the
same refusal, drawn to scale.

\FloatBarrier

\begin{collective}
This establishes the division of labour used later:
\begin{center}
\begin{tabular}{@{}ll@{}}
\toprule
Question & Primary tool\\
\midrule
How does a nonlinear sensor react locally? & Jacobian\\
How should residuals be scaled statistically? & Covariance / whitening\\
How strongly is a state direction constrained? & Fisher information\\
Which source carries which state direction? & GSVD\\
How are realised source errors linearly dependent? & CCA / cross-covariance\\
How much uncertainty is removed? & Mutual information\\
How distinguishable are hypotheses? & KL / Bhattacharyya / Chernoff\\
How can estimates be fused with unknown dependence? & Covariance intersection\\
\bottomrule
\end{tabular}
\end{center}
The tools are complementary. Replacing one by another usually changes the
question rather than answering it.
\end{collective}

\section{Generalized eigenvalues: two quadratic forms at once}
\label{sec:geneig}

Now the bridge to the GSVD.

\begin{definition}
Given symmetric $\mat{S}$ and symmetric positive definite $\mat{T}$, the \emph{generalized
eigenvalue problem} is
\[
   \mat{S}\vekt{x}=\lambda \mat{T} \vekt{x},\qquad \vekt{x}\neq\vekt{0}.
\]
The pair $(\mat{S},\mat{T})$ is called a \emph{matrix pencil}.
\end{definition}
\begin{practicebox}
A generalized eigenproblem compares one quadratic energy with another rather than with the Euclidean norm. This is exactly the situation when two sensing modalities constrain the same unknown with different directional strengths.
\end{practicebox}

\begin{theorem}[Simultaneous diagonalisation]\label{thm:simdiag}
If $\mat{T}$ is symmetric positive definite and $\mat{S}$ is symmetric, there exists an
invertible $\mat{X}$ such that
\[
   \mat{X}\T \mat{S}\mat{X}=\mat{\Lambda},\qquad \mat{X}\T \mat{T}\mat{X}=\mat{I},
\]
with $\mat{\Lambda}$ diagonal.
\end{theorem}
\begin{practicebox}
This theorem says that two quadratic forms can be made scalar-by-scalar comparable in one common coordinate system when the reference form is positive definite. It is the algebraic heart of relative information attribution.
\end{practicebox}

\begin{proof}[Proof --- and note that it is a recipe]
\textbf{Step 1: whiten $\mat{T}$.} Since $\mat{T}$ is positive definite, write
$\mat{T}=\mat{L}\mat{L}\T$ by Cholesky factorisation. Set
$\widetilde{\mat{S}}=\mat{L}^{-1}\mat{S}\mat{L}^{-\mathsf T}$. This matrix is symmetric.

\textbf{Step 2: diagonalise the whitened form.} By the spectral theorem,
$\widetilde{\mat{S}}=\mat{Q}\mat{\Lambda} \mat{Q}\T$ with $\mat{Q}$ orthogonal.

\textbf{Step 3: transform back.} Put $\mat{X}=\mat{L}^{-\mathsf T}\mat{Q}$. Then
\[
 \mat{X}\T \mat{T}\mat{X}=\mat{Q}\T \mat{L}^{-1}(\mat{L}\mat{L}\T)\mat{L}^{-\mathsf T}\mat{Q}=\mat{I},
\]
and
\[
 \mat{X}\T \mat{S}\mat{X}=\mat{Q}\T \mat{L}^{-1}\mat{S}\mat{L}^{-\mathsf T}\mat{Q}=\mat{\Lambda}.
\]
\end{proof}

\begin{remark}[Congruence, not similarity]
$\mat{X}\T \mat{S}\mat{X}$ is a \emph{congruence} transform, not a similarity transform
$\mat{X}^{-1}\mat{S}\mat{X}$. This distinction matters because quadratic forms transform by
congruence:
\[
   \vekt{x}\T \mat{S}\vekt{x}=\vekt{c}\T(\mat{X}\T \mat{S}\mat{X})\vekt{c}\qquad\text{when }\vekt{x}=\mat{X}\vekt{c}.
\]
The GSVD is fundamentally about comparing quadratic forms, not about applying
one linear map repeatedly.
\end{remark}
\begin{practicebox}
Covariances, energies, and information matrices transform by congruence under a change of coordinates. Keeping this distinct from similarity prevents incorrect claims about which numbers are coordinate-invariant.
\end{practicebox}

\begin{corollary}[Generalized Rayleigh quotient]\label{cor:genray}
Let $\mat{S}=\mat{A}\T \mat{A}$ and $\mat{T}=\mat{B}\T \mat{B}$, with $\mat{B}$ of full column rank. Then
\[
   \max_{\vekt{x}\neq\vekt{0}}\frac{\|\mat{A}\vekt{x}\|^2}{\|\mat{B}\vekt{x}\|^2}
   =\lambda_{\max}(\mat{A}\T \mat{A},\mat{B}\T \mat{B}),
\]
and the maximum is attained at a corresponding generalized eigenvector.
\end{corollary}
\begin{practicebox}
The largest generalized singular value is the strongest possible $\mat{A}$-over-$\mat{B}$ response ratio. It is a relative sensitivity measure; it should not be mistaken for a statistical hypothesis-separation metric.
\end{practicebox}

\begin{proof}
Use $\vekt{x}=\mat{X}\vekt{c}$ from Theorem~\ref{thm:simdiag}. Then the quotient becomes
\[
  \frac{\sum_i\lambda_i c_i^2}{\sum_i c_i^2},
\]
a weighted average of the generalized eigenvalues. Its maximum is therefore
$\lambda_{\max}$.
\end{proof}

This characterisation is mathematically useful but numerically dangerous if it
is implemented literally: forming $\mat{A}\T \mat{A}$ and $\mat{B}\T \mat{B}$ squares condition numbers.
The GSVD gives the same spectral comparison while working with $\mat{A}$ and $\mat{B}$
directly \cite{VanLoan1976,PaigeSaunders1981,GolubVanLoan2013}.

\begin{collective}
Two quadratic forms admit an exact common diagonal coordinate system under the
hypotheses above. For three or more generic forms, exact simultaneous
diagonalisation by one congruence is no longer available in general. That does
\emph{not} force a single replacement method. Depending on the scientific
question, sensible choices include pairwise analysis, one-agent-versus-rest,
approximate joint diagonalisation, or a higher-order GSVD. Higher-order GSVDs
do exist, but their structure and uniqueness are weaker than in the two-matrix
case \cite{Ponnapalli2011}.
\end{collective}

\section{Principal angles between subspaces}\label{sec:angles}

Principal angles solve a different problem: they compare \emph{subspaces in the
same ambient Euclidean space}. They are useful in this text, but they must not
be confused with the GSVD mixing angles introduced later.

\begin{definition}[Principal angles]
Let $\mathcal S,\mathcal T\subseteq\R^d$ be subspaces of dimensions $s\ge t$.
The principal angles
$0\le\theta_1\le\cdots\le\theta_t\le\pi/2$ are defined recursively by
\[
\cos\theta_k=\max_{\vekt{u}\in\mathcal S,\,\vekt{v}\in\mathcal T}\vekt{u}\T \vekt{v}
\]
subject to $\|\vekt{u}\|=\|\vekt{v}\|=1$ and orthogonality to all previously chosen principal
vectors.
\end{definition}
\begin{practicebox}
Principal angles provide a coordinate-free measure of how much two subspaces overlap. For sensing, they answer whether two agents constrain the same parts of the state space, independent of the bases used to represent those parts.
\end{practicebox}

\begin{theorem}[Bj\"orck--Golub characterisation]\label{thm:angles}
Let $\mat{Q}_{\mathcal S}$ and $\mat{Q}_{\mathcal T}$ have orthonormal columns spanning
$\mathcal S$ and $\mathcal T$. Then
\[
   \cos\theta_i=\sigma_i(\mat{Q}_{\mathcal S}\T \mat{Q}_{\mathcal T}).
\]
\end{theorem}
\begin{practicebox}
This result turns a geometric subspace question into an ordinary SVD of an overlap matrix. That makes principal angles both numerically practical and easy to validate in software.
\end{practicebox}

\begin{proof}[Sketch]
Write unit vectors as $\vekt{u}=\mat{Q}_{\mathcal S}\vekt{a}$ and $\vekt{v}=\mat{Q}_{\mathcal T}\vekt{b}$. Then
$\vekt{u}\T \vekt{v}=\vekt{a}\T(\mat{Q}_{\mathcal S}\T \mat{Q}_{\mathcal T})\vekt{b}$. Maximising this bilinear form
with successive orthogonality constraints is exactly the variational
characterisation of singular values. This classical numerical construction is
due to Bj\"orck and Golub \cite{BjorckGolub1973}.
\end{proof}

\begin{example}[Two observable subspaces]\label{ex:angles}
In $\R^3$, let
\[
\mathcal S=\spn\{\vekt{e}_1,\vekt{e}_2\},\qquad
\mathcal T=\spn\left\{\vekt{e}_1,\left(0,\half,\tfrac{\sqrt3}{2}\right)\T\right\}.
\]
Both displayed bases are orthonormal, so
\[
\mat{Q}_{\mathcal S}\T \mat{Q}_{\mathcal T}
=\begin{pmatrix}1&0\\0&\half\end{pmatrix}.
\]
Its singular values are $1$ and $1/2$. Hence the principal angles are
$0^\circ$ and $60^\circ$.
\end{example}
\begin{practicebox}
A zero principal angle identifies an exactly shared direction; a larger angle identifies increasing geometric separation. This diagnoses redundancy or complementarity of observable subspaces, not dependence of random errors.
\end{practicebox}

\begin{remark}[Geometry is not dependence]\label{rem:angles-not-independence}
Principal angles say how subspaces are oriented. They do \textbf{not} test
statistical independence.

For example, let two independent random errors have the same anisotropic
covariance,
\[
 \vekt{e}_{\mathcal A},\vekt{e}_{\mathcal B}\ \text{independent},\qquad
 \operatorname{Cov}(\vekt{e}_{\mathcal A})=
 \operatorname{Cov}(\vekt{e}_{\mathcal B})=\diag(100,1).
\]
Their dominant one-dimensional error subspaces are identical, so their
principal angle is $0^\circ$. Nevertheless their cross-covariance is zero and,
if they are jointly Gaussian, they are independent. The converse can fail as
well: two strongly dependent errors can have orthogonal marginal dominant
subspaces.

Therefore principal angles are excellent for \emph{subspace geometry}; paired
cross-covariance or canonical correlation is the right linear tool for
\emph{cross-agent statistical dependence}. We will make that distinction
explicit in Section~\ref{sec:commonmode}.
\end{remark}
\begin{practicebox}
Two sensors may have similarly oriented error ellipsoids and still be statistically independent, or have orthogonal marginal error directions while being strongly coupled. Dependence must be tested from paired cross-statistics, not from marginal subspace geometry alone.
\end{practicebox}

\begin{figure}[!htbp]
\centering
\begin{tikzpicture}[figbase,scale=0.86]
 \begin{scope}
  \draw[ax] (-3.9,0)--(3.9,0); \draw[ax] (0,-3.9)--(0,3.9);
  \draw[srcA,crv] (0,0) ellipse (2 and 1);
  \draw[srcB,crv] (0,0) ellipse (1 and 3);
  \draw[vec,srcA] (0,0)--(2,0);
  \draw[vec,srcB] (0,0)--(0,3);
  \node[lbl,text=srcA] at (2.55,-0.42) {$\bm\Sigma_{AA}$};
  \node[lbl,text=srcB] at (0.92,3.35) {$\bm\Sigma_{BB}$};
  \draw[warn,thincrv] (0.8,0) arc (0:90:0.8);
  \node[lbl,text=warn,font=\scriptsize] at (1.42,1.30) {$\theta_1=90^\circ$};
  \node[fcap] at (0,-4.75)
    {dominant error directions\\ are exactly perpendicular};
 \end{scope}
 \begin{scope}[xshift=9.6cm]
  \draw[ax] (-3.9,0)--(3.9,0) node[right,tick]{$e_{\mathcal A,1}$};
  \draw[ax] (0,-3.9)--(0,3.9) node[above,tick]{$e_{\mathcal B,1}$};
  \begin{scope}
   \clip (-3.7,-3.7) rectangle (3.7,3.7);
   \foreach \x/\y in {-0.05/0.11,0.70/-0.06,1.07/0.07,-0.35/0.42,1.09/0.24,-1.04/-0.33,
   -0.05/-0.43,-0.06/0.28,2.76/1.09,4.01/1.31,3.71/1.63,2.48/1.28,-0.28/-0.21,5.11/2.16,
   0.08/0.09,3.13/1.26,2.09/0.62,-1.98/-1.28,-0.08/0.35,1.18/0.50,-0.23/-0.08,2.43/1.17,
   -2.45/-1.82,-1.73/-0.76,0.94/1.32,-1.32/-1.14,-0.25/0.14,0.26/0.42,0.02/0.30,-2.75/-1.59,
   -0.33/-0.36,-0.05/-0.54,1.19/0.47,-1.98/-0.43,2.77/0.95,-0.95/-1.32,0.94/0.40,-2.62/-0.93,
   0.71/0.17,0.24/0.78,0.90/0.29,-0.68/-0.37,0.58/-0.22,0.10/-0.15,-2.43/-0.85,-0.06/0.26,
   0.50/0.69,-0.09/0.21,2.51/1.33,3.71/0.85,0.76/-0.04,-0.70/0.65,1.76/0.55,-0.49/-0.01,
   0.39/0.09,-1.49/-0.47,2.07/0.94,0.11/-0.41,-0.37/-0.55,-1.97/-0.84,-0.08/-0.30,0.57/-0.61,
   2.19/1.19,4.10/2.29,3.35/1.90,1.55/1.07,-0.00/-0.67,-2.73/-0.65,0.18/-0.04,0.48/-0.20,
   -2.10/-1.22,0.24/-0.24,1.46/0.13,-2.48/-1.23,-1.93/-0.90,1.44/0.53,1.12/0.53,0.80/0.24,
   2.88/1.00,-3.18/-1.78,2.04/1.10,-2.56/-1.84,0.52/-0.03,3.39/1.91,0.03/0.05,1.42/0.87,
   1.10/0.45,2.41/0.60,-2.58/-1.43,-0.58/-0.28,0.96/0.23,-1.20/-0.70,0.30/0.15,-2.46/-0.85,
   -0.67/-0.56,2.59/1.63,-1.90/-0.95,-1.21/-0.23,-1.81/-0.45,0.65/0.97,2.34/1.47,-1.13/-0.15,
   -3.99/-1.22,2.57/0.46,-1.46/-1.13,-0.12/0.31,3.63/0.78,-0.52/-0.23,0.48/0.24,1.97/0.26,
   0.50/-0.19,3.19/1.72,0.05/0.20,2.08/0.69,2.14/0.61,-0.26/-0.46,-0.77/-0.21,-3.80/-2.39,
   -1.75/-0.86,0.27/-0.51,0.79/0.70,0.15/0.11,0.39/0.69,0.41/-0.77,1.71/-0.06,6.57/2.84,
   -2.67/-1.23,2.49/0.76,-2.28/-1.00,-0.09/-0.06,-0.21/0.25,-1.14/-0.44,1.99/1.16,1.18/1.03,
   2.56/1.14,0.24/-0.47,-3.68/-1.08,0.79/0.71,-0.32/-1.29,-0.49/-0.26,0.02/-0.46,0.57/0.19,
   -2.42/-0.99,-0.25/0.55,1.17/0.38,3.36/2.26,-2.08/-0.57,-1.35/-0.58,-0.39/-0.29,0.40/0.21,
   -3.11/-1.21,0.21/-0.15,1.00/1.17,-1.02/-0.45,0.88/-0.07,-0.01/0.38,0.82/0.28,0.42/0.25,
   3.21/1.40,1.55/1.11}
     \fill[warn,opacity=0.5] (\x,\y) circle (1.5pt);
   \draw[warn,dashed,crv] (-3.7,-1.665)--(3.7,1.665);
  \end{scope}
  \node[lbl,text=warn] at (2.45,-2.75) {$\rho_1=0.9$};
  \node[fcap] at (0,-4.75)
    {the same collective:\\ first components $90\%$ correlated};
 \end{scope}
\end{tikzpicture}
\caption{\textbf{Subspace geometry and statistical dependence are different questions.} Left: the dominant error directions of the two sources are exactly perpendicular, so a common-mode index built on principal angles reports zero coupling. Right: the first components of the same two error vectors are $90\%$ correlated. All of the dependence sits in the cross-covariance block $\bm\Sigma_{AB}$, which the marginal ellipses never see. Together with the converse case --- independent sources whose dominant subspaces coincide --- principal angles fail as a dependence test in \emph{both} directions. Measuring cross-source dependence requires temporally aligned, paired residuals.}
\label{fig:cca-counterexample}
\end{figure}
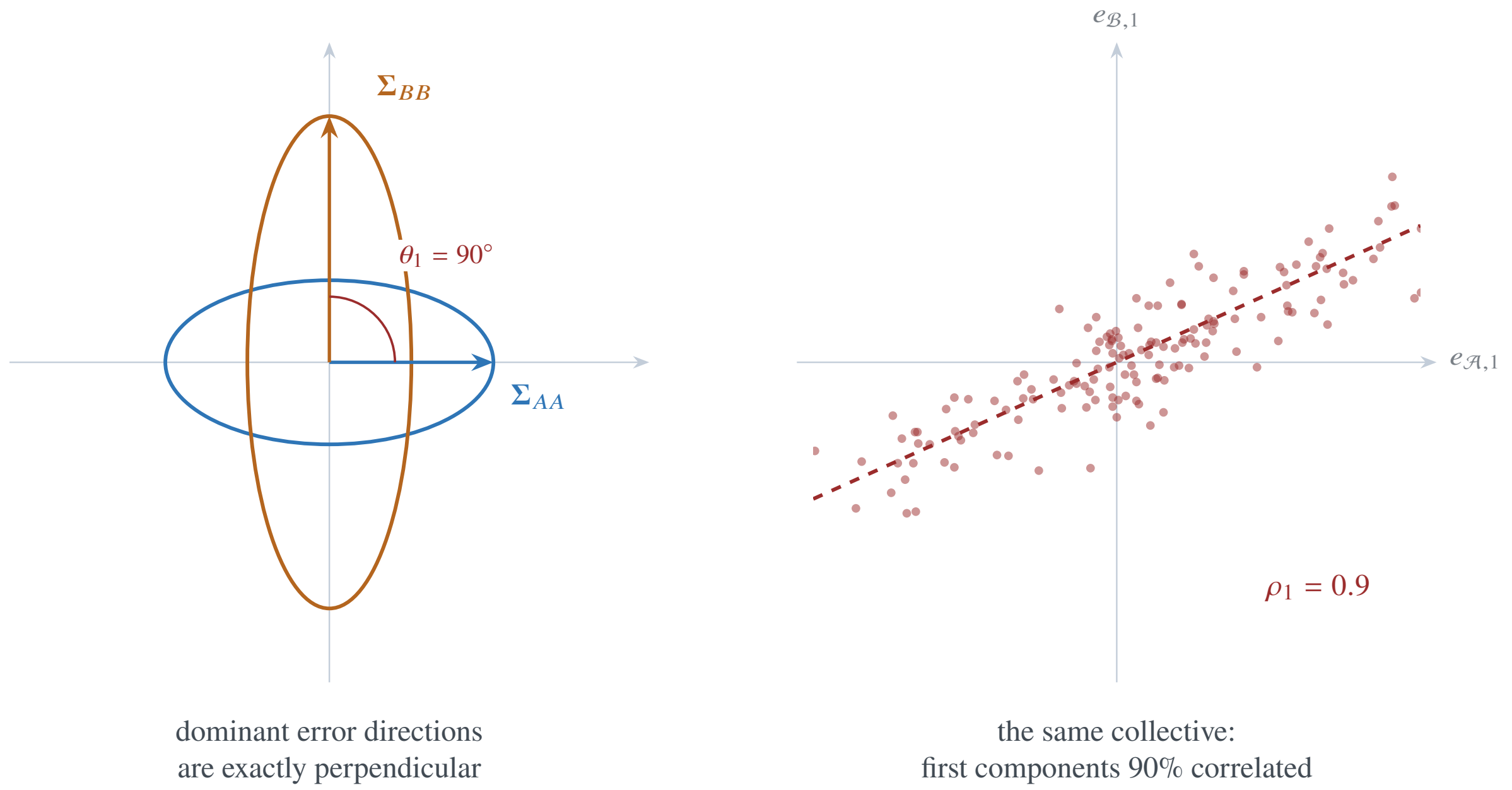

Figure~\ref{fig:cca-counterexample} visualises the converse case. The explicit positive-definite joint covariance and the corresponding CCA calculation are worked out in Section~\ref{sec:cca-counterexample}.

\FloatBarrier

\clearpage
\part*{Part II --- The Generalized SVD}

\section{Statement}

There are two useful ways to state the GSVD. The first is deliberately simple
and is the best one for learning the idea. The second handles arbitrary matrix
shapes and ranks. Keeping these two forms separate prevents a common source of
dimensional mistakes \cite{VanLoan1976,PaigeSaunders1981}.

\subsection{A transparent full-column-rank form}

\begin{theorem}[GSVD, transparent form]\label{thm:gsvd}
Let
\[
   \mat{A}\in\R^{m\times n},\qquad \mat{B}\in\R^{p\times n},
\]
with $n\le \min(m,p)$, and assume that the stacked matrix
\[
   \mat{C}=\begin{pmatrix}\mat{A}\\\mat{B}\end{pmatrix}
\]
has full column rank $n$. Then there exist orthogonal matrices
$\mat{U}\in\R^{m\times m}$ and $\mat{V}\in\R^{p\times p}$, an invertible
$\mat{X}\in\R^{n\times n}$, and numbers $\alpha_i,\beta_i\ge0$ such that
\begin{equation}\label{eq:gsvd}
 \mat{U}\T \mat{A}\mat{X}=\begin{pmatrix}\mat{D}_A\\\mat{0}\end{pmatrix},\qquad
 \mat{V}\T \mat{B}\mat{X}=\begin{pmatrix}\mat{D}_B\\\mat{0}\end{pmatrix},
\end{equation}
where
\[
\mat{D}_A=\diag(\alpha_1,\dots,\alpha_n),\qquad
\mat{D}_B=\diag(\beta_1,\dots,\beta_n),
\]
and
\begin{equation}\label{eq:alphabeta}
   \alpha_i^2+\beta_i^2=1\qquad(i=1,\dots,n).
\end{equation}
The pairs may be ordered so that $\alpha_1\ge\cdots\ge\alpha_n$ and therefore
$\beta_1\le\cdots\le\beta_n$.
\end{theorem}
\begin{practicebox}
This is the form to remember first: one shared state direction $\vekt{x}_i$, two output directions, and two nonnegative gains. After metric-aware scaling, $\alpha_i^2+\beta_i^2=1$ makes the relative split immediately interpretable.
\end{practicebox}

If $\vekt{x}_i$ is column $i$ of $\mat{X}$, and $\vekt{u}_i,\vekt{v}_i$ are the corresponding columns of
$\mat{U},\mat{V}$, then
\begin{equation}\label{eq:gsvdcols}
   \mat{A}\vekt{x}_i=\alpha_i \vekt{u}_i,
   \qquad
   \mat{B}\vekt{x}_i=\beta_i \vekt{v}_i.
\end{equation}
Because $\|\vekt{u}_i\|=\|\vekt{v}_i\|=1$,
\begin{equation}\label{eq:sigmaratio}
  \frac{\|\mat{A}\vekt{x}_i\|}{\|\mat{B}\vekt{x}_i\|}=\frac{\alpha_i}{\beta_i}=\sigma_i
\end{equation}
whenever $\beta_i>0$, with the natural $\infty$ convention when $\beta_i=0$.
The same state direction $\vekt{x}_i$ is therefore simple for both operators at once.
That is the central idea of the GSVD.

The generalized singular value is
\begin{equation}\label{eq:gsvsigma}
   \sigma_i=\frac{\alpha_i}{\beta_i}\in[0,\infty],
\end{equation}
with the conventions $\sigma_i=\infty$ when $\beta_i=0$ and $\sigma_i=0$ when
$\alpha_i=0$.

\paragraph{A new symbol that prevents a serious confusion.}
Because $\alpha_i^2+\beta_i^2=1$, define the \textbf{GSVD mixing angle}
\begin{equation}\label{eq:mixangle}
   \phi_i=\operatorname{atan2}(\beta_i,\alpha_i)\in[0,\pi/2].
\end{equation}
Then
\[
   \alpha_i=\cos\phi_i,\qquad \beta_i=\sin\phi_i,
   \qquad \sigma_i=\cot\phi_i.
\]
We use $\phi$ for GSVD mixing angles and reserve $\theta$ for principal angles
between subspaces. They are not the same object.

\begin{collective}
The normalisation in \eqref{eq:alphabeta} makes the two contributions a
partition of one pooled unit. Along a generalized direction,
$\phi_i=0$ means only $\mat{A}$ responds, $\phi_i=\pi/2$ means only $\mat{B}$ responds, and
$\phi_i=\pi/4$ means a 50--50 relative split. This says nothing yet about how
large the pooled information is in ordinary state units; that absolute scale is
encoded by $\mat{J}=\mat{A}\T \mat{A}+\mat{B}\T \mat{B}$.
\end{collective}

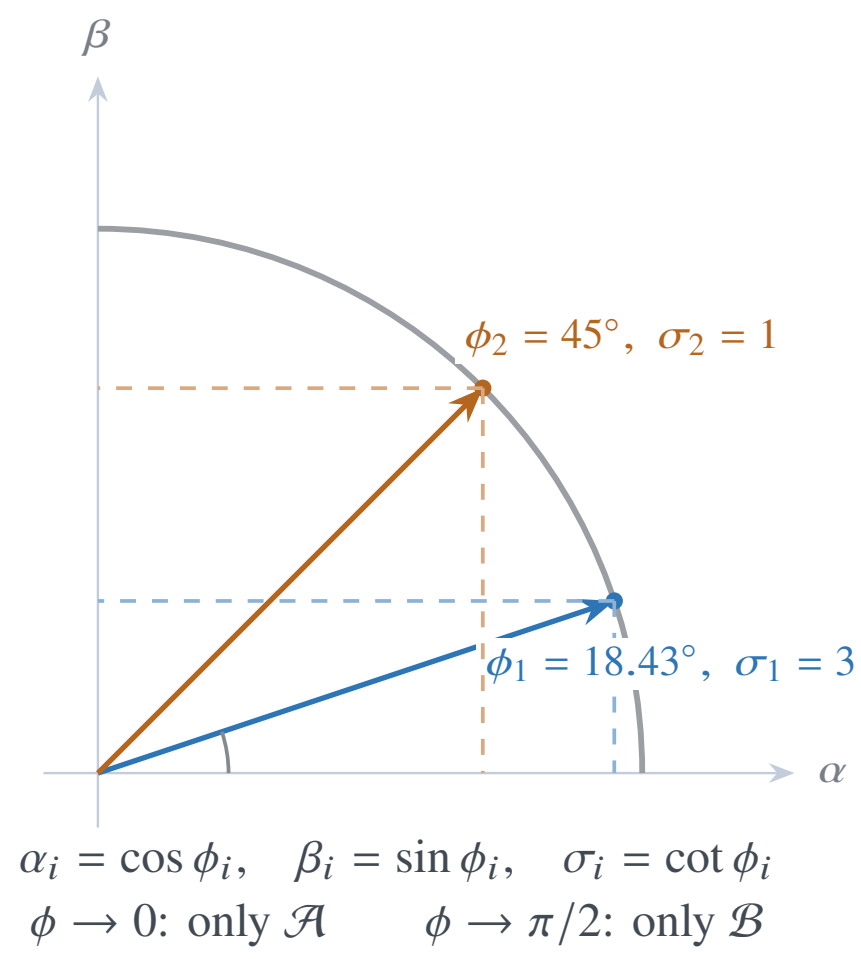
\begin{figure}[htbp]
\centering
\begin{tikzpicture}[figbase,scale=3.6]
  \draw[ax] (-0.10,0)--(1.28,0) node[right,tick]{$\alpha$};
  \draw[ax] (0,-0.10)--(0,1.28) node[above,tick]{$\beta$};
  \draw[struct!45,crv] (1,0) arc (0:90:1);
  \draw[vec,srcA] (0,0)--(0.9487,0.3162);
  \fill[srcA] (0.9487,0.3162) circle (0.45pt);
  \draw[srcA!55,dashed,thincrv] (0.9487,0.3162)--(0.9487,0) (0.9487,0.3162)--(0,0.3162);
  \node[lbl,text=srcA,font=\scriptsize] at (1.05,0.20) {$\phi_1=18.43^\circ,\ \sigma_1=3$};
  \draw[vec,srcB] (0,0)--(0.7071,0.7071);
  \fill[srcB] (0.7071,0.7071) circle (0.45pt);
  \draw[srcB!55,dashed,thincrv] (0.7071,0.7071)--(0.7071,0) (0.7071,0.7071)--(0,0.7071);
  \node[lbl,text=srcB,font=\scriptsize] at (0.96,0.80) {$\phi_2=45^\circ,\ \sigma_2=1$};
  \draw[struct!55,thincrv] (0.24,0) arc (0:18.43:0.24);
  \node[fcap] at (0.55,-0.22)
    {$\alpha_i=\cos\phi_i$,\quad $\beta_i=\sin\phi_i$,\quad $\sigma_i=\cot\phi_i$\\[1pt]
     $\phi\to0$: only $\mathcal A$\qquad $\phi\to\pi/2$: only $\mathcal B$};
\end{tikzpicture}
\caption{\textbf{The mixing angle collects the entire two-source split into one point on the unit circle.} Because $\alpha_i^2+\beta_i^2=1$ the pair is a partition of a single unit budget: neither source can gain along a direction without the other losing. The dashed projections read off the two shares, and $\sigma_i$ is the cotangent of the angle, so $\phi\to0$ means the direction belongs to $\mathcal A$ alone and $\phi\to\pi/2$ to $\mathcal B$ alone. $\phi_i$ is a property of the source split; it is \emph{not} a principal angle between the agents' observable subspaces.}
\label{fig:gsvd-mixing-circle}
\end{figure}

Figure~\ref{fig:gsvd-mixing-circle} places the whole split on the unit circle,
where $\sigma_i$ appears as a cotangent and the two shares can be read off as
projections.

\FloatBarrier

\subsection{The general compact form}\label{sec:general-gsvd}

When $m<n$ or $p<n$, the rectangular blocks in Theorem~\ref{thm:gsvd} cannot
possibly contain $n$ main-diagonal entries. The correct general statement needs
a slightly more flexible notion.

\begin{definition}[Paired-diagonal matrix]
A matrix $\mat{D}\in\R^{q\times r}$ is \emph{paired-diagonal} here if every column
contains at most one nonzero entry and no row is used by two different nonzero
columns. Thus
\[
   \mat{D}\T \mat{D}=\diag(d_1^2,\dots,d_r^2),
\]
even when $r>q$. The nonzeros need not sit on the ordinary main diagonal.
\end{definition}
\begin{practicebox}
Paired-diagonal structure is needed when one output space has fewer rows than the compact GSVD rank. It prevents a dimensionally impossible demand that every block provide a different diagonal row for every shared direction.
\end{practicebox}

\begin{theorem}[Compact GSVD for arbitrary shapes]\label{thm:gsvd-general}
Let $\mat{A}\in\R^{m\times n}$ and $\mat{B}\in\R^{p\times n}$ and let
\[
   r=\rank\begin{pmatrix}\mat{A}\\\mat{B}\end{pmatrix}.
\]
Then there exist orthogonal $\mat{U}\in\R^{m\times m}$ and
$\mat{V}\in\R^{p\times p}$, a full-column-rank matrix $\mat{X}\in\R^{n\times r}$, and
paired-diagonal matrices $\mat{C}_A\in\R^{m\times r}$,
$\mat{C}_B\in\R^{p\times r}$ such that
\begin{equation}\label{eq:compact-gsvd}
   \mat{U}\T \mat{A}\mat{X}=\mat{C}_A,\qquad \mat{V}\T \mat{B}\mat{X}=\mat{C}_B,
\end{equation}
and
\begin{equation}\label{eq:compact-gram}
   \mat{C}_A\T \mat{C}_A=\diag(\alpha_1^2,\dots,\alpha_r^2),\qquad
   \mat{C}_B\T \mat{C}_B=\diag(\beta_1^2,\dots,\beta_r^2),
\end{equation}
with $\alpha_i^2+\beta_i^2=1$.

The remaining $n-r$ state dimensions form exactly the common null space
\[
   \nullsp(\mat{A})\cap\nullsp(\mat{B}).
\]
No generalized singular value is assigned to those $0/0$ directions.
\end{theorem}
\begin{practicebox}
This is the form that covers real sensor suites: tall or wide matrices, rank deficiency, and common-null directions. Directions invisible to both agents are excluded explicitly instead of being assigned an arbitrary $0/0$ ratio.
\end{practicebox}

The full Paige--Saunders/LAPACK representation expresses the same structure by
orthogonal $\mat{U},\mat{V},\mat{Q}$, structured cosine/sine blocks $\mat{D}_1,\mat{D}_2$, and a common
upper-triangular block $\mat{R}_0$,
\[
   \mat{A}=\mat{U} \mat{D}_1 \mat{R}_0 \mat{Q}\T,\qquad \mat{B}=\mat{V} \mat{D}_2 \mat{R}_0 \mat{Q}\T.
\]
The integers usually called $\mat{K}$ and $\mat{L}$ satisfy $\mat{K}+\mat{L}=r$ and determine the
identity, cosine/sine and zero blocks. This is the form implemented by LAPACK's
\texttt{xGGSVD3} and exposed by Julia's \texttt{GeneralizedSVD}
\cite{PaigeSaunders1981,LAPACKGGSVD3,JuliaGSVD}. The compact form above is easier
to teach because it keeps the essential geometry visible without hiding it in
case-dependent block notation.

\begin{collective}
The split
\[
 \R^n=\text{collectively constrained subspace}\ \oplus\
       \bigl(\nullsp(\mat{A})\cap\nullsp(\mat{B})\bigr)
\]
is more fundamental than any generalized singular value. The dimension
$n-r$ tells you how many directions no agent in the pair constrains at all.
That is an observability statement, not an accuracy statement.
\end{collective}

\section{First proof: the whitening route}\label{sec:proof1}

This proof explains the idea with the least machinery. Assume the hypotheses of
Theorem~\ref{thm:gsvd} and, in addition, that $\mat{B}$ has full column rank.
We do \emph{not} invert a rectangular matrix.

\begin{proof}[Proof of the transparent form when $\rank(\mat{B})=n$]
\textbf{Step 1: use a thin QR factorisation of $\mat{B}$.}
Write
\[
   \mat{B}=\mat{Q}_B \mat{R}_B,
\]
where $\mat{Q}_B\in\R^{p\times n}$ has orthonormal columns and
$\mat{R}_B\in\R^{n\times n}$ is invertible upper triangular.

\textbf{Step 2: measure $\mat{A}$ in the input coordinates supplied by $\mat{B}$.}
Define
\[
   \mat{M}=\mat{A} \mat{R}_B^{-1}.
\]
Take the ordinary SVD
\[
   \mat{M}=\mat{U}_A\mat{\Sigma} \mat{W}\T,
   \qquad \mat{\Sigma}=\diag(s_1,\dots,s_n).
\]

\textbf{Step 3: normalise each ratio onto the unit circle.}
Set
\[
   \alpha_i=\frac{s_i}{\sqrt{1+s_i^2}},\qquad
   \beta_i=\frac{1}{\sqrt{1+s_i^2}},
\]
and $\mat{D}_B=\diag(\beta_i)$. Define
\[
   \mat{X}=\mat{R}_B^{-1}\mat{W} \mat{D}_B.
\]
Then $\alpha_i/\beta_i=s_i$ and
$\alpha_i^2+\beta_i^2=1$.

\textbf{Step 4: verify the two equations.}
For $\mat{B}$,
\[
   \mat{B}\mat{X}=\mat{Q}_B\mat{W}\mat{D}_B.
\]
The columns of $\mat{Q}_B\mat{W}$ are orthonormal, so complete them to an orthogonal matrix
$\mat{V}$; then $\mat{V}\T \mat{B}\mat{X}=\begin{psmallmatrix}\mat{D}_B\\\mat{0}\end{psmallmatrix}$.
For $\mat{A}$,
\[
   \mat{A}\mat{X}=\mat{A}\mat{R}_B^{-1}\mat{W}\mat{D}_B=\mat{M}\mat{W}\mat{D}_B
      =\mat{U}_A\mat{\Sigma} \mat{D}_B
      =\mat{U}_A\begin{pmatrix}\mat{D}_A\\\mat{0}\end{pmatrix},
\]
where $\mat{D}_A=\diag(\alpha_i)$, the zero block having $m-n$ rows. This is the
claimed GSVD, in the same padded form as the $\mat{B}$ equation above.
\end{proof}

\paragraph{What this proof teaches.}
When $\mat{B}$ has full column rank, the generalized singular values are the ordinary
singular values of $\mat{A} \mat{R}_B^{-1}$, where $\mat{R}_B$ is the square factor in a QR
factorisation of $\mat{B}$. If $\mat{B}$ is square, this reduces to the familiar
$\mat{A}\mat{B}^{-1}$. The QR version is the correct rectangular statement and avoids the
misleading idea that a nonsquare $\mat{B}$ has an ordinary inverse.

Swapping $\mat{A}$ and $\mat{B}$ reciprocates the generalized singular values. The
associated subspaces agree after the necessary reordering; individual vectors
can still change sign or rotate inside a repeated-value subspace.

\section{The compact CS decomposition}\label{sec:cs}

The general proof needs one fact about an orthonormal matrix whose rows are split
into two blocks. This is the compact geometric core of the cosine--sine
(CS) decomposition \cite{GolubVanLoan2013,EdelmanWang2020}.

\begin{theorem}[Compact two-block CS decomposition]\label{thm:cs}
Let $\mat{P}\in\R^{(m+p)\times r}$ have orthonormal columns and write
\[
   \mat{P}=\begin{pmatrix}\mat{P}_1\\\mat{P}_2\end{pmatrix},
   \qquad \mat{P}_1\in\R^{m\times r},\quad \mat{P}_2\in\R^{p\times r}.
\]
Then there exists an orthogonal $\mat{W}\in\R^{r\times r}$ such that the columns of
$\mat{T}_1=\mat{P}_1\mat{W}$ are mutually orthogonal, the columns of $\mat{T}_2=\mat{P}_2\mat{W}$ are mutually
orthogonal, and
\[
   \|\mat{T}_1\vekt{e}_i\|=\alpha_i,\qquad
   \|\mat{T}_2\vekt{e}_i\|=\beta_i,
   \qquad \alpha_i^2+\beta_i^2=1.
\]
After normalising the nonzero columns and completing orthonormal bases, there
exist orthogonal $\mat{U},\mat{V}$ with
\[
   \mat{U}\T \mat{T}_1=\mat{C}_A,
   \qquad
   \mat{V}\T \mat{T}_2=\mat{C}_B,
\]
where $\mat{C}_A,\mat{C}_B$ are paired-diagonal and satisfy
\eqref{eq:compact-gram}.
\end{theorem}
\begin{practicebox}
The CS decomposition is the engine behind the GSVD. Once the stacked system has orthonormal columns, each canonical direction carries one unit of total squared response, split exactly between the two row blocks.
\end{practicebox}

\begin{proof}
Because $\mat{P}\T \mat{P}=\mat{I}_r$,
\begin{equation}\label{eq:csconstraint}
   \mat{P}_1\T \mat{P}_1+\mat{P}_2\T \mat{P}_2=\mat{I}_r.
\end{equation}
Diagonalise the first symmetric matrix:
\[
   \mat{P}_1\T \mat{P}_1=\mat{W}\diag(\alpha_1^2,\dots,\alpha_r^2)\mat{W}\T.
\]
Equation~\eqref{eq:csconstraint} immediately gives
\[
   \mat{W}\T \mat{P}_2\T \mat{P}_2\mat{W}
   =\mat{I}-\diag(\alpha_i^2)
   =\diag(\beta_i^2),
\]
where $\beta_i=\sqrt{1-\alpha_i^2}$ in exact arithmetic. Thus
$\mat{T}_1=\mat{P}_1\mat{W}$ and $\mat{T}_2=\mat{P}_2\mat{W}$ each have mutually orthogonal columns, with the stated
norms. Their nonzero normalised columns can be inserted into orthonormal bases
of $\R^m$ and $\R^p$, producing $\mat{U}$ and $\mat{V}$.
\end{proof}

\begin{remark}[The numerically important version of the last line]\label{rem:sines}
The proof writes $\beta_i=\sqrt{1-\alpha_i^2}$ because it is short on paper.
Do not compute a tiny $\beta_i$ that way. If $\alpha_i$ is very close to one,
subtraction loses relative accuracy. In floating point, compute both quantities
from the data:
\[
   \alpha_i=\|\mat{P}_1\mat{W}\vekt{e}_i\|,\qquad
   \beta_i=\|\mat{P}_2\mat{W}\vekt{e}_i\|.
\]
This is exactly what the reference implementation does.
\end{remark}
\begin{practicebox}
Near an exclusive direction, computing a tiny sine from $\sqrt{1-\alpha_i^2}$ can lose most of its relative accuracy. Measuring both block responses directly is the numerically reliable implementation rule.
\end{practicebox}

\begin{remark}[A deeper version of the same trap: perturbed structural zeros]
\label{rem:structural-zero}
Remark~\ref{rem:sines} fixes one way a computed magnitude can be wrongly small:
a formula that loses digits. There is a second, quite different way, and the
same remedy does \emph{not} fix it.

\begin{lemma}\label{lem:forced-ones}
With $\mat{P}=\begin{psmallmatrix}\mat{P}_1\\\mat{P}_2\end{psmallmatrix}$ as above, at least
$\max(0,r-p)$ of the $\alpha_i$ equal exactly $1$, and at least
$\max(0,r-m)$ of the $\beta_i$ equal exactly $1$.
\end{lemma}
\begin{practicebox}
Dimensions alone can force exclusive directions before any numerical values are inspected. If one measurement block has too few rows, some jointly constrained directions must live entirely in the other block.
\end{practicebox}

\begin{proof}
In the frame $\mat{W}$ we have $\mat{P}_2\T \mat{P}_2=\mat{I}-\diag(\alpha_i^2)$, whose rank equals the
number of indices with $\alpha_i\ne1$. Since
$\rank(\mat{P}_2\T \mat{P}_2)=\rank(\mat{P}_2)\le p$, at most $p$ indices can satisfy
$\alpha_i\ne1$. If $r>p$, at least $r-p$ indices therefore have
$\alpha_i=1$ and hence $\beta_i=0$; if $r\le p$, the lower bound is simply
zero. Exchanging the two blocks gives the second claim.
\end{proof}

So whenever $r>p$, some $\beta_i$ vanish \emph{exactly}, as a matter of
structure rather than of numerical accident. In floating point they do not
necessarily vanish: the columns of $\mat{P}$ are orthonormal only to working
precision, so a vector $\mat{P}_2\mat{W}\vekt{e}_i$ that should be zero can appear with norm around
$10^{-14}$. Taking the norm computes that perturbed vector accurately; it does
not tell us whether the exact quantity was zero.

The converse problem is equally important. For
$\mat{A}=[1]$ and $\mat{B}=[\varepsilon]$ with $\varepsilon>0$, the second response is
small but genuinely finite:
\[
   \beta=\frac{\varepsilon}{\sqrt{1+\varepsilon^2}},
   \qquad
   \sigma=\frac{1}{\varepsilon}.
\]
At $\varepsilon=10^{-14}$, replacing ``small'' by ``zero'' would turn the
finite value $10^{14}$ into $\infty$. That changes the scientific conclusion
from \emph{very weakly observed} to \emph{structurally unobservable}.

A fixed magnitude threshold therefore cannot solve the problem. The robust
criterion is structural rank. Because $\mat{A}=\mat{P}_1\mat{R}$ and $\mat{B}=\mat{P}_2\mat{R}$ with $\mat{R}$ full row
rank, $\rank(\mat{A})=\rank(\mat{P}_1)$ and $\rank(\mat{B})=\rank(\mat{P}_2)$. The reference
implementation estimates those two ranks relative to each block's own scale and
keeps exactly that many largest $\alpha_i$ and $\beta_i$ as structurally
nonzero. This removes roundoff-created structural modes without collapsing a
tiny but finite sensor response merely because it is small in absolute value.

\emph{The general lesson is worth separating from the GSVD.} There are two
reasons a computed quantity can be small: a formula lost digits, or an exact
structural zero was perturbed. Reformulating the arithmetic fixes the first
only. The second requires a rank decision, and that decision should be made
explicitly and with a scale-aware tolerance.
\end{remark}
\begin{practicebox}
A tiny finite response and a structural zero have very different engineering meanings. The former is weak observability; the latter is blindness. A rank-aware rule is required to distinguish them reliably in floating point.
\end{practicebox}

\subsection{How the block magnitudes relate to principal angles}
\label{sec:which-angles}

It is tempting to say loosely that the $\alpha_i$ and $\beta_i$ ``are
principal-angle cosines''. That sentence needs two small but important
qualifications: ordinary principal angles stop at the smaller subspace
 dimension, and the GSVD pairs $\beta_i$ in the opposite order from the usual
singular-value convention. The exact statement is the following.

\begin{theorem}[Block magnitudes and principal angles]
\label{thm:which-angles}
Let $\mat{C}=\begin{psmallmatrix}\mat{A}\\\mat{B}\end{psmallmatrix}$, let $r=\rank(\mat{C})$, and let
\[
   \mathcal S_A=\R^m\times\{\vekt{0}\}\subset\R^{m+p},
   \qquad
   \mathcal S_B=\{\vekt{0}\}\times\R^p\subset\R^{m+p}
\]
be the coordinate subspaces associated with the two row blocks. Put
$q_A=\min(r,m)$ and $q_B=\min(r,p)$. If
$0\le\theta^{(A)}_1\le\cdots\le\theta^{(A)}_{q_A}\le\pi/2$ are the principal
angles between $\range(\mat{C})$ and $\mathcal S_A$, then
\[
   (\alpha_1,\ldots,\alpha_r)
   =\operatorname{sort}_{\downarrow}
     \bigl(\cos\theta^{(A)}_1,\ldots,\cos\theta^{(A)}_{q_A},
            \underbrace{0,\ldots,0}_{r-q_A}\bigr).
\]
Likewise, if $\theta^{(B)}_j$ are the principal angles between $\range(\mat{C})$ and
$\mathcal S_B$, then the \emph{multiset} of the $\beta_i$ equals those principal-angle
cosines padded with $r-q_B$ zeros. Under the GSVD convention used here,
$\alpha_i$ are ordered decreasingly and therefore $\beta_i$ increasingly; hence
\[
   (\beta_r,\ldots,\beta_1)
   =\operatorname{sort}_{\downarrow}
     \bigl(\cos\theta^{(B)}_1,\ldots,\cos\theta^{(B)}_{q_B},
            \underbrace{0,\ldots,0}_{r-q_B}\bigr).
\]
\end{theorem}
\begin{practicebox}
This theorem identifies exactly where principal-angle language belongs. The $\alpha_i$ and $\beta_i$ encode how the stacked measurement subspace sits relative to the two coordinate blocks; they are not, in general, the angles between the agents' observable subspaces in state space.
\end{practicebox}

\begin{proof}
Let $\mat{P}\in\R^{(m+p)\times r}$ have orthonormal columns spanning $\range(\mat{C})$ and
write $\mat{P}=\begin{psmallmatrix}\mat{P}_1\\\mat{P}_2\end{psmallmatrix}$. Let
$\mat{E}_A=\begin{psmallmatrix}\mat{I}_m\\\mat{0}\end{psmallmatrix}$; its columns are an
orthonormal basis of $\mathcal S_A$. By the Bj\"orck--Golub characterisation
(Theorem~\ref{thm:angles}), the $q_A$ principal-angle cosines are the singular
values of
\[
   \mat{P}\T \mat{E}_A=\mat{P}_1\T.
\]
The nonzero singular values of $\mat{P}_1\T$ and $\mat{P}_1$ coincide. In the compact CS
construction, $\mat{W}$ diagonalises $\mat{P}_1\T \mat{P}_1$ and the column norms
$\alpha_i=\|\mat{P}_1\mat{W}\vekt{e}_i\|$ are exactly the singular values of $\mat{P}_1$, extended by
zeros when $r>m$ and ordered decreasingly. This proves the first identity.
Replacing $\mat{E}_A$ by $\mat{E}_B=\begin{psmallmatrix}\mat{0}\\\mat{I}_p\end{psmallmatrix}$ gives the
same statement for $\mat{P}_2$ and the $\beta_i$. Because
$\alpha_i^2+\beta_i^2=1$, decreasing $\alpha_i$ means increasing $\beta_i$;
therefore the conventional decreasing singular-value order for the $\mat{B}$ block
is $(\beta_r,\ldots,\beta_1)$.
\end{proof}

\emph{Verified numerically:} over $500$ random cases with
$m,p,n\in\{1,\dots,5\}$, the padded principal-angle cosine lists agreed with the
computed $\alpha_i$ and reversed $\beta_i$ to within $2\times10^{-15}$.

\begin{remark}[Where the confusion comes from]\label{rem:angle-confusion}
Theorem~\ref{thm:which-angles} says precisely how the GSVD block magnitudes
encode principal-angle cosines --- with dimension-dependent padding and, for the
$\mat{B}$ block, a reversed ordering convention. These subspaces live in the
\textbf{measurement space} $\R^{m+p}$, not in the state space $\R^n$. They are therefore \textbf{not}, in
general, the principal angles between $\range(\mat{A}\T)$ and $\range(\mat{B}\T)$.
Section~\ref{sec:mix-not-principal} gives an explicit counterexample in which
the two angle sets differ in their values \emph{and} in how many of them there
are.
\end{remark}
\begin{practicebox}
Whenever angles are reported, state the ambient space and the two subspaces. Otherwise two mathematically different diagnostics can be plotted with the same symbol and lead to a wrong physical interpretation.
\end{practicebox}

\begin{collective}
The two angle families answer different questions and both are worth reporting.
$\phi_i$ answers: \emph{how is the pooled evidence along this direction split
between the two agents' measurement channels?} The state-space angles $\theta_i$
between $\range(\mat{A}\T)$ and $\range(\mat{B}\T)$ answer: \emph{do the two agents look at
the same part of the state at all?} A collective can score identically on the
first and completely differently on the second. Label them separately in any
figure or table.
\end{collective}

\section{Second proof: the stacking route (general case)}\label{sec:proof2}

This proof handles arbitrary shapes and ranks and never forms $\mat{A}\T \mat{A}$, $\mat{B}\T \mat{B}$
or an inverse of $\mat{B}$.

\begin{figure}[htbp]
\centering
\begin{tikzpicture}[figbase]
  \node[bx,draw=srcA,fill=softblue,minimum width=10mm] (A) at (0,0.5) {$\mat A$};
  \node[bx,draw=srcB,fill=srcB!7,minimum width=10mm] (B) at (0,-0.5) {$\mat B$};
  \node[left=2.5mm of A.west,yshift=-5mm] {$\mat C=$};
  \draw[flow] (0.75,0)--(2.45,0);
  \node[above] at (1.60,0.10) {QR / SVD};
  \node[bx,draw=srcA,fill=softblue,minimum width=10mm] (P1) at (3.2,0.5) {$\mat P_1$};
  \node[bx,draw=srcB,fill=srcB!7,minimum width=10mm] (P2) at (3.2,-0.5) {$\mat P_2$};
  \node[bx,draw=struct!45,fill=softgray,minimum width=10mm] (R) at (4.65,0) {$\mat R$};
  \node[struct!70,font=\scriptsize] at (3.2,1.35) {$\mat P\T\mat P=\mat I$};
  \draw[flow] (5.35,0)--(7.05,0);
  \node[above] at (6.20,0.10) {CS};
  \node[bx,draw=srcA,fill=softblue,minimum width=10mm] (DA) at (7.8,0.5) {$\mat D_\alpha$};
  \node[bx,draw=srcB,fill=srcB!7,minimum width=10mm] (DB) at (7.8,-0.5) {$\mat D_\beta$};
  \node[right=3.5mm of DA.east,yshift=-5mm,align=left,font=\scriptsize,struct!85]
     {$\alpha_i^2+\beta_i^2=1$,\\ forced by $\mat P\T\mat P=\mat I$};
  \node[fcap] at (4.3,-1.85)
     {the split of $\mat P$ carries all the difference between the two sources;\\
      the shared factor $\mat R$ carries the pooled geometry};
\end{tikzpicture}
\caption{\textbf{The general proof in one picture.} Stacking $\mat A$ over $\mat B$ and orthonormalising pushes all of the pooled geometry into the shared factor $\mat R$ and all of the difference between the sources into the split of $\mat P$. The normalisation $\alpha_i^2+\beta_i^2=1$ is then not an assumption but a consequence of $\mat P\T\mat P=\mat I$: stacking one operator on the other and orthonormalising divides each unit of response length between the two blocks.}
\label{fig:stacking-construction}
\end{figure}
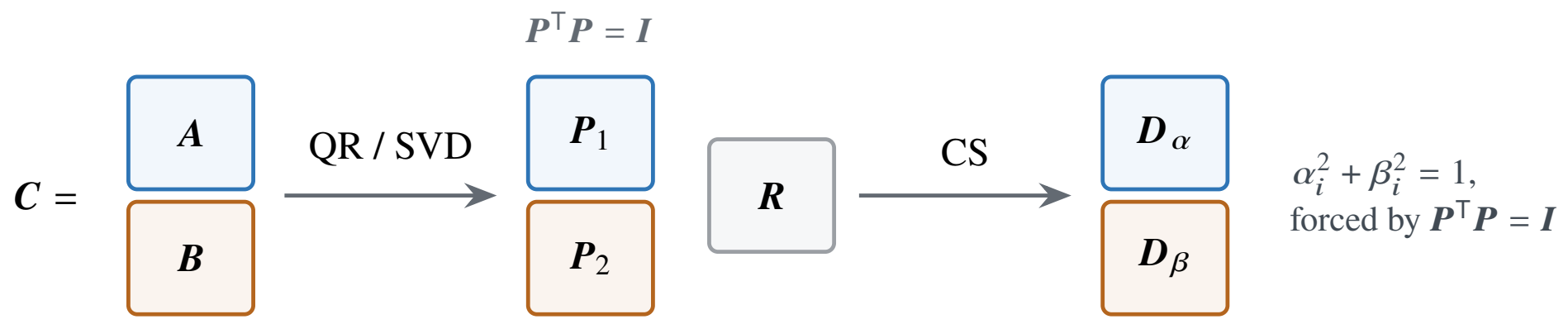

Figure~\ref{fig:stacking-construction} summarises the construction before the
formal argument: the shared factor $\mat{R}$ absorbs the pooled geometry, and the
split of $\mat{P}$ carries everything that distinguishes the two sources.

\FloatBarrier

\begin{proof}[Proof of Theorem~\ref{thm:gsvd-general}]
\textbf{Step 1: stack and factor.}
Let
\[
   \mat{C}=\begin{pmatrix}\mat{A}\\\mat{B}\end{pmatrix},\qquad r=\rank(\mat{C}).
\]
Compute a rank-revealing factorisation
\[
   \mat{C}=PR,
\]
where $\mat{P}\in\R^{(m+p)\times r}$ has orthonormal columns and
$\mat{R}\in\R^{r\times n}$ has full row rank. A thin rank-revealing QR or SVD gives
such a factorisation without forming $\mat{C}\T \mat{C}$.

\textbf{Step 2: split the orthonormal factor.}
Write
\[
   \mat{P}=\begin{pmatrix}\mat{P}_1\\\mat{P}_2\end{pmatrix}.
\]
Then
\[
   \mat{A}=\mat{P}_1\mat{R},\qquad \mat{B}=\mat{P}_2\mat{R},
\]
and $\mat{P}_1\T \mat{P}_1+\mat{P}_2\T \mat{P}_2=\mat{I}_r$.

\textbf{Step 3: apply the compact CS decomposition.}
Choose $\mat{W}$ from Theorem~\ref{thm:cs} and put
\[
   \mat{T}_1=\mat{P}_1\mat{W},\qquad \mat{T}_2=\mat{P}_2\mat{W}.
\]
Choose $\mat{U},\mat{V}$ so that $\mat{U}\T \mat{T}_1=\mat{C}_A$ and $\mat{V}\T \mat{T}_2=\mat{C}_B$.

\textbf{Step 4: choose the shared state directions.}
Define
\[
   \mat{Y}=\mat{W}\T \mat{R}\in\R^{r\times n}.
\]
It has full row rank. Choose specifically its Moore--Penrose right inverse
\begin{equation}\label{eq:Xminnorm}
   \mat{X}=\mat{Y}^+.
\end{equation}
Then $\mat{Y}\mat{X}=\mat{I}_r$, and
\[
   \mat{A}\mat{X}=\mat{P}_1\mat{R}\mat{X}=\mat{T}_1\mat{Y}\mat{X}=\mat{T}_1,
\]
so $\mat{U}\T \mat{A}\mat{X}=\mat{C}_A$; similarly $\mat{V}\T \mat{B}\mat{X}=\mat{C}_B$. This proves
\eqref{eq:compact-gsvd}--\eqref{eq:compact-gram}.
\end{proof}

The choice $\mat{X}=\mat{Y}^+$ is more than a convenience. Other right inverses also satisfy
$\mat{Y}\mat{X}=\mat{I}_r$ and therefore preserve the two diagonalisation equations, but the
Moore--Penrose choice is the minimum-norm one and gives the clean pseudoinverse
information identity in Section~\ref{sec:fisher}.

\begin{collective}
Stacking first does something conceptually important. It separates
\emph{collective strength} from \emph{source attribution}. The factor $\mat{R}$
contains the geometry and scale of the collectively constrained state
subspace. The split of the orthonormal factor $\mat{P}$ into $\mat{P}_1$ and $\mat{P}_2$ says how
that pooled content is divided between the two agents. The GSVD therefore gives
an exact linear attribution only after the pooled geometry has been normalised.
\end{collective}

\section{Properties, special cases, and what can go wrong}

\subsection{\texorpdfstring{$\mat{B}=\mat{I}$}{B = I} recovers the ordinary SVD}
Let $\mat{B}=\mat{I}_n$. Then $\|\mat{B}\vekt{x}\|=\|\vekt{x}\|$, so the generalized Rayleigh quotient is the
ordinary singular-value quotient. If $\mat{A}=\mat{U}\mat{\Sigma} \mat{V}\T$ and $s_i$ are the ordinary
singular values, choose
\[
   \alpha_i=\frac{s_i}{\sqrt{1+s_i^2}},\qquad
   \beta_i=\frac{1}{\sqrt{1+s_i^2}},\qquad
   \mat{X}=\mat{V}\diag(\beta_i).
\]
Then $\alpha_i/\beta_i=s_i$. Thus the ordinary SVD is recovered as the special
case in which the second quadratic form is the Euclidean metric.

\subsection{Relation to the generalized eigenproblem}
In the square full-rank form,
\[
   \mat{A}=\mat{U} \mat{D}_A \mat{X}^{-1},\qquad \mat{B}=\mat{V} \mat{D}_B \mat{X}^{-1}.
\]
Hence
\begin{equation}\label{eq:congruence}
 \mat{A}\T \mat{A}=\mat{X}^{-\mathsf T}\mat{D}_A^2\mat{X}^{-1},\qquad
 \mat{B}\T \mat{B}=\mat{X}^{-\mathsf T}\mat{D}_B^2\mat{X}^{-1}.
\end{equation}
Therefore
\[
   \mat{X}\T(\mat{A}\T \mat{A})\mat{X}=\mat{D}_A^2,\qquad
   \mat{X}\T(\mat{B}\T \mat{B})\mat{X}=\mat{D}_B^2,
\]
and whenever $\beta_i>0$,
\[
   \mat{A}\T \mat{A}\,\vekt{x}_i=\sigma_i^2 \mat{B}\T \mat{B}\,\vekt{x}_i.
\]
This is the generalized eigenvalue connection. It is a characterisation, not a
recommended floating-point algorithm.

\subsection{The pooled metric and the information identity}\label{sec:fisher}
Define
\[
   \mat{J}=\mat{A}\T \mat{A}+\mat{B}\T \mat{B}.
\]
Adding the two diagonal relations gives, in the full-column-rank case,
\begin{equation}\label{eq:Jmetric}
   \mat{X}\T \mat{J}\mat{X}=\mat{I},
\end{equation}
and therefore
\begin{equation}\label{eq:XXT}
   \mat{X}\mat{X}\T=\mat{J}^{-1}.
\end{equation}
In the rank-deficient compact form, if $\mat{X}$ is the Moore--Penrose choice
\eqref{eq:Xminnorm}, the corresponding identities are
\begin{equation}\label{eq:pseudoinfo}
   \mat{X}\T \mat{J}\mat{X}=\mat{I}_r,\qquad \mat{X}\mat{X}\T=\mat{J}^+.
\end{equation}
The second equality would not hold for an arbitrary right inverse.

\begin{collective}[ --- the central statistical interpretation]
Assume independent linear-Gaussian measurements
\[
 \vekt{z}_{\mathcal A}=\mat{H}_{\mathcal A}\vekt{x}+\vekt{\eta}_{\mathcal A},\qquad
 \vekt{\eta}_{\mathcal A}\sim\mathcal N(\vekt{0},\mat{\Sigma}_{\mathcal A}),
\]
\[
 \vekt{z}_{\mathcal B}=\mat{H}_{\mathcal B}\vekt{x}+\vekt{\eta}_{\mathcal B},\qquad
 \vekt{\eta}_{\mathcal B}\sim\mathcal N(\vekt{0},\mat{\Sigma}_{\mathcal B}),
\]
and whiten them as in \eqref{eq:scaledops}. Then
\[
   \mat{I}_{\mathcal A}=\mat{A}\T \mat{A},\qquad \mat{I}_{\mathcal B}=\mat{B}\T \mat{B},
   \qquad \mat{J}=\mat{I}_{\mathcal A}+\mat{I}_{\mathcal B}.
\]
For every generalized direction $\vekt{x}_i$ normalised by
$\vekt{x}_i\T \mat{J}\vekt{x}_i=1$,
\[
   \vekt{x}_i\T \mat{I}_{\mathcal A}\vekt{x}_i=\alpha_i^2,\qquad
   \vekt{x}_i\T \mat{I}_{\mathcal B}\vekt{x}_i=\beta_i^2.
\]
Hence $\alpha_i^2$ and $\beta_i^2$ are \textbf{exact relative Fisher
information shares} along a direction normalised by the pooled information.
This is the precise sense in which the GSVD attributes ``who contributes what''. 
\end{collective}

\begin{figure}[htbp]
\centering
\begin{tikzpicture}[figbase,scale=3.0]
  \draw[ax] (-1.45,0)--(1.55,0) node[right,tick]{$x_1$};
  \draw[ax] (0,-0.95)--(0,1.00) node[above,tick]{$x_2$};
  \draw[rulegray!170,dashed,thincrv] (0,0) circle (1);
  \node[struct!55,font=\scriptsize] at (0.66,0.86) {Euclidean unit circle};
  \draw[pooled,crv,rotate=152.496] (0,0) ellipse (1.1226 and 0.1992);
  \node[lbl,text=pooled,font=\scriptsize] at (-0.80,0.64)
     {$\{\vekt x:\vekt x\T\mat J\vekt x=1\}$};
  \draw[vec,srcA] (0,0)--(0,0.2236);
  \fill[srcA] (0,0.2236) circle (0.5pt);
  \node[lbl,text=srcA,font=\scriptsize] at (0.46,0.33) {$\vekt x_1=(0,\,0.2236)\T$};
  \draw[vec,srcB] (0,0)--(1,-0.5);
  \fill[srcB] (1,-0.5) circle (0.5pt);
  \node[lbl,text=srcB,font=\scriptsize] at (1.20,-0.42) {$\vekt x_2=(1,-0.5)\T$};
  \draw[warn,thincrv] (0,0.16) arc (90:-26.57:0.16);
  \node[lbl,text=warn,font=\scriptsize] at (0.285,0.045) {$116.57^\circ$};
  \node[fcap] at (0.05,-1.30)
    {both directions lie \emph{on} the teal ellipse: $\mat X\T\mat J\mat X=\mat I$.\\
     They are $\mat J$-orthonormal; only the Euclidean eye calls them oblique.};
\end{tikzpicture}
\caption{\textbf{The generalized directions are orthonormal in the pooled metric, not the Euclidean one.} Both columns of $\mat X$ lie exactly on the teal ellipse $\{\vekt x:\vekt x\T\mat J\vekt x=1\}$ while enclosing $116.57^\circ$ in the Euclidean sense. This is the geometric content of $\mat X\T\mat J\mat X=\mat I$: the obliqueness is an artefact of measuring angles with the wrong inner product. A collective of heterogeneous sources supplies its own natural metric, and in that metric the source-resolved directions are as orthogonal as any eigenbasis.}
\label{fig:pooled-metric-geometry}
\end{figure}
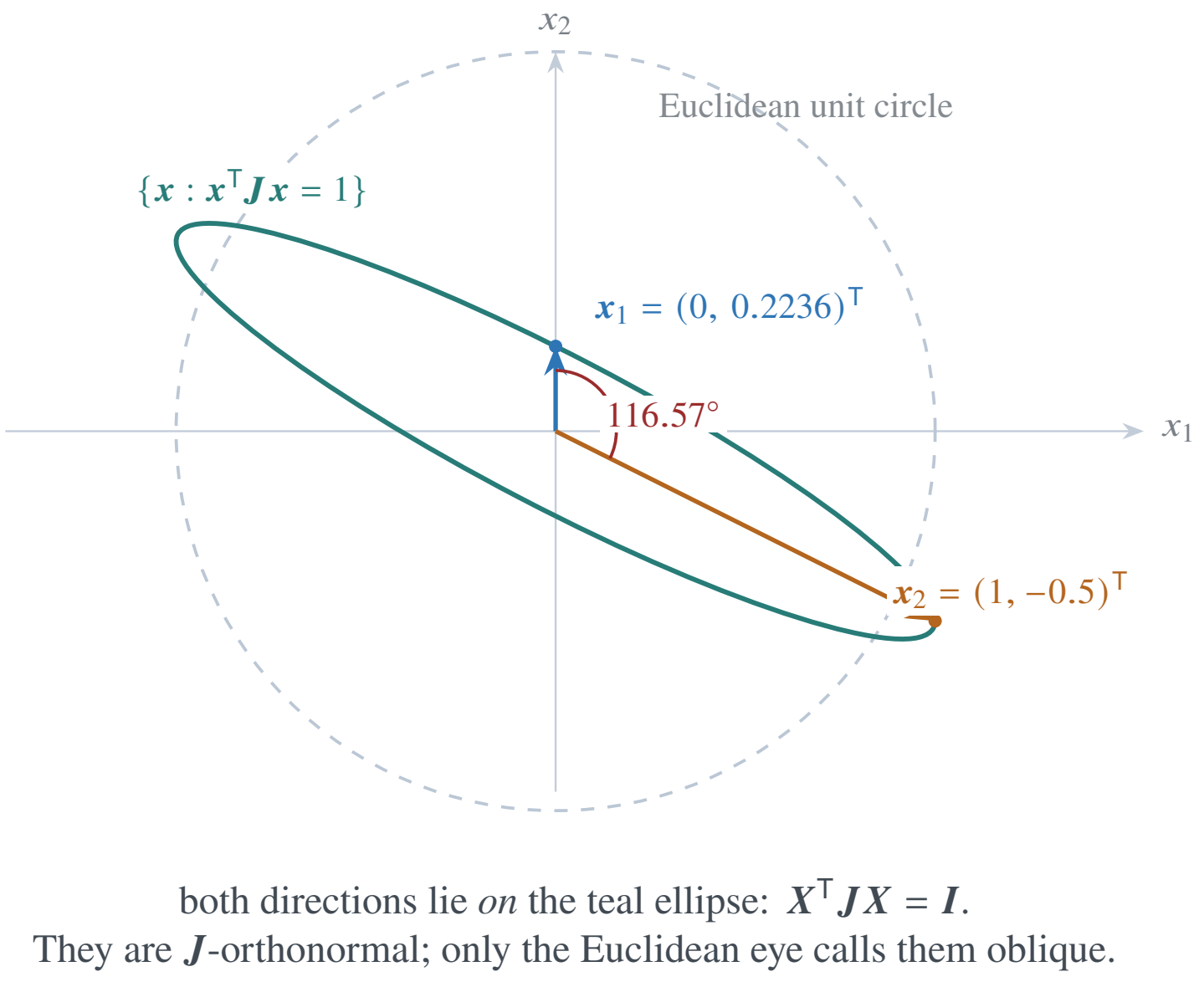

The two-dimensional matrix pair worked through in Part~III provides a concrete picture of this metric: its generalized directions look oblique in Euclidean coordinates but are exactly orthonormal under the pooled information metric. Figure~\ref{fig:pooled-metric-geometry} draws it.

\FloatBarrier

There is an important statistical distinction. $\mat{J}^{-1}$ is the Cram\'er--Rao
lower bound for unbiased estimators. In the linear-Gaussian model with no prior,
the generalized least-squares/maximum-likelihood estimator attains that bound,
so its covariance is exactly $\mat{J}^{-1}$. With a Gaussian prior
$\vekt{x}\sim\mathcal N(\vekt{\mu}_0,\mat{P}_0)$, however, the posterior covariance is
\[
   \mat{P}_{\rm post}=(\mat{P}_0^{-1}+\mat{J})^{-1}.
\]
The prior is then a third information source; $\mat{J}^{-1}$ alone is not the
posterior covariance.

\begin{remark}[Correlated agents require the full joint covariance]
If the two noises are correlated, write
\[
 \vekt{\eta}=\begin{pmatrix}\vekt{\eta}_{\mathcal A}\\\vekt{\eta}_{\mathcal B}\end{pmatrix},\qquad
 \mat{\Sigma}=
 \begin{pmatrix}
   \mat{\Sigma}_{AA}&\mat{\Sigma}_{AB}\\
   \mat{\Sigma}_{BA}&\mat{\Sigma}_{BB}
 \end{pmatrix},\qquad
 \mat{H}=\begin{pmatrix}\mat{H}_{\mathcal A}\\\mat{H}_{\mathcal B}\end{pmatrix}.
\]
The correct joint Fisher information is
\[
   \mat{I}_{\rm joint}=\mat{H}\T\mat{\Sigma}^{-1}\mat{H}.
\]
It is generally \emph{not} the sum of two separately whitened agent matrices.
Moreover, joint whitening by $\mat{\Sigma}^{-1/2}$ mixes the two measurement blocks,
so an additive attribution to individual agents is no longer canonical without
additional modelling choices. Principal angles do not repair this; dependence
must be estimated from paired data or explicitly modelled in $\mat{\Sigma}$.
\end{remark}
\begin{practicebox}
If the agent errors are correlated, separate whitening and simple information addition are no longer valid. The full joint covariance --- or another explicit dependency model --- must enter before information can be fused or attributed correctly.
\end{practicebox}

\subsection{Uniqueness and sensitivity}
The generalized singular values, understood with multiplicities and the
$0/\infty$ conventions, are invariant properties of the pair. Individual
vectors are less rigid:
\begin{enumerate}[leftmargin=1.6em,itemsep=2pt]
\item signs are arbitrary;
\item repeated values permit arbitrary orthogonal rotations inside the
      corresponding invariant subspace;
\item zero output factors make the associated output vector arbitrary.
\end{enumerate}

For interpretation, the stable object is therefore often a subspace rather than
one vector. A useful bounded separation measure is
\[
   g_i=\min_{j\neq i}|\alpha_i^2-\alpha_j^2|,
\]
or equivalently a gap between the mixing angles $\phi_i$. Small $g_i$ means that
a small perturbation can rotate individual generalized directions strongly.
This is the same invariant-subspace principle formalised by classical
perturbation theory \cite{DavisKahan1970,StewartSun1990}.

\subsection{Degenerate values and what they mean}
\begin{center}
\begin{tabular}{@{}llll@{}}
\toprule
$\alpha_i$ & $\beta_i$ & $\sigma_i$ & Meaning\\
\midrule
$>0$ & $>0$ & finite & both contribute; $\sigma_i$ is their relative balance\\
$1$ & $0$ & $\infty$ & $\mat{B}$ is blind, $\mat{A}$ is not\\
$0$ & $1$ & $0$ & $\mat{A}$ is blind, $\mat{B}$ is not\\
$0$ & $0$ & undefined & common-null direction; excluded\\
\bottomrule
\end{tabular}
\end{center}

Only the $0$ and $\infty$ cases certify a single-agent observability dependency.
A merely large or small finite ratio indicates imbalance, not loss of rank after
one agent fails.

That statement is correct, but on its own it leaves the reader without any
number for the case that actually occurs in practice, namely a large but finite
ratio. The next subsection supplies it.

\subsection{Resilience: what a finite ratio actually costs}\label{sec:resilience}

\begin{theorem}[Leave-one-agent-out inflation]\label{thm:resilience}
Assume the full-column-rank setting of Theorem~\ref{thm:gsvd}, normalised so
that $\mat{X}\T \mat{J}\mat{X}=\mat{I}_n$, and let $\vekt{c}=\mat{X}^{-1}\vekt{x}$ be the generalized coordinates. In the
independent linear-Gaussian model of Section~\ref{sec:fisher},
\[
   \Var(c_i)\big|_{\text{both agents}}=1.
\]
If $\mat{A}\T \mat{A}$ is invertible, then
\[
   \Var(c_i)\big|_{\mathcal A\ \text{only}}=\frac{1}{\alpha_i^{2}};
\]
if $\mat{B}\T \mat{B}$ is invertible, then symmetrically
\[
   \Var(c_i)\big|_{\mathcal B\ \text{only}}=\frac{1}{\beta_i^{2}}.
\]
Therefore the standard deviation of the generalized coordinate $c_i$ is
inflated by
\[
   \boxed{\ \frac{1}{\alpha_i}=\sqrt{1+\sigma_i^{-2}}\quad(\text{losing }\mathcal B),
   \qquad
   \frac{1}{\beta_i}=\sqrt{1+\sigma_i^{2}}\quad(\text{losing }\mathcal A),\ }
\]
whenever the corresponding single-agent information matrix is invertible.
\end{theorem}
\begin{practicebox}
This converts an abstract GSVD share into an operational failure cost. A design review can read $1/\alpha_i$ directly as the standard-deviation inflation of generalized coordinate $c_i$ when agent $\mathcal B$ disappears, and analogously for $1/\beta_i$.
\end{practicebox}

\begin{proof}
The pooled covariance is $\mat{J}^{-1}$, so the covariance of $\vekt{c}=\mat{X}^{-1}\vekt{x}$ is
$\mat{X}^{-1}\mat{J}^{-1}\mat{X}^{-\mathsf T}=(\mat{X}\T \mat{J}\mat{X})^{-1}=\mat{I}$, giving $\Var(c_i)=1$. With
$\mathcal B$ removed, the covariance is $(\mat{A}\T \mat{A})^{-1}$; transforming to $\vekt{c}$
gives
\[
   \mat{X}^{-1}(\mat{A}\T \mat{A})^{-1}\mat{X}^{-\mathsf T}
   =(\mat{X}\T \mat{A}\T \mat{A} \mat{X})^{-1}=\diag(\alpha_i^2)^{-1}.
\]
Hence $\Var(c_i)=1/\alpha_i^2$. The argument for $\mathcal B$ alone is identical
with $\mat{A},\alpha_i$ replaced by $\mat{B},\beta_i$. Finally
$\alpha_i^2=\sigma_i^2/(1+\sigma_i^2)$ and
$\beta_i^2=1/(1+\sigma_i^2)$ give the two displayed inflation factors.
\end{proof}

\emph{Verified numerically:} worst deviation $6.8\times10^{-12}$ over $300$
random pairs, checking both variances directly against $\mat{J}^{-1}$ and
$(\mat{A}\T \mat{A})^{-1}$.

\begin{remark}[Why this is the number to report]
$\sigma_i$ is a dimensionless \emph{ratio} and says nothing about consequences.
$1/\alpha_i$ is a \emph{factor on a standard deviation}, which is what a design
review or a safety case actually needs. The two degenerate rows of the table
above are just the endpoints of this one continuous quantity: $\alpha_i=0$ gives
$1/\alpha_i=\infty$ (a genuine single point of failure) and $\alpha_i=1$ gives
$1$ (the other agent contributes nothing there). Everything in between is now
quantified instead of merely labelled ``imbalanced''.
\end{remark}
\begin{practicebox}
A ratio such as $\sigma_i$ describes balance but not consequence. Reporting the ratio together with the leave-one-agent-out inflation gives both the structural explanation and the quantity an engineer can use in a tolerance or safety budget.
\end{practicebox}

\begin{collective}
This closes a practical gap. A collective can have no single point of failure in
the strict rank sense and still degrade by an order of magnitude when one agent
drops out --- and for a tracking, docking or intercept task an order of
magnitude is operationally the same as losing the direction.
Section~\ref{sec:worked-audit} works a numerical example in which exactly this
happens.
\end{collective}

\begin{figure}[!htbp]
\centering
\begin{tikzpicture}[figbase,scale=0.49]
  \begin{scope}
  \clip (-15.0,-12.6) rectangle (15.0,12.6);
  \draw[ax] (-14.2,0)--(14.6,0);
  \draw[ax] (0,-11.6)--(0,12.0);
  \foreach \x in {-10,-5,5,10}
     \draw[rulegray!170] (\x,-0.3)--(\x,0.3) node[below=1.5pt,tick]{$\x$};
  \foreach \y in {-10,-5,5,10}
     \draw[rulegray!170] (-0.3,\y)--(0.3,\y) node[left=1.5pt,tick,fill=white,inner sep=0.6pt]{$\y$};
  \draw[srcB,crv] (0,0) ellipse (50 and 1.00);
  \draw[srcA,crv] (0,0) ellipse (2.00 and 10.00);
  \draw[pooled,crv,fill=softgreen] (0,0) ellipse (1.9984 and 0.9950);
  \end{scope}
  \node[lbl,text=srcA] at (4.4,9.4) {LiDAR};
  \node[lbl,text=srcB] at (7.6,3.6) {camera};
  \node[lbl,text=pooled] at (7.2,-3.2) {pooled};
  \draw[srcB,-{Stealth[length=2.0mm]},line width=0.8pt] (10.6,2.7)--(14.4,1.35);
  \node[lbl,text=srcB,font=\scriptsize] at (8.4,5.6) {continues to $50$\,cm};
  \node[tick] at (12.6,-1.6) {along-range [cm]};
  \node[tick,rotate=90] at (2.9,7.2) {cross-range [cm]};
\end{tikzpicture}
\caption{\textbf{The sensor-suite audit drawn to scale, and what a scalar accuracy figure hides.} The two $1\sigma$ ellipses are almost perfectly complementary: the LiDAR is narrow along range and wide across it, the camera the reverse. The pooled ellipse is close to their intersection, so each direction is carried by whichever sensor is good at it. The consequence is strongly asymmetric and invisible in a single accuracy number: removing the camera inflates the cross-range standard deviation by $10.05\times$, removing the LiDAR inflates the along-range one by $25.02\times$. Neither generalized singular value is $0$ or $\infty$, so by the rank criterion there is no single point of failure --- which is true, and almost irrelevant.}
\label{fig:lidar-camera-audit}
\end{figure}
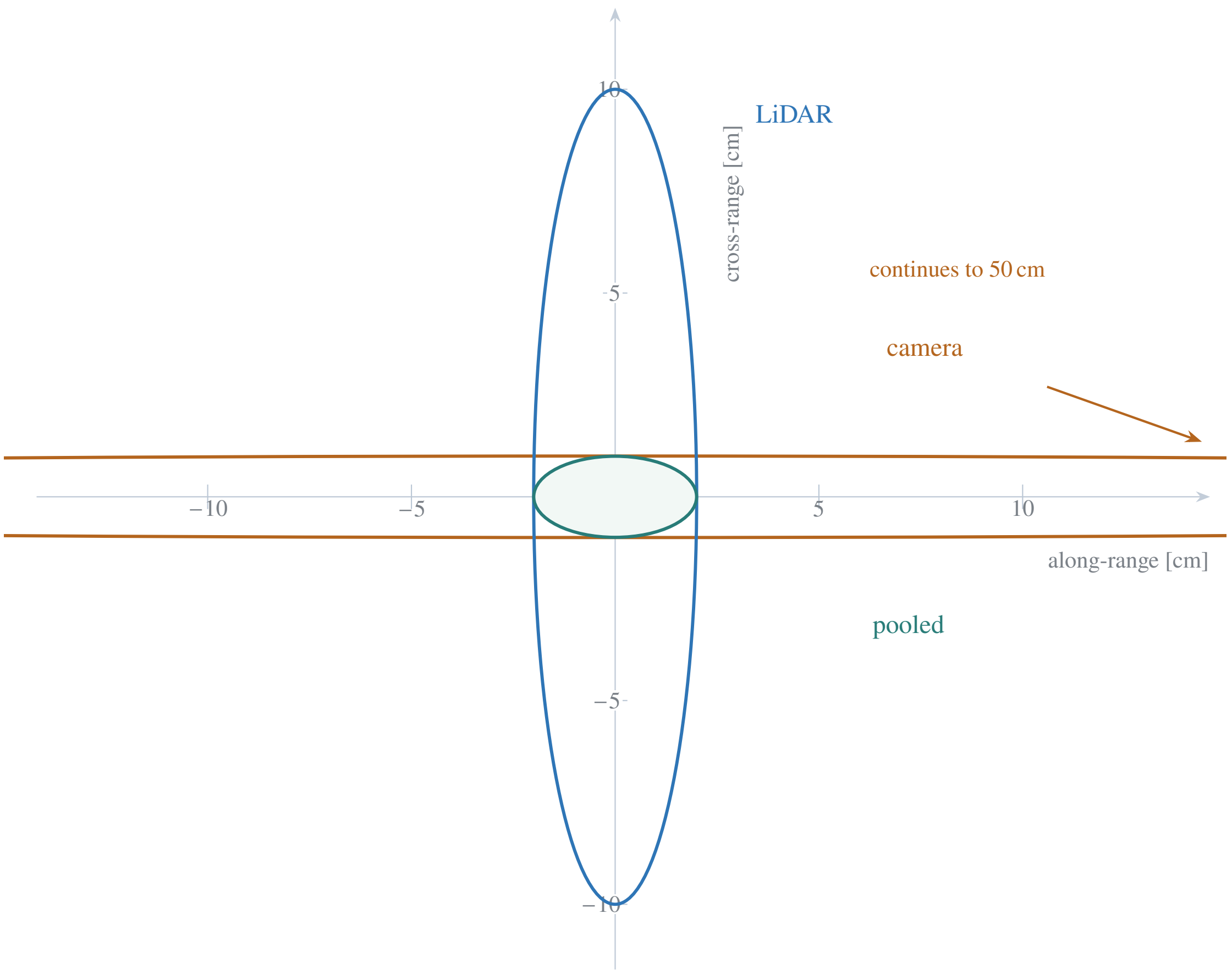

Figure~\ref{fig:lidar-camera-audit} previews that consequence with the verified LiDAR--camera audit developed in Section~\ref{sec:worked-audit}: the rank criterion remains benign while the leave-one-source-out inflation becomes operationally severe.

\FloatBarrier

\clearpage
\part*{Part III --- Worked Examples, Every Step}

Everything below uses one small pair of matrices, examined through five different
lenses, plus five short special-case examples. Working one pair thoroughly
teaches more than ten pairs superficially. All decimal values printed here were
verified in double precision.

\section{The running example}\label{sec:running}

\begin{equation}\label{eq:AB}
   \mat{A}=\begin{pmatrix}1&3\\2&3\end{pmatrix},\qquad
   \mat{B}=\begin{pmatrix}1&1\\0&1\end{pmatrix}.
\end{equation}

Think of it as: two agents, a two-dimensional state $\vekt{x}=(x_1,x_2)\T$, two
measurements each. Neither matrix is symmetric, neither is orthogonal, and $\mat{X}$
will come out genuinely oblique --- so the example still exhibits a genuinely non-Euclidean-orthogonal shared frame.

\subsection{Lens 1: the whitening route}\label{sec:lens1}

\textbf{Step 1: invert $\mat{B}$.} For a $2\times2$ matrix,
$\begin{psmallmatrix}a&b\\c&d\end{psmallmatrix}^{-1}
=\frac{1}{ad-bc}\begin{psmallmatrix}d&-b\\-c&a\end{psmallmatrix}$.
Here $ad-bc = 1\cdot1-1\cdot0=1$, so
\[
   \mat{B}^{-1}=\begin{pmatrix}1&-1\\0&1\end{pmatrix}.
\]
Check: $\mat{B}\mat{B}^{-1}=\begin{psmallmatrix}1&1\\0&1\end{psmallmatrix}
\begin{psmallmatrix}1&-1\\0&1\end{psmallmatrix}
=\begin{psmallmatrix}1&-1+1\\0&1\end{psmallmatrix}=\mat{I}$ \checkmark

\textbf{Step 2: form $\mat{M}=\mat{A}\mat{B}^{-1}$.}
\[
\mat{M}=\begin{pmatrix}1&3\\2&3\end{pmatrix}\begin{pmatrix}1&-1\\0&1\end{pmatrix}
 =\begin{pmatrix}1\cdot1+3\cdot0 & 1\cdot(-1)+3\cdot1\\
                 2\cdot1+3\cdot0 & 2\cdot(-1)+3\cdot1\end{pmatrix}
 =\begin{pmatrix}1&2\\2&1\end{pmatrix}.
\]
This is exactly the matrix whose SVD we computed in Section~\ref{sec:svdex}. So
we already know:
\[
   s_1=3,\quad s_2=1,\qquad
   \mat{U}=\frac{1}{\sqrt2}\begin{pmatrix}1&-1\\1&1\end{pmatrix},\qquad
   \mat{V}=\frac{1}{\sqrt2}\begin{pmatrix}1&1\\1&-1\end{pmatrix}.
\]
Here $s_i$ are the \emph{ordinary} singular values of $\mat{M}$, following the
notation of Section~\ref{sec:proof1}. They will turn out to equal the
generalized singular values $\sigma_i$ of the pair $(\mat{A},\mat{B})$, but that is a
conclusion of the construction, not a definition.

\textbf{Step 3: compute $\alpha_i$ and $\beta_i$.} Using
$\alpha_i=s_i/\sqrt{1+s_i^2}$, $\beta_i=1/\sqrt{1+s_i^2}$:
\begin{align*}
s_1=3: &\quad \sqrt{1+9}=\sqrt{10}=3.162278,\quad
   \alpha_1=\tfrac{3}{\sqrt{10}}=0.948683,\quad \beta_1=\tfrac{1}{\sqrt{10}}=0.316228,\\
s_2=1: &\quad \sqrt{1+1}=\sqrt{2}=1.414214,\quad
   \alpha_2=\tfrac{1}{\sqrt2}=0.707107,\quad \beta_2=\tfrac{1}{\sqrt2}=0.707107.
\end{align*}
Pythagoras check: $\alpha_1^2+\beta_1^2=\tfrac{9}{10}+\tfrac{1}{10}=1$
\checkmark; $\alpha_2^2+\beta_2^2=\half+\half=1$ \checkmark

\textbf{Step 4: build $\mat{X}=\mat{B}^{-1}\mat{V}\mat{D}_B$.} First
\[
\mat{V} \mat{D}_B=\frac{1}{\sqrt2}\begin{pmatrix}1&1\\1&-1\end{pmatrix}
\begin{pmatrix}\tfrac{1}{\sqrt{10}}&0\\0&\tfrac{1}{\sqrt2}\end{pmatrix}
=\begin{pmatrix}\tfrac{1}{\sqrt{20}}&\half\\[2pt] \tfrac{1}{\sqrt{20}}&-\half\end{pmatrix}
=\begin{pmatrix}0.223607&0.5\\0.223607&-0.5\end{pmatrix},
\]
using $\tfrac{1}{\sqrt2}\cdot\tfrac{1}{\sqrt{10}}=\tfrac{1}{\sqrt{20}}=0.223607$
and $\tfrac{1}{\sqrt2}\cdot\tfrac{1}{\sqrt2}=\half$. Then
\[
\mat{X}=\mat{B}^{-1}(\mat{V}\mat{D}_B)=\begin{pmatrix}1&-1\\0&1\end{pmatrix}
\begin{pmatrix}0.223607&0.5\\0.223607&-0.5\end{pmatrix}
=\begin{pmatrix}0.223607-0.223607 & 0.5-(-0.5)\\ 0.223607 & -0.5\end{pmatrix}
\]
\[
   \mat{X}=\boxed{\begin{pmatrix}0&1\\0.223607&-0.5\end{pmatrix}}
\]
Determinant $=0\cdot(-0.5)-1\cdot0.223607=-0.223607\ne0$, so $\mat{X}$ is invertible
\checkmark

\textbf{Step 5: verify the two GSVD equations.}
\[
\mat{A}\mat{X}=\begin{pmatrix}1&3\\2&3\end{pmatrix}\begin{pmatrix}0&1\\0.223607&-0.5\end{pmatrix}
  =\begin{pmatrix}0+0.670820 & 1-1.5\\ 0+0.670820 & 2-1.5\end{pmatrix}
  =\begin{pmatrix}0.670820&-0.5\\0.670820&0.5\end{pmatrix}.
\]
\begin{align*}
\mat{U}\T \mat{A}\mat{X}
&=\frac{1}{\sqrt2}\begin{pmatrix}1&1\\-1&1\end{pmatrix}
  \begin{pmatrix}0.670820&-0.5\\0.670820&0.5\end{pmatrix}\\
&=\frac{1}{\sqrt2}\begin{pmatrix}1.341641&0\\0&1\end{pmatrix}
 =\begin{pmatrix}0.948683&0\\0&0.707107\end{pmatrix}
 =\mat{D}_A\quad\checkmark
\end{align*}
\[
\mat{B}\mat{X}=\begin{pmatrix}1&1\\0&1\end{pmatrix}\begin{pmatrix}0&1\\0.223607&-0.5\end{pmatrix}
  =\begin{pmatrix}0.223607&0.5\\0.223607&-0.5\end{pmatrix},
\]
\begin{align*}
\mat{V}\T \mat{B}\mat{X}
&=\frac{1}{\sqrt2}\begin{pmatrix}1&1\\1&-1\end{pmatrix}
  \begin{pmatrix}0.223607&0.5\\0.223607&-0.5\end{pmatrix}\\
&=\frac{1}{\sqrt2}\begin{pmatrix}0.447214&0\\0&1\end{pmatrix}
 =\begin{pmatrix}0.316228&0\\0&0.707107\end{pmatrix}
 =\mat{D}_B\quad\checkmark
\end{align*}

\textbf{Step 6: the obliqueness, made concrete.} The two columns of $\mat{X}$ are
$\vekt{x}_1=(0,\,0.223607)\T$ and $\vekt{x}_2=(1,\,-0.5)\T$. Their inner product is
\[
   \vekt{x}_1\T \vekt{x}_2 = 0\cdot1 + 0.223607\cdot(-0.5) = -0.111803 \ne 0 .
\]
The angle between them is
$\arccos\!\bigl(-0.111803/(0.223607\cdot1.118034)\bigr)=\arccos(-0.447214)=116.57^\circ$.
\textbf{Not orthogonal --- and this is normal, not a mistake.}

\textbf{Step 7: the ratios.} From \eqref{eq:gsvsigma} and \eqref{eq:gsvdcols} we expect
$\|\mat{A}\vekt{x}_i\|/\|\mat{B}\vekt{x}_i\|=\sigma_i$. Directly:
\[
\mat{A}\vekt{x}_1=\binom{0.670820}{0.670820},\ \ \|\mat{A}\vekt{x}_1\|=0.948683;\qquad
\mat{B}\vekt{x}_1=\binom{0.223607}{0.223607},\ \ \|\mat{B}\vekt{x}_1\|=0.316228,
\]
ratio $=0.948683/0.316228=3.000000=\sigma_1$ \checkmark
\[
\mat{A}\vekt{x}_2=\binom{-0.5}{0.5},\ \|\mat{A}\vekt{x}_2\|=0.707107;\qquad
\mat{B}\vekt{x}_2=\binom{0.5}{-0.5},\ \|\mat{B}\vekt{x}_2\|=0.707107,
\]
ratio $=1=\sigma_2$ \checkmark

\begin{collective}
Read the result. Direction $\vekt{x}_1\propto\vekt{e}_2$ --- the pure second state-coordinate axis --- is seen
three times more strongly by agent $\mathcal A$ than by agent $\mathcal B$.
Direction $\vekt{x}_2\propto(2,-1)\T$ has an equal relative split. These are
relative statements only: neither finite ratio says that one agent can be
removed without loss of observability. To decide that, inspect the ranks or the
$0/\infty$ cases. For task allocation along $(0,1)\T$, however, the static
comparison clearly says that $\mathcal A$ contributes more information.
\end{collective}

\subsection{Lens 2: the stacking / CS route}\label{sec:twoRoutes}

Now the same pair via Section~\ref{sec:proof2}, which never inverts $\mat{B}$.

\textbf{Step 1: stack.}
\[
\mat{C}=\begin{pmatrix}\mat{A}\\\mat{B}\end{pmatrix}=\begin{pmatrix}1&3\\2&3\\1&1\\0&1\end{pmatrix},
\qquad
\mat{C}\T \mat{C}=\begin{pmatrix}1+4+1+0 & 3+6+1+0\\ 3+6+1+0 & 9+9+1+1\end{pmatrix}
     =\begin{pmatrix}6&10\\10&20\end{pmatrix}.
\]
(We form $\mat{C}\T \mat{C}$ here only to keep the hand arithmetic short; a real
implementation uses QR on $\mat{C}$ directly. See Remark~\ref{rem:sines} and
Section~\ref{sec:numerics}.)

\textbf{Step 2: Cholesky, $\mat{C}\T \mat{C}=\mat{R}\T \mat{R}$.} Write
$\mat{R}=\begin{psmallmatrix}r_{11}&r_{12}\\0&r_{22}\end{psmallmatrix}$. Matching
entries: $r_{11}^2=6\Rightarrow r_{11}=\sqrt6=2.449490$;
$r_{11}r_{12}=10\Rightarrow r_{12}=10/\sqrt6=4.082483$;
$r_{12}^2+r_{22}^2=20\Rightarrow r_{22}^2=20-\tfrac{100}{6}=\tfrac{10}{3}
\Rightarrow r_{22}=1.825742$. So
\[
\mat{R}=\begin{pmatrix}2.449490&4.082483\\0&1.825742\end{pmatrix},\qquad
\mat{R}^{-1}=\begin{pmatrix}0.408248&-0.912871\\0&0.547723\end{pmatrix}.
\]
(For upper triangular $2\times2$: $\mat{R}^{-1}=
\begin{psmallmatrix}1/r_{11} & -r_{12}/(r_{11}r_{22})\\ 0 & 1/r_{22}\end{psmallmatrix}$;
here $-4.082483/(2.449490\cdot1.825742)=-4.082483/4.472136=-0.912871$.)

\textbf{Step 3: split $\mat{Q}=\mat{C}\mat{R}^{-1}$.}
\[
\mat{Q}_1=\mat{A}\mat{R}^{-1}=\begin{pmatrix}1&3\\2&3\end{pmatrix}
\begin{pmatrix}0.408248&-0.912871\\0&0.547723\end{pmatrix}
=\begin{pmatrix}0.408248 & -0.912871+1.643168\\
                0.816497 & -1.825742+1.643168\end{pmatrix},
\]
\[
\mat{Q}_1=\begin{pmatrix}0.408248&0.730297\\0.816497&-0.182574\end{pmatrix},
\]
\[
\mat{Q}_2=\mat{B}\mat{R}^{-1}=\begin{pmatrix}0.408248&-0.912871+0.547723\\0&0.547723\end{pmatrix}
=\begin{pmatrix}0.408248&-0.365148\\0&0.547723\end{pmatrix}.
\]
Sanity check that $\mat{Q}\T \mat{Q}=\mat{I}$ --- first column squared norms:
$0.408248^2+0.816497^2+0.408248^2+0^2 = 0.166667+0.666667+0.166667=1.000000$
\checkmark; the cross term
$0.408248(0.730297)+0.816497(-0.182574)+0.408248(-0.365148)+0
=0.298142-0.149071-0.149071=0$ \checkmark

\textbf{Step 4: CS decomposition.} Following the proof of
Theorem~\ref{thm:cs}, take the SVD of $\mat{Q}_1$ via
\[
\mat{Q}_1\T \mat{Q}_1=\begin{pmatrix}0.833333&0.149071\\0.149071&0.566667\end{pmatrix}.
\]
Trace $=1.4$, determinant $=0.833333\cdot0.566667-0.149071^2=0.472222-0.022222=0.45$.
So the eigenvalues solve $\lambda^2-1.4\lambda+0.45=0$:
$\lambda=\tfrac{1.4\pm\sqrt{1.96-1.8}}{2}=\tfrac{1.4\pm0.4}{2}$, giving
$\lambda=0.9$ and $\lambda=0.5$. Hence
\[
   c_1=\sqrt{0.9}=0.948683,\qquad c_2=\sqrt{0.5}=0.707107 .
\]
Here $c_i$ are the CS cosine magnitudes; in the GSVD they are precisely
$\alpha_i=\cos\phi_i$, where $\phi_i$ is the GSVD mixing angle, not a
principal angle between the agents' observable subspaces.

\textbf{Compare with Lens 1: $c_i=\alpha_i$ exactly.} $0.948683$ and $0.707107$
--- the same numbers, from a completely different computation. That is the
theorem working.

For $\lambda=0.9$, the first eigenvector equation is
\[
  -0.066667\,w_1+0.149071\,w_2=0,
  \qquad w_1=2.236068\,w_2.
\]
After normalisation,
$\vekt{w}^{(1)}=(0.912871,\,0.408248)\T$. An orthogonal second vector is
$\vekt{w}^{(2)}=(-0.408248,\,0.912871)\T$:
\[
\mat{W}=\begin{pmatrix}0.912871&-0.408248\\0.408248&0.912871\end{pmatrix}.
\]

\textbf{Step 5: compute the sine factors from column norms (not from $\sqrt{1-\alpha_i^2}$).}
\[
\mat{T}=\mat{Q}_2\mat{W}=\begin{pmatrix}0.408248&-0.365148\\0&0.547723\end{pmatrix}
\begin{pmatrix}0.912871&-0.408248\\0.408248&0.912871\end{pmatrix}
\]
\[
\mat{T}=\begin{pmatrix}0.223607&-0.500000\\0.223607&0.500000\end{pmatrix}.
\]
Column norms: $\|\mat{T}\vekt{e}_1\|=\sqrt{2(0.223607)^2}=0.316228=\beta_1$ and
$\|\mat{T}\vekt{e}_2\|=\sqrt{2(0.5)^2}=0.707107=\beta_2$. These match Lens~1
exactly \checkmark\ (and indeed $\alpha_1^2+\beta_1^2=0.9+0.1=1$).
Normalising the columns gives
$\mat{V}=\frac{1}{\sqrt2}\begin{psmallmatrix}1&-1\\1&1\end{psmallmatrix}$.

\textbf{Step 6: assemble $\mat{X}=\mat{R}^{-1}\mat{W}$.}
\[
\mat{X}_{\text{CS}}=\mat{R}^{-1}\mat{W}=\begin{pmatrix}0.408248&-0.912871\\0&0.547723\end{pmatrix}
\begin{pmatrix}0.912871&-0.408248\\0.408248&0.912871\end{pmatrix}
\]
\[
=\begin{pmatrix}0.372678-0.372678 & -0.166667-0.833333\\
                0.223607 & 0.500000\end{pmatrix}
=\begin{pmatrix}0&-1\\0.223607&0.5\end{pmatrix}.
\]

\textbf{Step 7: compare the two routes.}
\[
\mat{X}_{\text{Lens 1}}=\begin{pmatrix}0&1\\0.223607&-0.5\end{pmatrix},
\qquad
\mat{X}_{\text{CS}}=\begin{pmatrix}0&-1\\0.223607&0.5\end{pmatrix}.
\]
The first columns are \emph{identical}; the second columns differ by a sign.
Checking the corresponding $\vekt{u}_2$ and $\vekt{v}_2$ shows they too are negated. So the
triple $(\vekt{x}_2,\vekt{u}_2,\vekt{v}_2)\to(-\vekt{x}_2,-\vekt{u}_2,-\vekt{v}_2)$ --- exactly the sign ambiguity discussed
in the uniqueness paragraph of Part~II, and it changes nothing, because $\mat{A}(-\vekt{x}_2)=\alpha_2(-\vekt{u}_2)$ still
holds. \textbf{Two independent derivations, one decomposition.}

\subsection{Lens 3: the generalized eigenproblem}

Verify $\mat{A}\T \mat{A}\,\vekt{x}_i=\sigma_i^2\,\mat{B}\T \mat{B}\,\vekt{x}_i$ directly.
\[
\mat{A}\T \mat{A}=\begin{pmatrix}1&2\\3&3\end{pmatrix}\begin{pmatrix}1&3\\2&3\end{pmatrix}
=\begin{pmatrix}1+4&3+6\\3+6&9+9\end{pmatrix}=\begin{pmatrix}5&9\\9&18\end{pmatrix},
\quad
\mat{B}\T \mat{B}=\begin{pmatrix}1&0\\1&1\end{pmatrix}\begin{pmatrix}1&1\\0&1\end{pmatrix}
=\begin{pmatrix}1&1\\1&2\end{pmatrix}.
\]
The characteristic polynomial of the pencil is
\[
\det(\mat{A}\T \mat{A}-\lambda \mat{B}\T \mat{B})
=\det\begin{pmatrix}5-\lambda&9-\lambda\\9-\lambda&18-2\lambda\end{pmatrix}
=(5-\lambda)(18-2\lambda)-(9-\lambda)^2 .
\]
Expand: $(5-\lambda)(18-2\lambda)=90-10\lambda-18\lambda+2\lambda^2
=2\lambda^2-28\lambda+90$, and $(9-\lambda)^2=\lambda^2-18\lambda+81$. Subtract:
\[
   2\lambda^2-28\lambda+90-\lambda^2+18\lambda-81=\lambda^2-10\lambda+9=(\lambda-9)(\lambda-1).
\]
So $\lambda\in\{9,1\}=\{\sigma_1^2,\sigma_2^2\}$ \checkmark

Eigenvectors: for $\lambda=9$, the first row gives $(5-9)v_1+(9-9)v_2=0
\Rightarrow -4v_1=0\Rightarrow v_1=0$, so $\vekt{v}\propto(0,1)\T\propto \vekt{x}_1$ \checkmark
For $\lambda=1$: $(5-1)v_1+(9-1)v_2=0\Rightarrow 4v_1+8v_2=0\Rightarrow
v_1=-2v_2$, so $\vekt{v}\propto(-2,1)\T\propto(1,-0.5)\T=\vekt{x}_2$ \checkmark

\subsection{Lens 4: pooled information and attribution}

Test \eqref{eq:XXT}. The pooled Gram matrix is
\[
\mat{J}=\mat{A}\T \mat{A}+\mat{B}\T \mat{B}
 =\begin{pmatrix}5&9\\9&18\end{pmatrix}
 +\begin{pmatrix}1&1\\1&2\end{pmatrix}
 =\begin{pmatrix}6&10\\10&20\end{pmatrix}.
\]
Its inverse is
\[
\mat{J}^{-1}=\frac1{20}\begin{pmatrix}20&-10\\-10&6\end{pmatrix}
      =\begin{pmatrix}1&-0.5\\-0.5&0.3\end{pmatrix}.
\]
Now
\[
\mat{X}\mat{X}\T
=\begin{pmatrix}0&1\\0.223607&-0.5\end{pmatrix}
 \begin{pmatrix}0&0.223607\\1&-0.5\end{pmatrix}
=\begin{pmatrix}1&-0.5\\-0.5&0.3\end{pmatrix}=\mat{J}^{-1}\quad\checkmark
\]
The same result can be stated in the pooled metric:
\[
   \mat{X}\T \mat{J}\mat{X}=\mat{I}.
\]
So the two visibly oblique columns of $\mat{X}$ are exactly orthonormal in the
$\mat{J}$-inner product.

\begin{collective}
Assume now that $\mat{A}$ and $\mat{B}$ are whitened independent measurement Jacobians.
Along $\vekt{x}_1$, the pooled information has been normalised to one and
\[
   \alpha_1^2=0.9,\qquad \beta_1^2=0.1.
\]
Thus agent $\mathcal A$ supplies $90\%$ and agent $\mathcal B$ $10\%$ of the
\emph{directional Fisher information}. Along $\vekt{x}_2$ the split is $50/50$.

The distinction is important: $\alpha_i^2$ and $\beta_i^2$ are relative shares;
$\mat{J}$ carries the absolute information scale. In a no-prior linear-Gaussian
estimation problem $\mat{J}^{-1}$ is both the Cram\'er--Rao bound and the covariance
of the efficient GLS/ML estimate. With a prior, the posterior covariance is
instead $(\mat{P}_0^{-1}+\mat{J})^{-1}$.
\end{collective}

\subsection{Lens 5: regularization and filter factors}\label{sec:tikhonov}

Consider generalized Tikhonov regularization
\begin{equation}\label{eq:tikh}
   \min_{\vekt{x}}\ \|\mat{A}\vekt{x}-\vekt{b}\|^2+\lambda^2\|\mat{B}\vekt{x}\|^2,
\end{equation}
where $\mat{A}$ is the data operator and $\mat{B}$ is the regularization operator. This is
one of the classical applications of the GSVD \cite{Hansen1998}.

\textbf{Derivation.} The normal equations are
\[
   (\mat{A}\T \mat{A}+\lambda^2\mat{B}\T \mat{B})\vekt{x}=\mat{A}\T \vekt{b}.
\]
In the full-rank GSVD coordinates $\vekt{c}=\mat{X}^{-1}\vekt{x}$,
\[
   c_i=\frac{\alpha_i(\vekt{u}_i\T \vekt{b})}{\alpha_i^2+\lambda^2\beta_i^2}.
\]
Equivalently,
\begin{equation}\label{eq:filter}
   \boxed{
   \vekt{x}_\lambda=\sum_i f_i\frac{\vekt{u}_i\T \vekt{b}}{\alpha_i}\vekt{x}_i,
   \qquad
   f_i=\frac{\sigma_i^2}{\sigma_i^2+\lambda^2}}
\end{equation}
for the components with $\alpha_i>0$. These are the generalized Tikhonov
filter factors. A component with $\sigma_i\gg\lambda$ is weakly attenuated;
with $\sigma_i\ll\lambda$ it is strongly attenuated.

\textbf{Worked numbers.} Take $\vekt{b}=(1,1)\T$ and $\lambda=1$. Since
\[
\mat{U}\T \vekt{b}=\binom{1.414214}{0},
\]
we have
\[
   f_1=\frac9{10}=0.9,\qquad f_2=\frac12=0.5.
\]
Thus
\[
 c_1=0.9\frac{1.414214}{0.948683}=1.341641,\qquad c_2=0,
\]
and
\[
   \vekt{x}_\lambda=1.341641\binom{0}{0.223607}
            =\binom{0}{0.300000}.
\]
A direct solve of
\[
\begin{pmatrix}6&10\\10&20\end{pmatrix}\vekt{x}=\binom36
\]
gives exactly the same result.

One GSVD gives the whole $\lambda$ sweep:
\begin{center}
\begin{tabular}{@{}cccl@{}}
\toprule
$\lambda$ & $f_1$ & $f_2$ & $\vekt{x}_\lambda$\\
\midrule
$0$   & $1$      & $1$      & $(0,\ 0.333333)\T$\\
$1$   & $0.9$    & $0.5$    & $(0,\ 0.300000)\T$\\
$2$   & $0.692308$ & $0.2$  & $(0,\ 0.230769)\T$\\
$10$  & $0.082569$ & $0.009901$ & $(0,\ 0.027523)\T$\\
$\to\infty$ & $\to0$ & $\to0$ & $\to(0,0)\T$\\
\bottomrule
\end{tabular}
\end{center}

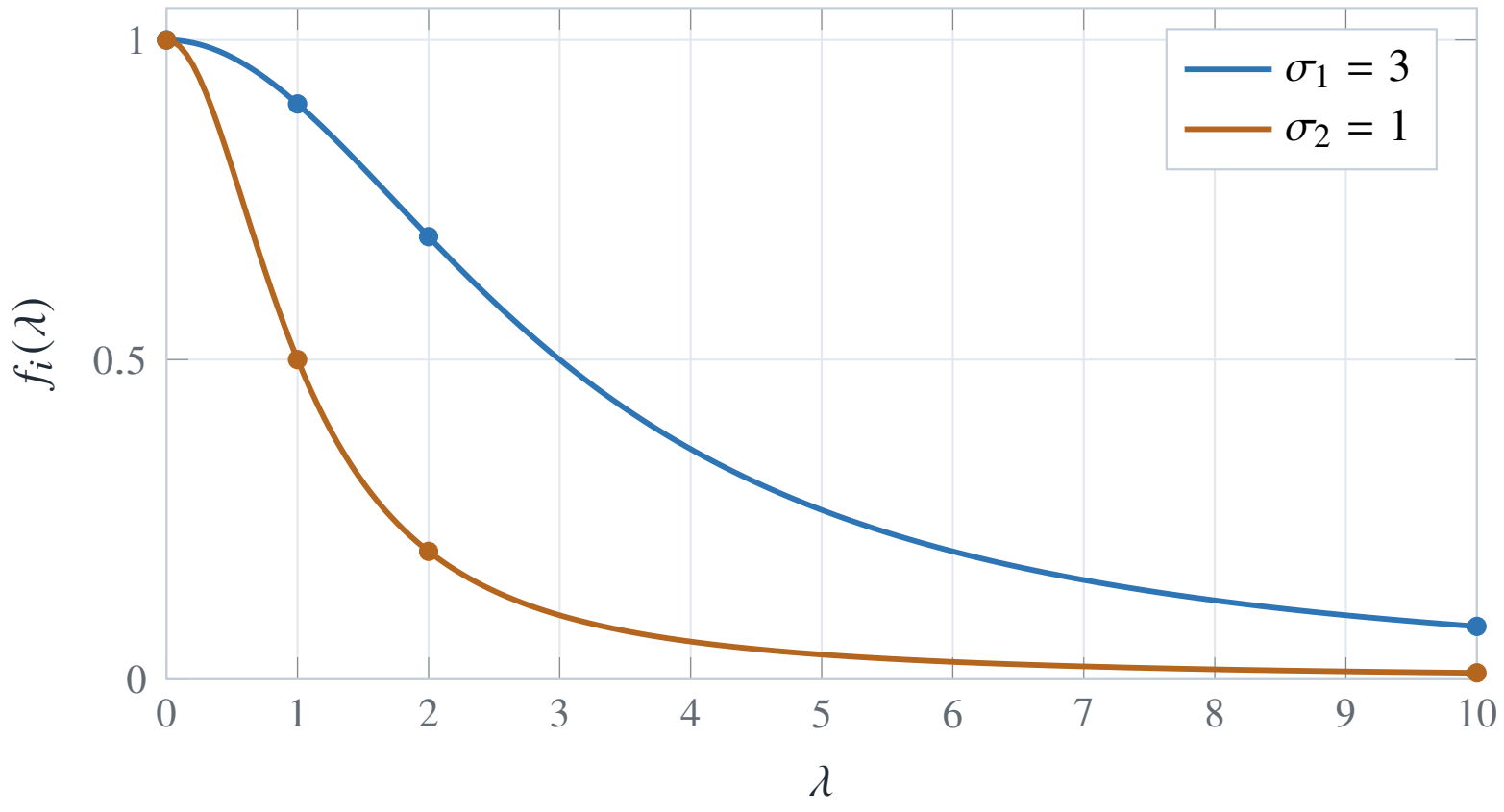
\begin{figure}[htbp]
\centering
\begin{tikzpicture}
\begin{axis}[width=10.6cm,height=6.2cm,xlabel={$\lambda$},ylabel={$f_i(\lambda)$},
  xmin=0,xmax=10,ymin=0,ymax=1.05,grid=both,grid style={rulegray!70},
  axis line style={rulegray!150},
  legend style={font=\footnotesize,draw=rulegray!150},legend pos=north east,
  tick label style={font=\scriptsize,color=struct!70},
  label style={font=\footnotesize,color=struct}]
  \addplot[srcA,line width=1.1pt,domain=0:10,samples=200]{9/(9+x^2)};
  \addlegendentry{$\sigma_1=3$}
  \addplot[srcB,line width=1.1pt,domain=0:10,samples=200]{1/(1+x^2)};
  \addlegendentry{$\sigma_2=1$}
  \addplot[only marks,mark=*,srcA,mark size=1.7pt,forget plot]
     coordinates {(0,1)(1,0.9)(2,0.692308)(10,0.082569)};
  \addplot[only marks,mark=*,srcB,mark size=1.7pt,forget plot]
     coordinates {(0,1)(1,0.5)(2,0.2)(10,0.009901)};
\end{axis}
\end{tikzpicture}
\caption{\textbf{Regularization is a low-pass filter in the generalized basis.} A component with $\sigma_i\gg\lambda$ passes almost unattenuated; one with $\sigma_i\ll\lambda$ is suppressed. The curves show the rates that the four tabulated values only sample: at $\lambda=1$ the weak mode has already lost half its weight while the strong mode retains $90\%$. Read as a precision balance, $f_i$ is the fraction of the directional precision contributed by the data term and $1-f_i$ the fraction contributed by the prior.}
\label{fig:tikhonov-filters}
\end{figure}

Figure~\ref{fig:tikhonov-filters} shows the two filter factors as continuous
functions of $\lambda$, of which the table samples only four points. It also
reads $f_i$ as the share of the directional precision supplied by the data term
rather than by the prior.

\FloatBarrier

\begin{collective}
The filter factor $f_i$ should not be called ``the fraction of the solution that
came from the data''. The solution is an estimator, and such a causal
attribution is not defined by the algebra alone. What \emph{is} exact is the
precision balance in that generalized direction. If the quadratic penalty is
interpreted as a zero-mean Gaussian prior with precision
$\lambda^2\mat{B}\T \mat{B}$, then
\[
   \frac{\alpha_i^2}{\alpha_i^2+\lambda^2\beta_i^2}=f_i
\]
is the fraction of the \emph{total directional precision} contributed by the
data term, while $1-f_i$ is the fraction contributed by the prior precision.
That is the statistically defensible interpretation.
\end{collective}

\section{Special cases, each done fully}

\subsection{Example: \texorpdfstring{$\mat{B}=\mat{I}$}{B = I} recovers the ordinary SVD}
Take $\mat{A}=\begin{psmallmatrix}1&2\\2&1\end{psmallmatrix}$ and $\mat{B}=\mat{I}_2$. From
Section~\ref{sec:svdex}, $\{\sigma_1(\mat{A}),\sigma_2(\mat{A})\}=\{3,1\}$. The GSVD gives
$\vekt{\alpha}=(0.948683,\,0.707107)$, $\vekt{\beta}=(0.316228,\,0.707107)$, so
\[
\frac{\alpha_1}{\beta_1}=\frac{0.948683}{0.316228}=3,\qquad
\frac{\alpha_2}{\beta_2}=1 ,
\]
exactly the ordinary singular values \checkmark\ The ordinary singular values are recovered exactly in this special case.

\subsection{Example: rank-deficient \texorpdfstring{$\mat{B}$}{B}, and \texorpdfstring{$\sigma=\infty$}{sigma = infinity}}
Let
\[
   \mat{A}=\begin{pmatrix}1&0\\0&1\end{pmatrix},\qquad
   \mat{B}=\begin{pmatrix}1&0\\0&0\end{pmatrix}.
\]
Agent $\mathcal B$ is blind to $\vekt{x}_2$. Try a diagonal $\mat{X}=\diag(t_1,t_2)$ with
$\mat{U}=\mat{V}=\mat{I}$. Then $\mat{A}\mat{X}=\diag(t_1,t_2)$ so $\vekt{\alpha}=(t_1,t_2)$, and $\mat{B}\mat{X}=\diag(t_1,0)$
so $\vekt{\beta}=(t_1,0)$. Impose $\alpha_i^2+\beta_i^2=1$:
\[
   i=1:\ t_1^2+t_1^2=1\Rightarrow t_1=\tfrac{1}{\sqrt2}=0.707107;
   \qquad
   i=2:\ t_2^2+0=1\Rightarrow t_2=1 .
\]
So
\[
\mat{X}=\begin{pmatrix}0.707107&0\\0&1\end{pmatrix},\quad
\vekt{\alpha}=(0.707107,\,1),\quad \vekt{\beta}=(0.707107,\,0),\quad
\vekt{\sigma}=(1,\,\infty).
\]
Under the decreasing-$\sigma_i$ convention of Theorem~\ref{thm:gsvd} one would
list the pairs the other way round, $\vekt{\alpha}=(1,\,0.707107)$,
$\vekt{\beta}=(0,\,0.707107)$, $\vekt{\sigma}=(\infty,\,1)$ --- which is what the reference
implementation returns. The ordering is a convention; the pairing
$(\alpha_i,\beta_i,\vekt{x}_i)$ is what matters.
The infinite generalized singular value is not a numerical failure --- it is the
correct, meaningful answer: direction $\vekt{x}_2=\vekt{e}_2$ lies in $\nullsp(\mat{B})$, so the
ratio $\|\mat{A}\vekt{x}\|/\|\mat{B}\vekt{x}\|$ is genuinely unbounded there. Note also the information
shares: along $\vekt{x}_2$, $\alpha_2^2=1$ and $\beta_2^2=0$ --- agent $\mathcal A$
supplies $100\%$.

\begin{collective}
$\sigma_i=\infty$ is the \textbf{single-point-of-failure detector}. Any infinite
generalized singular value in a sensor-suite audit means: exactly one agent
observes that state direction, and losing it makes the direction unobservable.
This is a design finding, available on paper before anything is built.
\end{collective}

\subsection{Example: a common null space, and why \texorpdfstring{$\sigma$}{sigma} is undefined}
Let
\[
\mat{A}=\begin{pmatrix}1&0&0\\0&1&0\end{pmatrix},\qquad
\mat{B}=\begin{pmatrix}1&1&0\end{pmatrix}.
\]
The third coordinate is invisible to both: $\mat{A}\vekt{e}_3=\vekt{0}$ and $\mat{B}\vekt{e}_3=\vekt{0}$. The stack
$\mat{C}$ has rank $r=2<3=n$, so Theorem~\ref{thm:gsvd} does not apply and we use
Theorem~\ref{thm:gsvd-general}: $\mat{X}\in\R^{3\times2}$, and the decomposition simply
does not assign a generalized singular value to $\vekt{e}_3$. Numerically,
$\vekt{\alpha}=(1,\,0.577350)$ and $\vekt{\beta}=(0,\,0.816497)$, i.e.\ $\vekt{\sigma}=(\infty,\,0.707107)$
over the two-dimensional collectively observable subspace.

\emph{Do not patch the $0/0$ direction with a large or small number.} It is not
a big ratio or a small one; it is the absence of any constraint. Silently
assigning it a value is how an ``observability check'' comes to certify an
unobservable system.

\subsection{Example: complementary sensors, and clean information shares}
\label{ex:complementary}
Two agents, two states. Agent $\mathcal A$ is strong on $\vekt{x}_1$ and weaker on
$\vekt{x}_2$; agent $\mathcal B$ is the mirror image:
\[
   \mat{A}=\begin{pmatrix}1&0\\0&0.5\end{pmatrix},\qquad
   \mat{B}=\begin{pmatrix}0.5&0\\0&1\end{pmatrix}.
\]
The pooled matrix is
\[
\mat{J}=\mat{A}\T \mat{A}+\mat{B}\T \mat{B}=\diag(1.25,1.25),\qquad \mat{J}^{-1}=\diag(0.8,0.8).
\]
Choose $\mat{X}=\diag(\sqrt{0.8},\sqrt{0.8})$. Then
\[
\mat{A}\mat{X}=\diag(0.894427,0.447214),\qquad
\mat{B}\mat{X}=\diag(0.447214,0.894427),
\]
so, after ordering by decreasing $\sigma_i$,
\[
(\alpha_1^2,\alpha_2^2)=(0.8,0.2),\qquad
(\beta_1^2,\beta_2^2)=(0.2,0.8),\qquad
\vekt{\sigma}=(2,0.5).
\]

\begin{collective}
Along $\vekt{x}_1$, agent $\mathcal A$ supplies $80\%$ of the pooled directional
information; along $\vekt{x}_2$, only $20\%$. Agent $\mathcal B$ supplies the
complementary shares.

One subtle correction matters for resource allocation. If the standard
deviation of one sensor is halved, its \emph{unnormalised Fisher information}
quadruples. Its \emph{share} does not generally quadruple, because shares must
still sum to one. If the old share is
\[
   s=\frac{a}{a+b},
\]
then after quadrupling $a$ the new share is
\begin{equation}\label{eq:shareupdate}
   s'=\frac{4a}{4a+b}=\frac{4s}{1+3s}.
\end{equation}
For $s=0.8$, this gives $s'=16/17\approx0.941$, not $3.2$. The saturation in
\eqref{eq:shareupdate} is exactly why improving an already dominant sensor has
diminishing returns in relative share.
\end{collective}

\subsection{Example: GSVD mixing angles are not principal angles}
\label{sec:mix-not-principal}

Theorem~\ref{thm:which-angles} identified the $\alpha_i$ as principal-angle
cosines in the \emph{measurement} space. This example shows that they are not
the principal angles between the agents' observable subspaces in \emph{state}
space, in a setting where nothing is degenerate.

Let
\[
   \mat{A}=\begin{pmatrix}1&0&0\\0&1&0\end{pmatrix},
   \qquad
   \mat{B}=\begin{pmatrix}0&1&0\\0&0&1\end{pmatrix}.
\]
Agent $\mathcal A$ observes $(x_1,x_2)$ and agent $\mathcal B$ observes
$(x_2,x_3)$: one shared coordinate, one private coordinate each.

\textbf{Principal angles in state space.} Here
$\range(\mat{A}\T)=\spn\{\vekt{e}_1,\vekt{e}_2\}$ and $\range(\mat{B}\T)=\spn\{\vekt{e}_2,\vekt{e}_3\}$. Both bases are
already orthonormal, so
\[
\mat{Q}_A\T \mat{Q}_B=\begin{pmatrix}\vekt{e}_1\T \vekt{e}_2 & \vekt{e}_1\T \vekt{e}_3\\ \vekt{e}_2\T \vekt{e}_2 & \vekt{e}_2\T \vekt{e}_3\end{pmatrix}
=\begin{pmatrix}0&0\\1&0\end{pmatrix},
\]
whose singular values are $1$ and $0$. Hence
\[
   \theta_1=0^\circ,\qquad \theta_2=90^\circ .
\]

\textbf{GSVD mixing angles.} The pooled matrix is
\[
   \mat{J}=\mat{A}\T \mat{A}+\mat{B}\T \mat{B}=\diag(1,1,0)+\diag(0,1,1)=\diag(1,2,1),
\]
which is invertible, so $r=3$. Since $\mat{J}$ is diagonal, $\mat{X}$ is diagonal, and
$\mat{X}\T \mat{J}\mat{X}=\mat{I}$ forces $\mat{X}=\diag(1,\tfrac{1}{\sqrt2},1)$. Then
\[
\mat{A}\mat{X}=\begin{pmatrix}1&0&0\\[2pt]0&\tfrac{1}{\sqrt2}&0\end{pmatrix},
\qquad
\mat{B}\mat{X}=\begin{pmatrix}0&\tfrac{1}{\sqrt2}&0\\[2pt]0&0&1\end{pmatrix}.
\]
Reading off column norms,
\[
   \vekt{\alpha}=(1,\ 0.707107,\ 0),\qquad \vekt{\beta}=(0,\ 0.707107,\ 1),
\]
so $\alpha_i^2+\beta_i^2=1$ throughout \checkmark, and
\[
   \vekt{\phi}=(0^\circ,\ 45^\circ,\ 90^\circ),
   \qquad
   \vekt{\sigma}=(\infty,\ 1,\ 0).
\]

\textbf{The comparison.}
\[
   \vekt{\phi}=(0^\circ,45^\circ,90^\circ)\ \ \text{has three entries;}
   \qquad
   \vekt{\theta}=(0^\circ,90^\circ)\ \ \text{has two.}
\]
They differ in their values \emph{and} in how many of them there are. No
reordering, padding or change of convention reconciles the two lists. This
settles the matter: \textbf{GSVD mixing angles and principal angles between the
agents' state-space observable subspaces are different objects.}

\begin{collective}
The example is also the cleanest possible picture of a small collective. There
is one $\mathcal A$-exclusive direction ($\sigma=\infty$, i.e.\ $\vekt{x}_1$), one
$\mathcal B$-exclusive direction ($\sigma=0$, i.e.\ $\vekt{x}_3$), and one genuinely
shared direction ($\sigma=1$, i.e.\ $\vekt{x}_2$) on which both agents contribute
equally. By Theorem~\ref{thm:resilience} the resilience factors are $\infty$ for
the two exclusive directions and $\sqrt2$ for the shared one: losing either
agent costs everything along its private coordinate and a factor $1.41$ on the
shared one.
\end{collective}

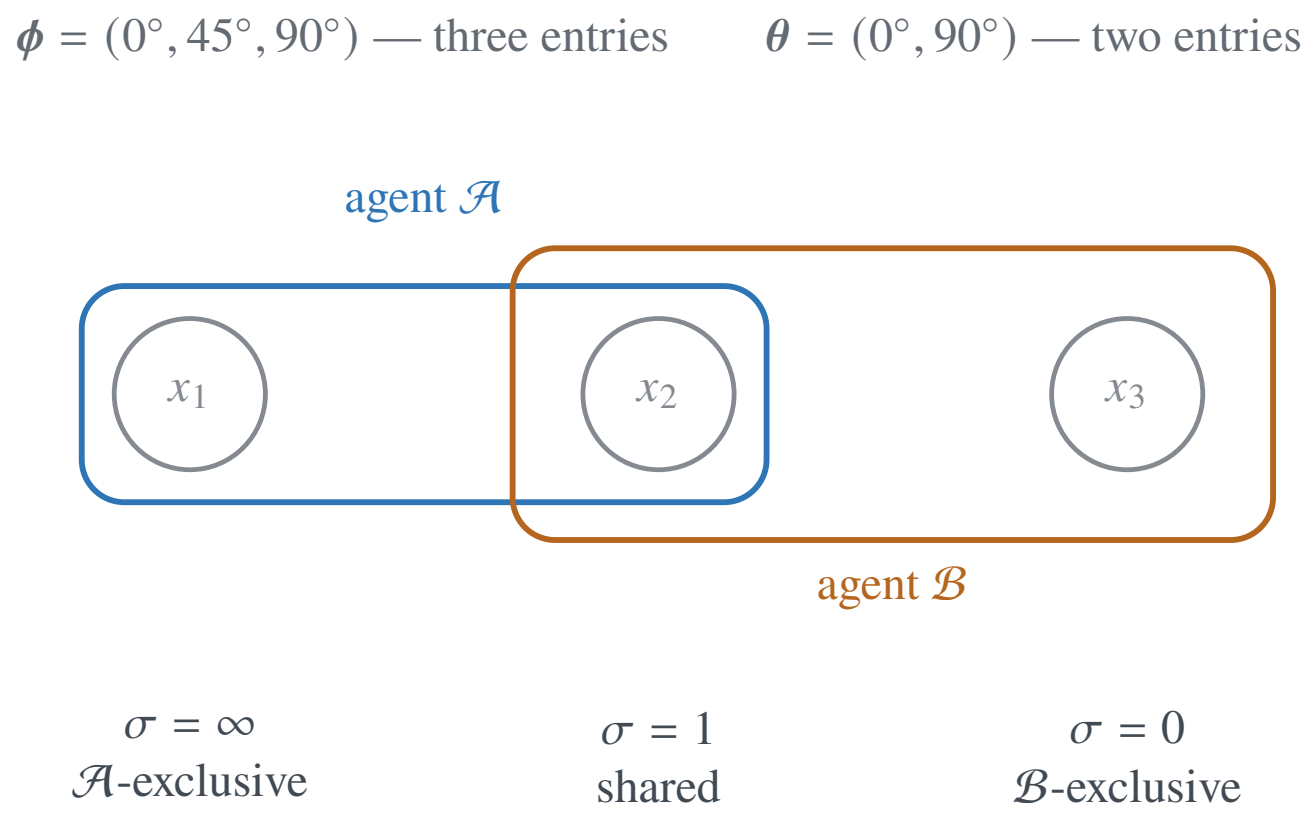
\begin{figure}[htbp]
\centering
\begin{tikzpicture}[figbase,scale=1.0]
  \foreach \i/\x in {1/0,2/3.1,3/6.2}
     \node[draw,struct!55,line width=0.9pt,circle,minimum size=10mm] (X\i) at (\x,0) {$x_\i$};
  \draw[srcA,crv,rounded corners=8pt]
     ($(X1.north west)+(-3.5mm,3.5mm)$) rectangle ($(X2.south east)+(3.5mm,-3.5mm)$);
  \draw[srcB,crv,rounded corners=8pt]
     ($(X2.north west)+(-6mm,6mm)$) rectangle ($(X3.south east)+(6mm,-6mm)$);
  \node[lbl,text=srcA] at (1.55,1.28) {agent $\mathcal A$};
  \node[lbl,text=srcB] at (4.65,-1.28) {agent $\mathcal B$};
  \node[fcap] at (0,-2.4) {$\sigma=\infty$\\ $\mathcal A$-exclusive};
  \node[fcap] at (3.1,-2.4) {$\sigma=1$\\ shared};
  \node[fcap] at (6.2,-2.4) {$\sigma=0$\\ $\mathcal B$-exclusive};
  \node[fcap,text=struct!70] at (3.1,2.35)
     {$\bm\phi=(0^\circ,45^\circ,90^\circ)$ --- three entries\qquad
      $\bm\theta=(0^\circ,90^\circ)$ --- two entries};
\end{tikzpicture}
\caption{\textbf{The complete anatomy of a two-source collective in one picture.} Agent $\mathcal A$ observes $(x_1,x_2)$ and agent $\mathcal B$ observes $(x_2,x_3)$, so one coordinate is shared and each agent holds one privately. All three canonical cases appear at once: $\sigma=\infty$ for the $\mathcal A$-exclusive direction, $\sigma=1$ for the shared one, $\sigma=0$ for the $\mathcal B$-exclusive one. Note also the counts --- three mixing angles against two state-space principal angles. The two families cannot be compared term by term, which is the cleanest available demonstration that they are different objects.}
\label{fig:shared-private-coordinates}
\end{figure}

Figure~\ref{fig:shared-private-coordinates} draws this collective in full,
including the count that settles the question of this example: three mixing
angles against two state-space principal angles.

\FloatBarrier

\clearpage
\part*{Part IV --- What It Means for Collective Systems}

The GSVD is powerful, but it answers one particular question. This part puts it
next to three neighbouring tools so that each is used for the problem it
actually solves:
\begin{center}
\begin{tabularx}{\textwidth}{@{}X l@{}}
\toprule
\textbf{Question} & \textbf{Tool}\\
\midrule
Who contributes how much information? & GSVD\\
How are two subspaces oriented? & Principal angles\\
Are two agents' errors statistically coupled? & Cross-covariance / CCA\\
Which action best separates hypotheses? & KL/Bhattacharyya/Chernoff or task-specific information gain\\
\bottomrule
\end{tabularx}
\end{center}
Confusing these questions produces attractive but false interpretations.

\section{The sensor-suite audit}

For each agent, linearise the measurement model at an operating point and choose
an explicit residual metric. For independent Gaussian sensors, whiten:
\[
   \mat{A}=\mat{\Sigma}_{\mathcal A}^{-1/2}\mat{H}_{\mathcal A},\qquad
   \mat{B}=\mat{\Sigma}_{\mathcal B}^{-1/2}\mat{H}_{\mathcal B}.
\]
Then compute the stack rank, the pooled matrix $\mat{J}=\mat{A}\T \mat{A}+\mat{B}\T \mat{B}$, and the GSVD.
Before interpreting eigenvalues of $\mat{J}$, fix the state metric. If $\vekt{x}=\mat{S}\vekt{\xi}$ maps
dimensionless or otherwise normalised coordinates $\vekt{\xi}$ into the physical state,
then the operators become $\mat{A}\mat{S}$ and $\mat{B}\mat{S}$ and the pooled matrix in those
coordinates is
\[
   \mat{J}_{\xi}=\mat{S}\T \mat{J} \mat{S}.
\]
The generalized singular values are invariant under such a common invertible
reparameterisation, but the ordinary eigenvalues and condition number of $\mat{J}$
are not. A useful audit table therefore contains:

\begin{center}
\begin{tabular}{@{}lll@{}}
\toprule
\textbf{Quantity} & \textbf{Meaning} & \textbf{Design interpretation}\\
\midrule
$n-r$ & common-null dimensions & new sensor or new geometry required\\
$\lambda_{\min}^{+}(\mat{J})$ & weakest nonzero pooled information & absolute weak direction\\
$\kappa^{+}(\mat{J})$ & conditioning on observable subspace & anisotropy / numerical fragility\\
$\sigma_i=\infty$ & only $\mat{A}$ responds & single point of failure\\
$\sigma_i=0$ & only $\mat{B}$ responds & single point of failure\\
$\alpha_i^2,\beta_i^2$ & relative information shares & attribution / resource balance\\
$1/\alpha_i,\,1/\beta_i$ & leave-one-agent-out std inflation & resilience / redundancy budget\\
$g_i$ & separation of neighbouring shares & vector interpretability\\
\bottomrule
\end{tabular}
\end{center}
Here $\lambda_{\min}^{+}$ means the smallest positive eigenvalue and
$\kappa^{+}$ is the condition number restricted to the collectively observable
subspace.

\begin{collective}
A balanced GSVD direction, $\sigma_i\approx1$, is not automatically redundant.
Redundancy requires two things: both agents contribute in the same direction
\emph{and} the pooled direction is sufficiently constrained. Statistical
independence is a third question again. The GSVD gives the relative split;
$\mat{J}$ gives absolute strength; Section~\ref{sec:commonmode} tests cross-agent
linear dependence from paired data.
\end{collective}

\subsection{A worked audit: LiDAR and camera}\label{sec:worked-audit}

The table above is abstract until it is filled in. Here it is, computed by hand
for a realistic pair. Every number was verified numerically.

\textbf{Setup.} A landmark at range $d=10$\,m straight ahead. The state is the
planar position error of that landmark resolved into
$\vekt{x}=(\Delta_\parallel,\Delta_\perp)\T$, along-range and cross-range, both in
metres.

\textbf{Agent $\mathcal A$, LiDAR.} Range noise $\sigma_r=0.02$\,m. Bearing
noise $\sigma_\psi=0.01$\,rad, which at $10$\,m corresponds to a cross-range
error of $d\sigma_\psi=0.1$\,m. Whitening each row by its standard deviation, as
required by \eqref{eq:scaledops},
\[
   \mat{A}=\diag\!\left(\tfrac{1}{0.02},\ \tfrac{1}{0.1}\right)=\diag(50,\ 10)\ \ [\mathrm{m}^{-1}].
\]

\textbf{Agent $\mathcal B$, stereo camera.} Stereo range noise at $10$\,m is
$\sigma_r=0.5$\,m --- disparity-based depth degrades quadratically with range.
Bearing noise $\sigma_\psi=0.001$\,rad gives a cross-range error of $0.01$\,m.
So
\[
   \mat{B}=\diag\!\left(\tfrac{1}{0.5},\ \tfrac{1}{0.01}\right)=\diag(2,\ 100).
\]

\textbf{Step 1: pooled information.}
\[
   \mat{J}=\mat{A}\T \mat{A}+\mat{B}\T \mat{B}=\diag(50^2+2^2,\ 10^2+100^2)=\diag(2504,\ 10100).
\]

\textbf{Step 2: the shared frame.} $\mat{J}$ is diagonal, so $\mat{X}$ is diagonal, and
$\mat{X}\T \mat{J}\mat{X}=\mat{I}$ forces
\[
   \mat{X}=\diag\!\left(\tfrac{1}{\sqrt{2504}},\ \tfrac{1}{\sqrt{10100}}\right)
    =\diag(0.0199840,\ 0.00995037).
\]
The columns of $\mat{X}$ happen to be mutually Euclidean-orthogonal here because
$\mat{X}$ is diagonal and the two agents share eigenvectors. The matrix $\mat{X}$ itself is
\emph{not} orthogonal unless those diagonal entries have unit magnitude.
Obliqueness appears when the information geometries are misaligned; it is not
forced in every example.

\textbf{Step 3: the magnitudes.}
\[
\mat{A}\mat{X}=\diag\!\left(\tfrac{50}{\sqrt{2504}},\ \tfrac{10}{\sqrt{10100}}\right)
  =\diag(0.999201,\ 0.099504),
\]
\[
\mat{B}\mat{X}=\diag\!\left(\tfrac{2}{\sqrt{2504}},\ \tfrac{100}{\sqrt{10100}}\right)
  =\diag(0.039968,\ 0.995037).
\]
Check: $\alpha_1^2+\beta_1^2=\tfrac{2500}{2504}+\tfrac{4}{2504}=1$ \checkmark\ and
$\alpha_2^2+\beta_2^2=\tfrac{100}{10100}+\tfrac{10000}{10100}=1$ \checkmark

\textbf{Step 4: the spectrum.} For a diagonal pair $\sigma_i$ is just the ratio
of the entries,
\[
   \vekt{\sigma}=\left(\tfrac{50}{2},\ \tfrac{10}{100}\right)=(25,\ 0.1).
\]

\textbf{Step 5: read the audit.}

\begin{center}
\begin{tabular}{@{}lrr@{}}
\toprule
& along-range $\Delta_\parallel$ & cross-range $\Delta_\perp$\\
\midrule
$\sigma_i$ & $25$ & $0.1$\\
LiDAR share $\alpha_i^2$ & $0.9984$ & $0.0099$\\
camera share $\beta_i^2$ & $0.0016$ & $0.9901$\\
\midrule
pooled std $1/\sqrt{\mat{J}_{ii}}$ & $2.00$\,cm & $1.00$\,cm\\
LiDAR alone & $2.00$\,cm & $10.0$\,cm\\
camera alone & $50.0$\,cm & $1.00$\,cm\\
\midrule
lose camera, $1/\alpha_i$ & $\times1.0008$ & $\boldsymbol{\times10.05}$\\
lose LiDAR, $1/\beta_i$ & $\boldsymbol{\times25.02}$ & $\times1.005$\\
\bottomrule
\end{tabular}
\end{center}

The boxed factors can be checked against the raw physics without any GSVD:
LiDAR-alone cross-range is $0.1$\,m against a pooled $0.00995$\,m, a ratio of
$10.05$ \checkmark; camera-alone along-range is $0.5$\,m against a pooled
$0.01998$\,m, a ratio of $25.02$ \checkmark\ Theorem~\ref{thm:resilience}
reproduces the physics exactly.

Figure~\ref{fig:lidar-camera-audit}, introduced in Section~\ref{sec:resilience}, draws these same directional standard deviations on one physical scale.

\begin{collective}
Three conclusions follow that the spectrum alone would not have given.

\emph{(i) There is no single point of failure --- and that fact is nearly
irrelevant.} Neither $\sigma_i$ is $0$ or $\infty$, so by
Section~\ref{sec:resilience} the suite stays observable if either agent drops
out. But losing the camera costs a factor $10$ in cross-range. For a tracking or
docking task that is operationally the same as losing the direction.
\textbf{Report $1/\alpha_i$, not just $\sigma_i$.}

\emph{(ii) The complementarity is quantified rather than asserted.} The LiDAR
carries $99.8\%$ of the along-range information and the camera $99.0\%$ of the
cross-range information. This is the textbook range/bearing complementarity, now
as a number.

\emph{(iii) It says where money should go.} Improving the LiDAR's bearing
accuracy raises $\alpha_2^2=0.0099$; by the saturation formula
\eqref{eq:shareupdate}, even a fourfold information gain moves the share only to
about $0.0385$, and the pooled cross-range standard deviation improves from
$0.995$\,cm to $0.980$\,cm --- a $1.5\%$ gain. The same effort spent on the
camera moves the dominant term instead. The audit identifies the saturated knob
\emph{before} anything is procured.
\end{collective}

\subsection{How much the noise model has to be trusted}\label{sec:noise-sens}

Every information claim in this text rests on the whitening matrices of
\eqref{eq:scaledops}, that is, on the assumed noise covariances. It is worth
knowing how hard that assumption is working.

Suppose agent $\mathcal A$'s true noise standard deviation along some direction
is $k$ times the assumed one. Its true information is then $1/k^2$ of the
assumed value, so a claimed share $s$ is really
\begin{equation}\label{eq:noise-sens}
   s'=\frac{s}{s+k^2(1-s)} .
\end{equation}
For a claimed $s=0.9$:

\begin{center}
\begin{tabular}{@{}lcccc@{}}
\toprule
$k$ (true$/$assumed noise) & $1$ & $1.5$ & $2$ & $3$\\
\midrule
true share $s'$ & $0.900$ & $0.800$ & $0.692$ & $0.500$\\
\bottomrule
\end{tabular}
\end{center}

A factor-of-two error in one noise model turns a claimed $90\%$ contribution
into $69\%$; a factor of three turns dominance into a coin flip.

\begin{collective}
\textbf{The attribution is only as good as the covariance, and the sensitivity
is quadratic.} This puts noise-model validation upstream of every information
share reported here, not alongside it. Datasheet error models are frequently
optimistic by exactly the factors in this table, so a published share should
carry the provenance of its noise model and, ideally, a sensitivity band from
\eqref{eq:noise-sens}. An audit computed from datasheet values and an audit
computed from measured values are different scientific claims.
\end{collective}

\section{Active perception: use GSVD as a fast diagnostic, not the objective}

Suppose a current uncertainty direction $\vekt{d}$ lies in the GSVD subspace and write
\[
   \vekt{d}=\mat{X}\vekt{\gamma}.
\]
Because the GSVD diagonalises both quadratic forms,
\begin{equation}\label{eq:directional-info}
 \vekt{d}\T \mat{A}\T \mat{A}\vekt{d}=\sum_i\gamma_i^2\alpha_i^2,
 \qquad
 \vekt{d}\T \mat{B}\T \mat{B}\vekt{d}=\sum_i\gamma_i^2\beta_i^2.
\end{equation}
This gives a cheap answer to a static question: \emph{which current agent
contributes more information along $\vekt{d}$?}

That is useful as a pre-filter for task allocation. It is not, by itself, a
principled active-perception objective. An action changes geometry, visibility,
noise, and often the state distribution. The correct action criterion should be
computed from the \emph{post-action} model: for example expected reduction in
trace, log-determinant, entropy, or a task-specific risk. GSVD can cheaply expose
which directions and agents deserve attention before those more expensive
predictions are evaluated.

\section{Common-mode errors: dependence requires paired statistics}
\label{sec:commonmode}

This is where the earlier principal-angle idea must be replaced by the right
statistical object.

Suppose we have paired residuals or estimation errors
\[
   \vekt{e}_{\mathcal A}^{(k)}\in\R^{d_A},\qquad
   \vekt{e}_{\mathcal B}^{(k)}\in\R^{d_B},\qquad k=1,\dots,n_s,
\]
measured on the same episodes. Estimate the covariance blocks
\[
   \mat{\Sigma}_{AA}=\operatorname{Cov}(\vekt{e}_{\mathcal A}),\qquad
   \mat{\Sigma}_{BB}=\operatorname{Cov}(\vekt{e}_{\mathcal B}),\qquad
   \mat{\Sigma}_{AB}=\operatorname{Cov}(\vekt{e}_{\mathcal A},\vekt{e}_{\mathcal B}).
\]
When the marginal covariances are nonsingular, form the whitened
cross-covariance
\begin{equation}\label{eq:Kcca}
   \mat{K}=\mat{\Sigma}_{AA}^{-1/2}\mat{\Sigma}_{AB}\mat{\Sigma}_{BB}^{-1/2}.
\end{equation}
The singular values
\[
   1\ge\rho_1\ge\rho_2\ge\cdots\ge0
\]
are the canonical correlations. They identify pairs of whitened error directions
with maximal linear correlation. This is canonical correlation analysis (CCA),
going back to Hotelling \cite{Hotelling1936}; its close relation to principal
angles after whitening/orthonormalisation is classical \cite{BjorckGolub1973}.

\paragraph{What the numbers mean.}
\begin{itemize}[leftmargin=1.6em,itemsep=2pt]
\item $\rho_1\approx0$: no detectable \emph{linear} cross-agent correlation in
      the measured residuals.
\item $\rho_1\approx1$: there exists a pair of linear residual directions that
      are almost perfectly correlated.
\item The full spectrum tells whether dependence is concentrated in one mode or
      spread across many.
\end{itemize}
For jointly Gaussian residuals,
\[
   \mat{\Sigma}_{AB}=\mat{0}
   \quad\Longleftrightarrow\quad
   \rho_i=0\ \text{for all }i
   \quad\Longleftrightarrow\quad
   \vekt{e}_{\mathcal A}\ \text{and}\ \vekt{e}_{\mathcal B}\ \text{are independent}.
\]
Outside the Gaussian case, zero canonical correlation rules out linear
correlation, not arbitrary nonlinear dependence.

\begin{collective}
For a research programme on collective failure modes, this separation is
powerful:
\begin{enumerate}[leftmargin=1.6em,itemsep=2pt]
\item GSVD on whitened Jacobians: \emph{where should the two agents contribute?}
\item CCA on paired realised errors: \emph{where do their failures actually
      co-vary?}
\item Compare the two spectra over geometry, weather, scene class, shared maps,
      shared models, and communication regimes.
\end{enumerate}
A geo-referenced multi-agent dataset with good ground truth is exactly the
experimental instrument needed for this comparison. The scientific object is
not a single ``common-mode index'' but a direction-resolved dependence spectrum
with uncertainty estimates.
\end{collective}

\subsection{Why subspace geometry cannot substitute for this}
\label{sec:cca-counterexample}

Remark~\ref{rem:angles-not-independence} showed one failure direction:
\emph{independent} agents whose dominant error subspaces coincide, giving
$\theta_1=0^\circ$ while $\rho_1=0$. The converse failure is at least as
dangerous, and it is the one that matters for a safety argument.

\begin{example}[Orthogonal error subspaces, strongly coupled agents]
\label{ex:cca-converse}
Let the paired residuals $\vekt{e}_{\mathcal A},\vekt{e}_{\mathcal B}\in\R^2$ have
\[
\mat{\Sigma}_{AA}=\begin{pmatrix}4&0\\0&1\end{pmatrix},\qquad
\mat{\Sigma}_{BB}=\begin{pmatrix}1&0\\0&9\end{pmatrix},\qquad
\mat{\Sigma}_{AB}=\begin{pmatrix}1.8&0\\0&0\end{pmatrix}.
\]

\textbf{This is a valid joint covariance.} The only coupled pair is
$(e_{\mathcal A,1},e_{\mathcal B,1})$, with block
$\begin{psmallmatrix}4&1.8\\1.8&1\end{psmallmatrix}$ whose eigenvalues are
\[
   \tfrac{4+1}{2}\pm\sqrt{\left(\tfrac{4-1}{2}\right)^2+1.8^2}
   =2.5\pm\sqrt{2.25+3.24}=2.5\pm2.343075,
\]
that is $4.843075$ and $0.156925$, both positive. The remaining two coordinates
are independent with variances $1$ and $9$. The full spectrum is
$(0.156925,\ 1,\ 4.843075,\ 9)$, so $\mat{\Sigma}\succ0$ \checkmark

\textbf{What principal angles report.} The dominant error direction of
$\mathcal A$ is the top eigenvector of $\mat{\Sigma}_{AA}$, namely $\vekt{e}_1$ (variance
$4$); for $\mathcal B$ it is $\vekt{e}_2$ (variance $9$). The principal angle between
$\spn\{\vekt{e}_1\}$ and $\spn\{\vekt{e}_2\}$ is
\[
   \theta_1=90^\circ,\qquad \cos\theta_1=0 .
\]
A common-mode index built on principal angles reports \textbf{zero coupling}.

\textbf{What canonical correlation reports.} Whitening with
$\mat{\Sigma}_{AA}^{-1/2}=\diag(\tfrac12,1)$ and
$\mat{\Sigma}_{BB}^{-1/2}=\diag(1,\tfrac13)$ gives
\[
\mat{K}=\begin{pmatrix}\tfrac12&0\\0&1\end{pmatrix}
  \begin{pmatrix}1.8&0\\0&0\end{pmatrix}
  \begin{pmatrix}1&0\\0&\tfrac13\end{pmatrix}
 =\begin{pmatrix}0.9&0\\0&0\end{pmatrix},
\qquad
\rho_1=0.9,\quad \rho_2=0 .
\]
The first error components of the two agents are $90\%$ correlated.
\end{example}

Figure~\ref{fig:cca-counterexample}, introduced in Section~\ref{sec:angles}, shows the same construction geometrically: orthogonal dominant marginal directions coexist with a canonical correlation of \(0.9\).

\begin{practicebox}
This counterexample is important for fusion design: marginal error geometry may look complementary while paired errors are strongly correlated. CCA detects the cross-agent coupling that principal angles of marginal error subspaces cannot see.
\end{practicebox}

Taken together with Remark~\ref{rem:angles-not-independence}, principal angles
between marginal error subspaces can report $0^\circ$ for independent agents and
$90^\circ$ for strongly coupled ones. They are \textbf{uninformative about
dependence in both directions}, not merely imprecise. The replacement by
canonical correlations is a necessary correction, not a refinement.

\begin{collective}
The mechanism is easy to state and easy to miss. Principal angles compare the
\emph{marginal} second-order structure of each agent separately, and marginals
carry no information about the joint distribution. All of the coupling lives in
$\mat{\Sigma}_{AB}$, which the marginals never see. Any diagnostic computed from
per-agent residual statistics alone --- however sophisticated --- inherits this
blindness.

The practical consequence is a data-collection requirement with a hard deadline:
estimating $\mat{\Sigma}_{AB}$ needs \textbf{temporally aligned, per-episode paired
residuals against ground truth}, not per-agent error statistics pooled
separately. If a recording campaign is designed without that pairing, $\rho_i$
is simply not computable afterwards, and no post-processing recovers it. This
belongs in the recording protocol, where it is cheap, rather than in the
analysis, where it is impossible.
\end{collective}

In finite samples, especially when $d_A$ or $d_B$ is comparable to $n_s$, sample
covariances are noisy or singular. Then use rank-revealing factorizations,
shrinkage, or regularised CCA and report confidence intervals or bootstrap
stability. A large empirical $\rho_1$ without uncertainty is not yet evidence of
a reproducible common mode.

\section{Cross-view representation alignment and CCA}

CCA also provides a clean tool for comparing learned representations. Let
$\mat{Y}_1\in\R^{n_s\times d_1}$ and $\mat{Y}_2\in\R^{n_s\times d_2}$ contain centred features
for the same $n_s$ scenes. In the full-rank classical setting, whiten each view or,
equivalently, use QR/SVD factors. The canonical correlations are singular
values of the cross-map between the whitened view subspaces.

High canonical correlation means that some \emph{linear variation} is shared
between the two representations. Low canonical correlation means that the two
linear feature spaces share little variation. Neither statement alone implies
that one modality is useless or that the other contains unique task-relevant
information. Task labels and downstream performance are separate evidence.

\begin{collective}
For camera--LiDAR, robot--robot, or edge--cloud representation studies, CCA is a
useful diagnostic before fusion: it tells you what linear structure is already
shared. The GSVD, in contrast, is appropriate when both operators act on the
same underlying state and you want a relative quadratic-form attribution.
Similar algebra; different scientific question.
\end{collective}

\section{Discriminant analysis with singular scatter matrices}

Classical linear discriminant analysis solves a generalized Rayleigh problem of
the form
\[
   \max_{\vekt{w}\neq\vekt{0}}\frac{\vekt{w}\T \mat{S}_b \vekt{w}}{\vekt{w}\T \mat{S}_w \vekt{w}},
\]
where $\mat{S}_b$ and $\mat{S}_w$ are between-class and within-class scatter matrices. When
$\mat{S}_w$ is singular, the ordinary inverse-based derivation breaks down. A GSVD of
suitable factors of the scatter operators handles the algebraic singularity
without first forming $\mat{S}_w^{-1}$. This is the basis of the GSVD discriminant
analysis developed by Howland and Park \cite{HowlandPark2004}.

There is nevertheless a statistical warning. If $\mat{S}_w$ is singular because the
feature dimension is large and the sample size is small, the problem is not
only algebraically singular; it is statistically underdetermined. Infinite or
very large generalized singular values may reflect sampling geometry rather
than stable discrimination. Regularisation and out-of-sample validation are
still required.

\section{Collective decision-making: use a discrimination criterion}
\label{sec:decision}

The ratio
\[
   \frac{\|\mat{A}\vekt{x}\|^2}{\|\mat{B}\vekt{x}\|^2}
\]
measures relative response energy. It does \textbf{not} in general measure how
well two hypotheses can be distinguished.

A two-line counterexample makes this clear. Let
\[
   \mat{A}=\mat{I},\qquad \mat{B}=-\mat{I}.
\]
Then $\mat{A}\T \mat{A}=\mat{B}\T \mat{B}=\mat{I}$, so every generalized singular value is $1$. Yet for every
$\vekt{x}\neq\vekt{0}$ the two predictions $\mat{A}\vekt{x}=\vekt{x}$ and $\mat{B}\vekt{x}=-\vekt{x}$ are different, and under small
measurement noise they are easy to distinguish. The GSVD ratio has discarded
the sign that matters for hypothesis separation.

For two Gaussian hypotheses with the same covariance $\mat{\Sigma}$ and means
$\vekt{\mu}_0,\vekt{\mu}_1$, the natural squared separation is the Mahalanobis distance
\begin{equation}\label{eq:mahal}
   d^2=(\vekt{\mu}_1-\vekt{\mu}_0)\T\mat{\Sigma}^{-1}(\vekt{\mu}_1-\vekt{\mu}_0),
\end{equation}
and
\[
   d_{\mathrm{KL}}(\mathcal N(\vekt{\mu}_1,\mat{\Sigma})\|\mathcal N(\vekt{\mu}_0,\mat{\Sigma}))
   =\frac12d^2.
\]
If $\vekt{\mu}_1-\vekt{\mu}_0=\mat{D}\vekt{x}$ and a design constraint is $\vekt{x}\T \mat{W} \vekt{x}=1$, the best linear
direction solves
\[
   \max_{\vekt{x}\neq\vekt{0}}
   \frac{\vekt{x}\T \mat{D}\T\mat{\Sigma}^{-1}\mat{D} \vekt{x}}{\vekt{x}\T \mat{W} \vekt{x}},
\]
a generalized eigenvalue problem involving the \emph{difference operator} $\mat{D}$,
not a GSVD of the two hypothesis matrices separately.

For unequal covariances, use the full Gaussian KL divergence, Bhattacharyya
distance, Chernoff information, or the actual Bayes risk appropriate to the
decision problem \cite{Kailath1967,CoverThomas2006}.

\begin{collective}
This gives a clean division of labour for active collective perception. Use the
GSVD to understand \emph{which agent supplies information about which state
directions}. Use a divergence or expected decision loss to choose \emph{which
action best separates the hypotheses that matter}. An action can be excellent
for attribution balance and poor for discrimination, or vice versa.
\end{collective}

\section{Related work in these application fields, and what is open}
\label{sec:relatedwork}

A foundations text that motivates methods should say what already exists. The survey below
was carried out by targeted literature search on 7~September 2026. It is a map,
not a systematic review. Representative sources are cited explicitly; any later
novelty claim in a research paper should still be supported by its own documented,
systematic search.

\subsection{What is established}

\textbf{The decomposition itself is mature.} Van Loan's original formulation
\cite{VanLoan1976}, the Paige--Saunders general form \cite{PaigeSaunders1981},
the textbook treatment \cite{GolubVanLoan2013} and the recent geometric account
\cite{EdelmanWang2020} leave little to add mathematically. Its established
application areas are regularization of discrete ill-posed problems
\cite{Hansen1998}, discriminant analysis with singular scatter matrices
\cite{HowlandPark2004}, and comparative genomics \cite{Ponnapalli2011}. None of
these is a sensing application.

\textbf{Observability grading in sensor fusion exists, but is built differently.}
There is an established line of work on ``observable degree analysis'' for
multi-sensor systems, using observability-derived matrices and estimation-error
covariance to grade how well states are determined; a representative
multi-sensor treatment is Hu et al.~\cite{HuEtAl2018}. This answers \emph{how strongly is this state observed},
for one system as a whole. It does not decompose that answer into per-agent
contributions, and it is not built on a pairwise decomposition.

\textbf{The SVD is used heavily in robotics, for other purposes.} Manipulator
kinematics and dexterity analysis, point-set registration and extrinsic
calibration all use the ordinary SVD. These are alignment and conditioning
problems, not attribution problems.

\textbf{Direction-resolved degeneracy analysis is established in state estimation.}
Zhang, Kaess and Singh~\cite{ZhangKaessSingh2016} analyse the local Hessian of an
optimization-based estimator, identify poorly conditioned eigen-directions, and
suppress the update along those directions; their camera--LiDAR demonstration
made this a direct predecessor of a large degeneracy-detection literature.
The distinction relevant here is therefore narrow but important: Hessian
analysis asks \emph{whether} a pooled state direction is weakly constrained,
whereas a pairwise GSVD can ask \emph{which measurement block carries the
constraint} in a shared state direction. A GSVD application must not claim
novelty for direction-resolved degeneracy detection itself.

\textbf{Cooperative and collective perception is a large, fast-moving field.}
Recent surveys of V2X cooperative perception catalogue architectures, fusion
schemes, uncertainty sources and communication constraints
\cite{HuangEtAl2025}. There is also an established literature on
\emph{redundancy mitigation} in collective perception services, where the
redundancy being reduced is primarily transmitted object/message content
\cite{LobatoEtAl2023}. That work is about \emph{communication} redundancy: deciding which
detections need not be transmitted because a neighbour has already reported
them. It is not about information-theoretic redundancy between agents'
estimates.

\textbf{Correlated errors are recognised and are handled conservatively.} A standard instrument is covariance intersection (CI) and its descendants.
CI was introduced specifically for fusion with unknown cross-correlation
\cite{JulierUhlmann1997}; a recent retrospective reviews a quarter century of
work on conservative distributed fusion under uncertain correlation
\cite{ForslingEtAl2024}. These methods fuse estimates whose cross-correlation is
not fully known by constructing a covariance bound that remains conservative
under the stated conditions. This is a genuine and widely used solution, and it
should be acknowledged as such. Its acknowledged cost is conservatism: the
fused covariance is pessimistic, and recent work is explicitly directed at
tightening the bound by exploiting whatever structural knowledge of the
correlation happens to be available.

\textbf{Canonical correlation is already an established multiview tool.}
Classical CCA dates to Hotelling \cite{Hotelling1936}; deep CCA generalises the
idea to nonlinear learned representations \cite{AndrewEtAl2013}. In that
literature CCA is used primarily as a \emph{representation-learning or fusion
mechanism}: learn two views whose coordinates are maximally correlated. The use
proposed here is different: apply CCA to paired estimation residuals as a
\emph{diagnostic} of realised cross-agent error dependence.

\subsection{Where the openings are}

Four gaps follow from the above, in increasing order of ambition.

\paragraph{1. The GSVD as a sensor-suite audit instrument.}
In the targeted literature search performed for this text, I did not
identify a sensing or multi-robot paper that uses the GSVD of two whitened
measurement Jacobians as a sensor-suite audit and reports the full table ---
common-null dimension, the $0$ and $\infty$ cases, information shares, and
spectral gaps. This is an evidence-limited literature observation, not a proof
of novelty; a publication should document its own systematic search. The construction is
elementary and the paper is small, but the instrument does not currently exist
in the sensing literature, and Section~\ref{sec:worked-audit} shows it produces
design conclusions that a raw accuracy comparison does not.

\paragraph{2. Information shares and the resilience factor.}
The identities of Section~\ref{sec:fisher} and Theorem~\ref{thm:resilience} ---
that $\alpha_i^2$ and $\beta_i^2$ are exact relative Fisher-information shares
along $\mat{J}$-normalised directions, and that losing an agent inflates the standard
deviation of the corresponding generalized coordinate by exactly
$1/\alpha_i$ --- were not identified in the targeted sensing and multi-robot
literature reviewed for this text. That is a bounded search result, not a
universal novelty claim. The identities themselves are elementary; their value
is that they convert a spectrum into quantities a design review can act on.

\paragraph{3. Measuring cross-agent error dependence instead of bounding it.}
This is the substantial opening. Covariance intersection exists because the
cross-correlation is not fully known. In a recording campaign with ground truth,
however, cross-agent dependence is at least \emph{estimable}: paired,
temporally aligned residuals from repeated episodes support estimates of the
cross-covariance and the canonical correlations of
Section~\ref{sec:commonmode}, together with finite-sample uncertainty. The CCA
spectrum gives a direction-resolved summary of dependence; it does not make the
cross-covariance known exactly.

Two things make this attractive. First, the field has stated the need: work on
tightening covariance intersection is explicitly seeking structural knowledge of
correlation, and track-to-track fusion papers routinely list conditional
independence as an assumption whose relaxation is left to future work. Second,
the measurement is only possible with an instrument that most groups do not
have --- synchronised multi-agent recordings with reliable ground truth ---
which converts a dataset from a resource into an experimental capability.

The scientific object is the discrepancy between the two spectra: the
\emph{predicted} division of labour from the GSVD of the Jacobians, and the
\emph{realised} co-variation of the errors from CCA on the residuals, compared
across geometry, weather, scene class, shared maps, shared models and
communication regime.

\begin{researchhypothesis}
Under controlled marginal error statistics and matched geometry, the
miscalibration or performance loss of a fusion rule that assumes independence
should increase with the leading canonical correlation of the paired residuals,
and the degradation should concentrate in generalized directions where one
agent carries a large fraction of the nominal information. This is a
falsifiable empirical hypothesis; neither monotonicity nor any proportional law
follows from the GSVD or CCA alone.
\end{researchhypothesis}

\paragraph{4. Feeding the measurement back into the fusion rule.}
A measured canonical-correlation spectrum is a diagnostic, not a complete
fusion model: the numbers $\rho_i$ alone do not determine the full
cross-covariance. The stronger programme is therefore to estimate a structured
cross-covariance (with uncertainty), use CCA to expose its dominant coupled
modes, and then exploit that structure in a fusion method that preserves a
stated consistency guarantee. If this yields a demonstrably less conservative
fusion rule than correlation-agnostic CI, the work closes the loop from
instrument to model to algorithm.

\paragraph{Beyond two agents.} All of the above is pairwise or
one-against-the-rest. The one-against-the-rest framing is exact and requires no
new GSVD theory; I did not identify it as a standard audit construction in the
targeted literature search. Genuine $k$-way structure is a different problem,
for the reasons given in Section~\ref{sec:limitations}.

\begin{collective}
A note on sequencing, since these differ sharply in risk. Items~1 and~2 are
low-risk and largely expository: the mathematics is settled, and the
contribution is instrumentation and clear statement. They are worth doing early
because they establish the vocabulary that items~3 and~4 need. Item~3 is a real
research programme with genuine statistical difficulty --- estimating dependence
from finite residuals, with honest confidence statements, is harder than
computing a decomposition --- and it is the one that requires the recording
infrastructure. Item~4 is the payoff, and it should not be promised before
item~3 has produced measurements worth feeding forward.
\end{collective}

\section{Honest limitations}\label{sec:limitations}

\begin{enumerate}[leftmargin=1.6em,itemsep=5pt]
\item \textbf{Pairwise core.} The classical GSVD is a decomposition of two
matrices. For $k$ agents there are two honest framings. The first is
\emph{one-against-the-rest}: let $\mat{A}$ be agent $i$ and let $\mat{B}$ be the stack of
all the others. This is exact, needs no new theory, and answers the question
that usually matters --- what does agent $i$ add? Theorem~\ref{thm:resilience}
then reads directly as the cost of losing agent $i$, which is precisely the
resilience statement a design review wants. The second is genuinely joint
$k$-way structure, where the two-matrix theory does not extend: three generic
symmetric forms cannot be simultaneously diagonalised by a single congruence.
Higher-order GSVDs exist, but their structure and uniqueness are weaker than in
the two-matrix case and should not be treated as a trivial extension
\cite{Ponnapalli2011}. The boundary deserves to be stated plainly, because it is
also what justifies the harder work: pairwise complementarity has an exact
spectral theory with a clean information interpretation; three-way
complementarity does not, and establishing what replaces it is a research
question rather than a computation.

\item \textbf{Linear and local.} For nonlinear sensors, a Jacobian-based GSVD
is a local statement around the linearisation point. Along a trajectory, repeat
or accumulate the analysis explicitly.

\item \textbf{Metrics are part of the model.} Raw rows with incompatible units
must not be compared by Euclidean norms. State the residual metrics or whiten by
noise covariance before giving information-theoretic meaning to the spectrum.

\item \textbf{Relative is not absolute.} $\sigma_i$ and
$\alpha_i^2/\beta_i^2$ describe relative contributions. Use $\mat{J}$ or an equivalent
absolute information measure to decide whether the collective is strong or weak
in that direction.

\item \textbf{Dependence is separate.} The additive Fisher interpretation
requires independent measurement blocks. Test or model cross-agent dependence;
do not infer it from principal angles of marginal subspaces.

\item \textbf{Numerical rank is a modelling decision near the threshold.}
The common-null dimension depends on a numerical rank tolerance in finite
precision. Report the tolerance or singular-value gap when a rank decision is
scientifically important.

\item \textbf{The educational stacking algorithm has a scale-validity
range.} In the experiments of Section~\ref{sec:scale-precondition}, accuracy
starts to deteriorate for block-norm ratios near $10^{6}$ and can be lost by
about $10^{8}$. Whitening supplies the statistically correct residual metric,
but does not guarantee comparable block norms. If a large ratio remains after
the metric is fixed, treat it as part of the physical problem and prefer a
production GSVD; do not separately rescale the two agents without recognising
that such a rescaling changes the generalized singular values.

\item \textbf{Computation scales with the dense problem size.} Dense GSVD is
appropriate for state dimensions from small to moderate. For large sparse
systems, exploit structure, iterative methods, or reduced subspaces rather than
assuming a dense decomposition is cheap.
\end{enumerate}

\clearpage

\clearpage
\part*{Part V --- The GSVD Through the Cognition Loop}
\label{part:cognitionloop}

\section{From decomposition to cognitive function}
\label{sec:cognition-loop}

The GSVD is most natural when two measurement operators act on a common state.
A cognitive robotic system, however, does not stop at state estimation. It
senses, estimates, represents knowledge, plans, decides and acts. The scientific
question is therefore not whether the GSVD can be inserted mechanically into
every layer, but where its particular structure --- a direction-resolved
comparison of two operators --- adds information that the established methods
of that layer do not already provide.

This chapter follows that loop. For the visual map below, it distinguishes four operational roles.

\begin{enumerate}[leftmargin=1.6em,itemsep=3pt]
\item \textbf{Core operator:} the layer already contains paired linearised
      information operators on a common state. Here the GSVD can be a primary
      analysis tool.
\item \textbf{Diagnostic:} established machinery solves the layer, while the
      GSVD exposes source-resolved directions that deserve attention or can enter
      an action-conditioned objective.
\item \textbf{Interface:} the layer consumes an information description produced
      upstream, but its own representation is not itself a canonical paired linear
      operator on a shared state. The GSVD can annotate or structure that interface,
      not replace the layer's native machinery.
\item \textbf{Constraint:} at the control layer, source-resolved information can
      enter as a condition on admissible behaviour or resilience. It does not supply
      the control law and never substitutes for a stability or safety proof.
\end{enumerate}

\begin{collective}[ --- Reading rule]
A high application impact does not imply that the GSVD itself should dominate
that layer. The strongest research programme is one in which the decomposition
is used where its assumptions are natural and its output is then handed to the
appropriate planning, decision or control machinery.
\end{collective}

\begin{figure}[htbp]
\centering
\begin{tikzpicture}[figbase]
  \def\rad{4.3}
  \node[bx,draw=pooled,fill=pooled!6,minimum width=28mm,minimum height=13mm] (S1) at (-90:\rad)
     {Multimodal\\Estimation\\[1pt]\textcolor{pooled}{\scriptsize core}};
  \node[bx,draw=pooled,fill=pooled!6,minimum width=28mm,minimum height=13mm] (S2) at (-150:\rad)
     {Collective\\Navigation\\[1pt]\textcolor{pooled}{\scriptsize core}};
  \node[bx,draw=struct!45,fill=struct!5,minimum width=28mm,minimum height=13mm] (S3) at (150:\rad)
     {Knowledge\\Representation\\[1pt]\textcolor{struct!75}{\scriptsize interface}};
  \node[bx,draw=srcA,fill=srcA!6,minimum width=28mm,minimum height=13mm] (S4) at (90:\rad)
     {Planning\\[1pt]\textcolor{srcA}{\scriptsize diagnostic}};
  \node[bx,draw=struct!45,fill=struct!5,minimum width=28mm,minimum height=13mm] (S5) at (30:\rad)
     {Decision\\[1pt]\textcolor{struct!75}{\scriptsize interface}};
  \node[bx,draw=srcB,fill=srcB!6,minimum width=28mm,minimum height=13mm] (S6) at (-30:\rad)
     {Control\\[1pt]\textcolor{srcB}{\scriptsize constraint}};
  \foreach \a in {-90,-150,-210,-270,-330,-390}
     \draw[flow] (\a-21:\rad) arc (\a-21:\a-39:\rad);
  \node[text=struct,font=\small\bfseries] at (0,0) {Cognition Loop};
\end{tikzpicture}

\vspace{0.6em}
\begin{tikzpicture}[figbase]
  \foreach \i/\col/\role in {0/pooled/{core operator}, 1/srcA/{diagnostic},
                             2/struct!55/{interface}, 3/srcB/{constraint}}
  { \draw[\col,line width=1.5pt] (\i*3.3,0)--(\i*3.3+0.45,0);
    \node[anchor=west,font=\scriptsize,text=struct!85] at (\i*3.3+0.58,0) {\role}; }
\end{tikzpicture}
\caption{\textbf{The GSVD's legitimate claim contracts around the
cognition loop.} The four roles say how much of a layer the decomposition may
own.\ \ \emph{Core operator}: the layer's central computation \emph{is} a GSVD of
whitened Jacobians on a shared state, because the objects it needs already
exist.\ \ \emph{Diagnostic}: established machinery solves the layer, and the GSVD
supplies a cheap pre-filter --- which state directions and which sources deserve
attention --- before the expensive evaluation is run.\ \ \emph{Interface}: the
layer's own representation is not a metric state vector, so the GSVD can at most
annotate it, attaching source provenance to beliefs it does not itself
compute.\ \ \emph{Constraint}: the GSVD contributes a condition the controller
must respect, never the control law and never the stability proof. Read
clockwise from multimodal estimation, the claim shrinks in exactly that order.
The one-word roles compress the ownership table, which remains the authoritative
statement; the figure shows only the direction of travel.}
\label{fig:cognition-loop}
\end{figure}
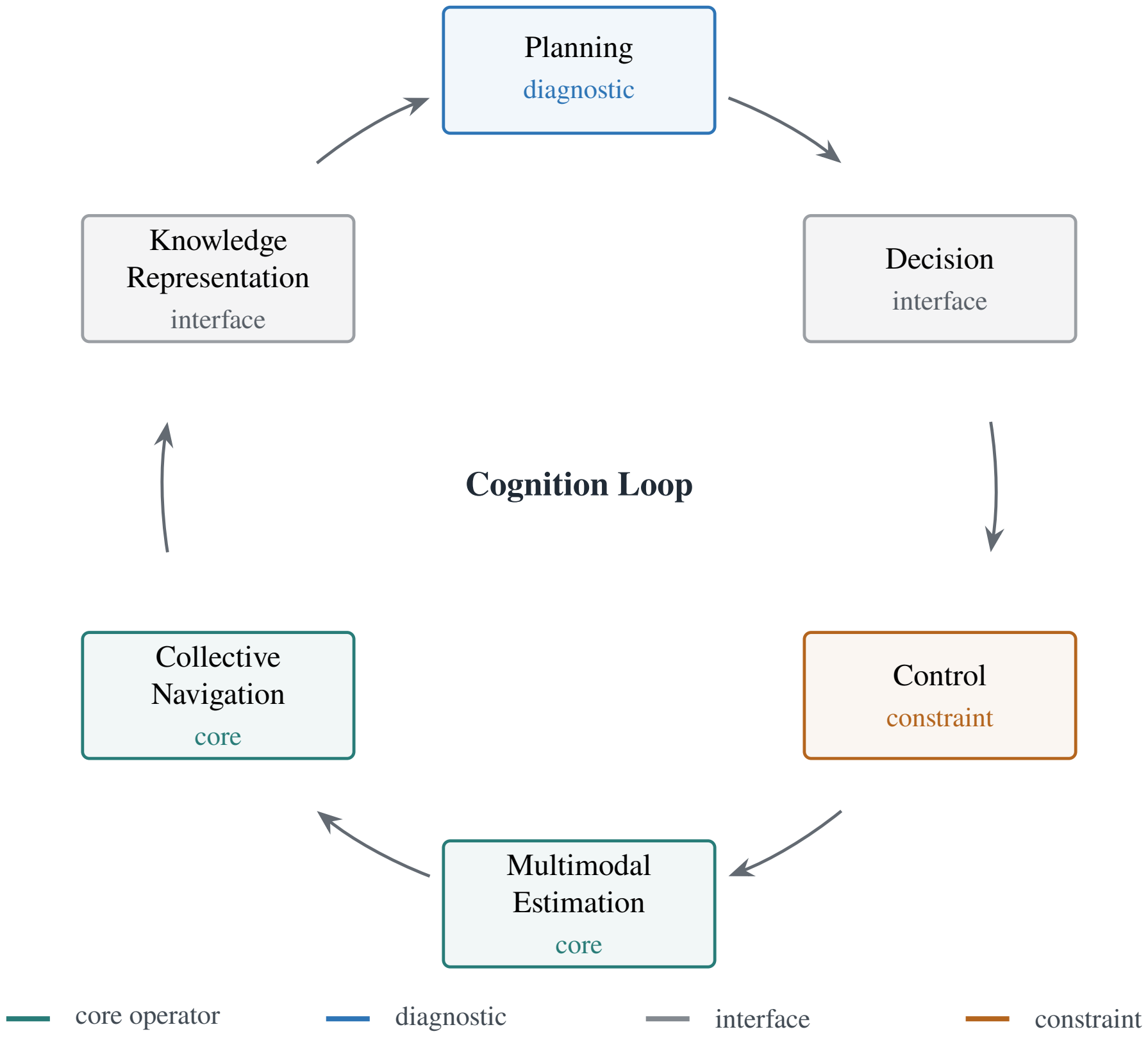

Figure~\ref{fig:cognition-loop} compresses the four roles into a single view of
the loop, grouping the six layers by how much of each the decomposition may own.
The ownership table of Section~\ref{sec:loop-boundary} remains the authoritative
statement.

\FloatBarrier

\subsection{A qualitative map of scientific leverage}

Table~\ref{tab:cognition-loop-map} intentionally uses qualitative categories.
The literature does not support decimal scores such as $9.5$ versus $9.0$, and
the categories are not statistically independent.

\begin{table}[htbp]
\centering
\small
\caption{Role of the GSVD and expected impact along the cognition loop.}
\label{tab:cognition-loop-map}
\begin{tabularx}{\textwidth}{>{\raggedright\arraybackslash}p{2.85cm}>{\raggedright\arraybackslash}p{2.15cm}>{\raggedright\arraybackslash}p{2.3cm}>{\raggedright\arraybackslash}p{2.55cm}>{\raggedright\arraybackslash}X}
\toprule
\textbf{Layer / research field} & \textbf{GSVD role} &
\textbf{Scientific leverage} & \textbf{Application impact} &
\textbf{Main reason}\\
\midrule
Multi-modal INS/GNSS/camera/LiDAR estimation & Core operator & Very high & Very high &
Direct access to whitened Jacobians; can separate pooled degeneracy from source contribution.\\
Collective integrated aided navigation & Core operator / audit & Very high & Very high &
Agent-versus-rest decomposition exposes information responsibility, failure sensitivity and complementarity.\\
Distributed / centralized knowledge representation & Interface / indirect & High, high-risk & High &
Promising for epistemic provenance of metric and probabilistic beliefs; no canonical GSVD exists for arbitrary symbolic facts.\\
Collective / collaborative planning & Action-conditioned diagnostic & Very high & Very high &
Can make information structure a planning objective, but comparison requires a stable reference representation across candidate actions.\\
Collective / collaborative decision making & Interface to belief / utility & Very high, high-risk & Very high &
Can expose who carries decision-relevant evidence; Dec-POMDP and Bayesian decision machinery remain primary.\\
Collective control & Constraint / diagnostic & High--very high & Very high &
Can link information responsibility to formation, resilience and control allocation; stability and safety must still be proved by control theory.\\
\bottomrule
\end{tabularx}
\end{table}

The ordering of the following sections follows the cognitive flow rather than a
publication ranking: local multimodal estimation, collective estimation,
knowledge, planning, decision and control.

\section{Layer 1: multi-modal state estimation}
\label{sec:loop-multimodal}

\subsection{Landscape: multimodal fusion and degeneracy}

Tightly coupled LiDAR--IMU--camera state estimation is now a mature and broad
field. Recent systematic reviews describe nonlinear least squares and factor
graphs, tightly coupled residual models, calibration and robust fusion as the
main computational backbone~\cite{FanEtAl2025}. Degeneracy is likewise not a
new topic: weak state directions are commonly detected from the eigenstructure
of the local Hessian, following the line exemplified by Zhang, Kaess and
Singh~\cite{ZhangKaessSingh2016}. Thus a contribution cannot be ``we analyse
state directions''; the established literature already does so.

\subsection{Where the GSVD adds something}

Let two sensor groups produce whitened local Jacobians
\[
\mat A=\mat\Sigma_A^{-1/2}\mat H_A,
\qquad
\mat B=\mat\Sigma_B^{-1/2}\mat H_B.
\]
The pooled local information is
\[
\FisherMat=\mat A\T\mat A+\mat B\T\mat B.
\]
A Hessian eigendecomposition can identify whether a direction is weak in the
pooled problem. The GSVD adds a different question: for a shared generalized
direction $\vekt x_i$,
\[
\|\mat A\vekt x_i\|^2=\alpha_i^2,
\qquad
\|\mat B\vekt x_i\|^2=\beta_i^2,
\qquad
\alpha_i^2+\beta_i^2=1
\]
under the information-normalised convention of Part~II. Hence the same weak or
strong state direction can be classified by \emph{who carries it}.

\begin{practicebox}
\textbf{Example --- a tunnel traversal.} Suppose the local estimator reports a
poorly conditioned yaw--gyro-bias direction. Pooled Hessian analysis says that
this direction is fragile. A camera-versus-LiDAR/INS GSVD may additionally show
$\alpha_i^2=0.82$ for the camera block and $\beta_i^2=0.18$ for the remaining
sensors. The actionable statement is no longer merely ``yaw is weak'' but
``most of the nominal constraint on this weak mode comes from vision''. A
camera dropout is therefore qualitatively different from losing an equally
accurate sensor that contributes to other modes.
\end{practicebox}

Natural pairings include absolute versus relative sensing,
vision versus LiDAR/inertial sensing, or exteroceptive versus proprioceptive
blocks. For more than two sources, the classical pairwise GSVD remains useful
for hypothesis-driven comparisons such as ``GNSS versus everything else''.
This volume deliberately does not introduce a new multi-source attribution
theory; that is a separate research question rather than a prerequisite.

\subsection{Impact: source-resolved sensor audit and failure diagnosis}

The scientific leverage is \textbf{very high} because the GSVD addresses a
precise gap between two established quantities: pooled conditioning and sensor
identity. The application impact is also \textbf{very high}: sensor-health
management, graceful degradation, sensor-suite design and failure diagnosis all
need to know not only the uncertainty but which modality supports a critical
state direction. The near-term research risk is comparatively low because the
required objects --- residuals, Jacobians and covariances --- already exist in
modern navigation back ends.

\section{Layer 2: collective integrated aided navigation}
\label{sec:loop-collective-nav}

\subsection{Landscape: cooperative localization and distributed fusion}

Cooperative localization combines local navigation with inter-agent relative
measurements and communication. Contemporary reviews cover distributed
filtering, heterogeneous sensor suites, relative ranging and bearing, consensus,
information fusion and the continuing problem of correlation and consistency
in multi-UAV systems~\cite{ZhangEtAl2025Coop}. Robust distributed metric-semantic
SLAM systems such as Kimera-Multi further show that globally useful estimates
can be produced using peer-to-peer communication while handling perceptual
outliers~\cite{TianEtAl2022KimeraMulti}. The open problem is therefore not
whether collective estimation can be computed; it can. The sharper question is
how the resulting observability is distributed over the team.

\subsection{Agent-versus-rest GSVD}

For an agent $a$, collect its whitened measurement block in $\mat A_a$ and all
other available blocks in $\mat A_{\neg a}$. The pair
\[
(\mat A_a,\mat A_{\neg a})
\]
provides a direct source-versus-collective comparison. A generalized direction
with
\[
\alpha_i^2\approx1,\qquad \beta_i^2\approx0
\]
is a warning that the team has concentrated information responsibility in
agent $a$ along that direction. Conversely,
\[
\alpha_i^2\approx\beta_i^2
\]
indicates a more balanced division of nominal information between that agent
and the remainder of the team.

The pair $(\mat A_a,\mat A_{\neg a})$ answers the question for one agent at a
time. The following elementary identity answers it for the whole team at once,
and it requires no simultaneous diagonalisation.

\begin{proposition}[Team shares along a fixed direction]\label{prop:teamshares}
Let $\mat J=\sum_{k=1}^{N}\mat J_k$ with $\mat J_k=\mat A_k\T\mat A_k$ the
whitened information of source $k$, and let $\vekt x$ satisfy
$\vekt x\T\mat J\vekt x=1$. Then
\[
   s_k(\vekt x)=\vekt x\T\mat J_k\vekt x=\|\mat A_k\vekt x\|^2\ \ge 0,
   \qquad
   \sum_{k=1}^{N}s_k(\vekt x)=1 .
\]
\end{proposition}
\begin{proof}
The shares are quadratic forms evaluated at one fixed vector, so
$\sum_k\vekt x\T\mat J_k\vekt x=\vekt x\T\mat J\vekt x=1$; nonnegativity is
$\|\mat A_k\vekt x\|^2\ge0$.
\end{proof}

Proposition~\ref{prop:teamshares} extends the two-source partition
$\alpha_i^2+\beta_i^2=1$ to any number of sources. It also locates the role of
the GSVD precisely: the partition itself is free, and what the decomposition
supplies is the \emph{choice of directions}. For $N=2$ the GSVD produces the
canonical $\mat J$-orthonormal basis in which both sources are simultaneously
diagonal. For $N>2$ no such basis exists in general, so the directions must be
fixed by some other criterion --- an agent-versus-rest GSVD, a task-relevant
direction, or an approximate joint diagonalisation. The share identity holds in
every case.

\begin{practicebox}
Two cautions. The shares are comparable only when the directions are held
fixed; recomputing a basis per configuration reintroduces the difficulty
discussed in Section~\ref{sec:moving-direction}. And $s_k$ is a share of
\emph{nominal} information: it inherits the accuracy of the noise model
$\mat\Sigma_k$ used for whitening, with the quadratic sensitivity quantified in
Section~\ref{sec:noise-sens}.
\end{practicebox}

\begin{practicebox}
\textbf{Example --- two vehicles with GNSS and relative vision.} Vehicle A has
an open-sky GNSS solution; vehicle B is partly shadowed but observes A through a
relative visual constraint. A pooled covariance may remain small. An
A-versus-rest GSVD can nevertheless reveal that global horizontal position is
almost entirely supported by A, while relative baseline and yaw are shared.
The system is accurate, but not equally resilient in all directions.
\end{practicebox}

That distinction is central for collective navigation. Accuracy, resilience
and integrity are not synonyms. A small covariance states that the current
estimate is precise under the assumed model. It does not by itself state how
that precision is distributed over agents or how it will change under a
specific failure.

\subsection{Correlation is the limiting assumption}

If agent errors share maps, models, atmospheric disturbances or previous
messages, the measurement blocks cannot automatically be treated as independent.
Covariance Intersection was developed precisely for fusion when
cross-correlation is unknown~\cite{JulierUhlmann1997,ForslingEtAl2024}. The GSVD
therefore does not replace correlation-aware fusion. A disciplined workflow is:
\[
\begin{array}{c}
\text{model/measure dependence}\\[-1mm]
\downarrow\\[-1mm]
\text{construct a defensible information model}\\[-1mm]
\downarrow\\[-1mm]
\text{apply direction-resolved source analysis}
\end{array}
\]
CCA is useful upstream as a diagnostic on paired residuals, but its canonical
correlations do not by themselves specify the full cross-covariance.

\subsection{Impact: collective observability, resilience, and integrity}

This layer has the strongest long-term combination of \textbf{very high
scientific leverage} and \textbf{very high application impact}. Multi-agent
systems need to answer which platform can fail, which relative observation is
critical, and where additional geometry or communication is worth paying for.
The hard part is empirical rather than algebraic: cross-agent stochastic
structure must be measured under realistic shared disturbances and repeated
geometry. That measurement burden is also what can make the resulting research
position difficult to copy.

\section{Layer 3: distributed and centralized knowledge representation}
\label{sec:loop-knowledge}

\subsection{Landscape: metric-semantic and distributed knowledge}

Modern robotic knowledge representations increasingly connect metric geometry,
objects, places, humans and semantic relations. 3D Dynamic Scene Graphs were
introduced explicitly as actionable spatial representations for planning and
decision making~\cite{RosinolEtAl2020}. Kimera-Multi showed distributed
metric-semantic reconstruction in a multi-robot setting
\cite{TianEtAl2022KimeraMulti}, and MR-COGraphs extends the trend toward
communication-efficient open-vocabulary 3D scene graphs
\cite{GuEtAl2025MRCOGraphs}. The representation problem is therefore moving
from ``where are the landmarks?'' toward structured, queryable and distributed
world knowledge.

\subsection{Where the GSVD is natural --- and where it is not}

A scene-graph proposition such as ``object 17 is a chair'' is not a vector in a
canonical Euclidean state space. Applying the GSVD directly to symbolic facts
would therefore be mathematically artificial. Its natural entry point is the
\emph{metric or probabilistic support} beneath a knowledge item.

Suppose a scene-graph node carries a local Gaussian pose belief
\[
\vekt x_o\sim\mathcal N(\hat{\vekt x}_o,\mat P_o)
\]
and two agents contribute linearised evidence $\mat A$ and $\mat B$ about that
pose. A GSVD can annotate the metric component with direction-resolved
provenance: agent A may determine depth, while agent B mainly determines
lateral position. The semantic node remains a graph entity; the GSVD describes
how its continuous geometric support is distributed.

\begin{practicebox}
\textbf{Example --- shared object knowledge.} Robot A sees a cabinet frontally
with a camera; robot B sees it obliquely with LiDAR. Both agree on the semantic
class. The central map could store one object node, yet the geometry of that
node need not be equally supported in all directions. A local camera-versus-
LiDAR GSVD can attach an epistemic provenance record to the pose or shape
estimate without pretending that the categorical label itself is a generalized
singular vector.
\end{practicebox}

For semantic beliefs, language embeddings or discrete propositions, mutual
information, probabilistic graphical models, provenance logic and PID-like
concepts are generally more natural. The GSVD becomes relevant again only after
a justified linearisation or latent state has been defined.

\subsection{Impact: provenance-aware collective knowledge}

The scientific leverage is \textbf{high but high-risk}. The promising question
is not ``GSVD for knowledge graphs'', but whether collective knowledge should
carry explicit \emph{epistemic provenance}: who supplied a belief, which parts
are independently supported, and which parts rely on shared evidence. The
application impact is \textbf{high} for distributed semantic mapping,
open-vocabulary querying, long-term autonomy and human--robot collaboration,
because communication and trust decisions depend on provenance. The principal
risk is representational: there is no universal state space across metric,
semantic and linguistic knowledge.

\section{Layer 4: collective and collaborative planning}
\label{sec:loop-planning}

\subsection{Landscape: informative planning and active sensing}

Multi-robot planning spans path planning, task allocation, informative path
planning and active sensing. Recent reviews emphasise uncertainty, dynamic
environments, coordination and heterogeneous teams
\cite{Banik2025PathPlanning,KAEtAl2024MRTA}. Information-aware planning is not
new: distributed robotic sensor networks have been controlled by gradients of
mutual information for more than a decade~\cite{JulianEtAl2012}. Consequently,
a GSVD-based planner must offer more than another scalar information-gain
objective.

\subsection{From information amount to information structure}

Let a candidate collective action $\vekt u$ determine expected measurement
Jacobians
\[
\mat A(\vekt u),\qquad \mat B(\vekt u).
\]
The pooled information gain answers how much information is expected. A GSVD
can additionally expose whether the planned geometry makes that information
balanced or concentrated between two sources. For a fixed interpretation
frame, the planner can therefore prefer an action that turns an almost
exclusive direction into a jointly supported direction, even if two candidate
actions have similar $\log\det\FisherMat$.

\begin{practicebox}
\textbf{Example --- cooperative viewpoint selection.} Two robots observe a
pedestrian at an intersection. A move that improves both robots from the same
viewpoint may increase total Fisher information but preserve almost identical
viewing geometry. A second move may produce a slightly smaller total gain while
creating complementary bearing directions. If a critical state component such
as lateral velocity is then supported by both robots rather than one, the
second move can be more valuable for resilience and future decision making.
\end{practicebox}

\subsection{The moving-direction trap}\label{sec:moving-direction}

A crucial methodological constraint appears here. If a fresh GSVD is computed
for every candidate action and its $i$th generalized vector changes with
$\vekt u$, then comparing ``mode $i$'' across actions may compare different
physical directions. Optimization objectives must therefore compare information
in a stable reference representation: for example a fixed state basis over the
planning horizon, a transported subspace, or an explicitly tracked subspace.
This is a general consequence of basis dependence, not a new GSVD theorem.

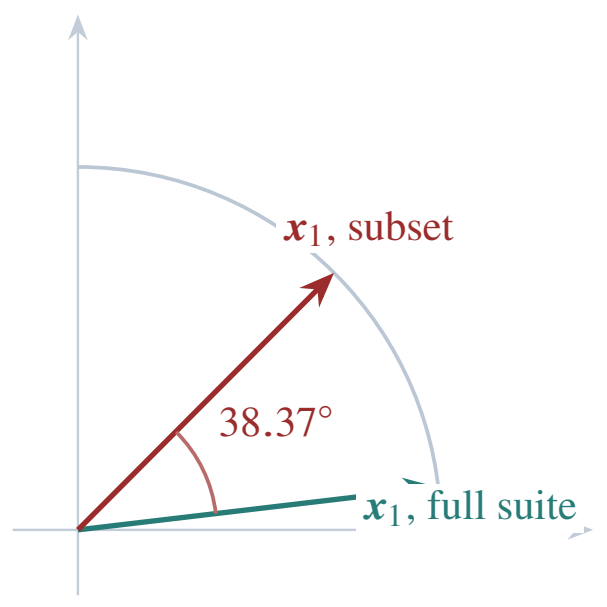
\begin{figure}[htbp]
\centering
\begin{tikzpicture}[figbase,scale=2.4]
  \draw[ax] (-0.18,0)--(1.42,0); \draw[ax] (0,-0.18)--(0,1.42);
  \draw[rulegray!170,thincrv] (1,0) arc (0:90:1);
  \draw[vec,pooled] (0,0)--(0.9931,0.1172);
  \draw[vec,warn] (0,0)--(0.7059,0.7083);
  \node[lbl,text=pooled,font=\scriptsize] at (1.08,0.06) {$\vekt x_1$, full suite};
  \node[lbl,text=warn,font=\scriptsize] at (0.80,0.83) {$\vekt x_1$, subset};
  \draw[warn!70,thincrv] (0.38,0.045) arc (6.73:45.1:0.38);
  \node[lbl,text=warn,font=\scriptsize] at (0.55,0.30) {$38.37^\circ$};
  \node[fcap] at (0.55,-0.52)
     {``mode $1$'' names a different physical direction in each configuration.\\
      Shares are comparable only in a fixed reference frame.};
\end{tikzpicture}
\caption{\textbf{Mode indices are not stable across configurations.} The leading generalized direction of the full sensor suite and that of a subset differ by $38.37^\circ$, although both are called ``mode~$1$''. Comparing their information shares would therefore compare different physical directions. Any objective that ranks configurations, sensor subsets or candidate actions must be evaluated in one fixed reference frame; recomputing a basis per configuration silently changes the thing being compared.}
\label{fig:moving-direction-trap}
\end{figure}

Figure~\ref{fig:moving-direction-trap} illustrates the effect: the leading
generalized direction of a full sensor suite and that of a subset can differ
substantially --- here by $38.37^\circ$ --- although both are indexed as
mode~$1$.

\FloatBarrier

\subsection{Impact: active shaping of collective information}

The scientific leverage is \textbf{very high}: planning is where
source-resolved information can change behaviour rather than merely explain an
estimate. The application impact is also \textbf{very high} for active
perception, formation choice, communication-aware navigation and sensor
deployment. The risk is higher than in estimation because expected information,
visibility, communication and dynamics must be predicted across a horizon.

\section{Layer 5: collective and collaborative decision making}
\label{sec:loop-decision}

\subsection{Landscape: decentralized decisions and task-oriented communication}

Decentralized decision making already has a deep theory. Finite-horizon
Dec-POMDPs are NEXP-hard even for two agents, illustrating why decentralized
partial information is fundamentally harder than a centralized POMDP
\cite{BernsteinEtAl2002}. Decentralized stochastic control therefore relies on
information states, common-information formulations and approximations
\cite{MahajanMannan2016}. In parallel, goal-oriented communication increasingly
asks which messages matter for a shared task rather than merely whether they
can be transmitted faithfully~\cite{CharalambousEtAl2026}.

The GSVD is not an alternative to any of these frameworks. Its possible role is
to make a continuous part of the belief state source-aware before the decision
problem is solved.

\subsection{Decision-relevant source responsibility}

Suppose utility depends locally on a scalar task variable
\[
y=\vekt c\T\vekt x.
\]
If the uncertainty of $y$ is driven by a state subspace that a pairwise GSVD
identifies as almost exclusively supported by agent A, then A has epistemic
importance for that decision even if B has comparable overall state accuracy.
That fact can influence delegation, query selection or communication priority.

\begin{practicebox}
\textbf{Example --- who should resolve a crossing conflict?} Robot A has the
better geometric estimate of free-space clearance; robot B has the better
viewpoint for pedestrian lateral motion. A single global confidence score can
rank one robot above the other and hide this specialization. A
source-resolved belief can instead state which continuous decision variable is
supported by which agent. The decision layer can then request the relevant
message or assign the subdecision to the agent that owns the critical evidence.
\end{practicebox}

For genuinely semantic targets --- intent, class, affordance or a discrete
hypothesis --- mutual information or PID can be a more direct language than
GSVD. The scientifically coherent interface is therefore
\[
\begin{array}{c}
\text{metric directional attribution}\\[-1mm]
\downarrow\\[-1mm]
\text{task-relevant information state}\\[-1mm]
\downarrow\\[-1mm]
\text{Bayesian / Dec-POMDP / optimization decision rule}
\end{array}
\]

\subsection{Impact: epistemic delegation and task-oriented communication}

The scientific leverage is \textbf{very high but high-risk}. A successful
formalism could connect epistemic specialization to role assignment and
communication, but it must show that source awareness improves decisions rather
than merely providing an elegant explanation. The application impact is
\textbf{very high} for heterogeneous robot teams, connected vehicles,
assistive robots and communication-limited autonomy.

\section{Layer 6: collective control}
\label{sec:loop-control}

\subsection{Landscape: distributed and resilient control}

Collective control includes consensus, formation control, containment,
distributed MPC, event-triggered control and resilient coordination. Recent
formation-control reviews emphasise practical constraints such as actuator
limits, communication delay, faults, disturbances and heterogeneous dynamics
\cite{FarooqEtAl2025}. Information-aware multi-robot control is also established:
robots have been moved to maximize mutual information in distributed sensor
networks~\cite{JulianEtAl2012}. The novelty cannot therefore be simply ``use
information in the control loop''.

\subsection{Information structure as a control constraint}

The sharper possibility is to control the \emph{distribution} of observability.
Consider a formation whose tracking error is small, but whose yaw estimate is
supported almost entirely by one platform. A conventional formation controller
can regard the geometry as satisfactory; a source-aware supervisor can regard
it as epistemically fragile and request a formation change before the critical
sensor or communication link is lost.

A useful conceptual separation is
\[
\underbrace{\text{control objective}}_{\text{stability, tracking, safety}}
\qquad\text{and}\qquad
\underbrace{\text{information objective}}_{\text{observability structure}}.
\]
The latter can enter as a soft cost, a constraint, or a supervisory trigger. It
does not replace a Lyapunov proof, MPC feasibility argument or control-barrier
certificate.

\begin{practicebox}
\textbf{Example --- formation under sensor loss.} Three mobile robots maintain a
triangular formation. All pose errors are currently small. A pairwise
agent-versus-rest GSVD shows that one relative-heading direction is almost
exclusively constrained by the front robot's camera. The controller may enlarge
or rotate the formation so that a second robot obtains an independent geometric
baseline. The value of the GSVD is diagnostic: it identifies the direction and
source concentration; the formation controller is still responsible for
executing a stable, collision-free motion.
\end{practicebox}

A deeper control-theoretic bridge is the relation between observability and
control authority. Classical control already studies controllability and
observability Gramians and the directions that are simultaneously easy to
observe and control. In a collective system one may additionally ask whether
the agent that can actuate a critical mode also possesses the information
needed to control it, or whether communication from another agent is essential.
That source-resolved extension is scientifically attractive, but it is a
research programme rather than an established consequence of the GSVD.

\subsection{Impact: information-conditioned collective control}

The scientific leverage is \textbf{high to very high}; the application impact
is \textbf{very high}. Formation resilience, cooperative manipulation,
platooning and humanoid or mobile robot teams can all benefit if information
fragility becomes visible before it produces a control failure. The evidential
bar is correspondingly high: a control paper must demonstrate improved safety,
stability margin, feasibility or closed-loop task performance, not merely a
more interpretable spectrum.

\section{What the GSVD should and should not own}
\label{sec:loop-boundary}

The six layers suggest a useful boundary of responsibility.

\begin{table}[htbp]
\centering
\small
\caption{Method ownership across the cognition loop.}
\label{tab:method-ownership}
\begin{tabularx}{\textwidth}{>{\raggedright\arraybackslash}p{3.0cm}>{\raggedright\arraybackslash}p{4.0cm}>{\raggedright\arraybackslash}X}
\toprule
\textbf{Layer} & \textbf{Primary established machinery} &
\textbf{Legitimate GSVD contribution}\\
\midrule
Multimodal estimation & Factor graphs, Kalman/information filtering, Hessian degeneracy analysis &
Pairwise source attribution along shared state directions.\\
Collective estimation & Distributed estimation, cooperative localization/SLAM, CI and correlation-aware fusion &
Agent-versus-rest information responsibility and failure-oriented audit.\\
Knowledge representation & Scene graphs, semantic maps, probabilistic graphs, embeddings &
Epistemic provenance for continuous or explicitly defined latent subspaces.\\
Planning & Search, optimization, belief-space / informative planning, task allocation &
Action-conditioned structure of information, provided the comparison frame is stable.\\
Decision making & Bayesian decision theory, Dec-POMDPs, team theory, MARL &
Source-aware continuous belief features and communication/delegation diagnostics.\\
Control & Consensus, formation control, MPC, CBFs, resilient/event-triggered control &
Observability-structure constraints or supervisory signals; not the stability proof.\\
\bottomrule
\end{tabularx}
\end{table}

The pattern is deliberate. Near sensing and estimation, the GSVD can sit at the
centre because the mathematical objects it needs already exist. Farther around
the cognition loop, it becomes an information interface whose outputs should be
consumed by methods native to that layer.

\section{Research leverage versus application leverage}
\label{sec:loop-leverage}

Three conclusions follow from the surveyed landscape.

\paragraph{First, multimodal estimation is the fastest falsifiable test.}
It has mature baselines, accessible Jacobians and direct comparisons to Hessian
degeneracy analysis. A useful result must show that source-resolved spectra
explain or predict a failure mode, sensor subset or degradation pattern that
pooled eigenvalues and covariance alone do not.

\paragraph{Second, collective navigation is the strongest long-term empirical
position.} It combines source attribution with the difficult, experimentally
expensive issue of cross-agent dependence. This is where synchronized
multi-agent datasets, controlled ground truth and repeated geometry can become
scientific infrastructure rather than merely evaluation data.

\paragraph{Third, knowledge, planning, decision and control are not optional
``applications at the end''.} They are the layers in which an information
representation can become cognition. Their research risk is higher because the
GSVD must be coupled to semantic representations, prediction, utility and
closed-loop dynamics. Their potential impact is nevertheless large precisely
because they determine communication, action and safety.

\begin{collective}[ --- Cognition-loop interpretation]
The scientifically coherent long-term object is not ``GSVD everywhere''. It is
\emph{source-resolved collective cognition}: retain direction and provenance
where a metric state makes them meaningful; retain dependence and task
relevance when the representation becomes probabilistic or semantic; and let
planning, decision and control act on that enriched information state.
\end{collective}

\section{A final application example across all six layers}
\label{sec:loop-endtoend}

Consider two mobile robots and one instrumented wheelchair moving through a
campus environment. Each platform carries IMU and camera; one robot also carries
LiDAR, and the wheelchair has RTK-capable GNSS outdoors.

\paragraph{1. Multi-modal estimation.}
Within the wheelchair, a GNSS-versus-local-sensors GSVD can show which position,
heading or calibration directions are currently provided by absolute versus
relative sensing.

\paragraph{2. Collective navigation.}
When the team exchanges relative observations, an agent-versus-rest GSVD can
show whether a collective position or heading direction depends excessively on
a single platform.

\paragraph{3. Knowledge representation.}
The shared scene graph can store a doorway, pedestrian and traversable corridor,
while metric entities retain provenance about which platforms support their
geometry. Semantic labels remain semantic objects; they are not forced into the
GSVD.

\paragraph{4. Planning.}
The robots can choose viewpoints that increase information in task-relevant
state directions or reduce concentration on one source, while keeping the
comparison representation fixed over the candidate horizon.

\paragraph{5. Decision making.}
If the wheelchair must decide whether a crossing maneuver is safe, the system
can distinguish the agent that best supports clearance geometry from the agent
that best supports pedestrian motion, and request only the relevant information.

\paragraph{6. Control.}
Finally, the formation or motion controller can react when a safety-critical
state direction becomes epistemically fragile, while all actual stability,
collision-avoidance and actuator constraints remain within the control design.

The same mathematical decomposition has therefore changed role along the loop:
central in estimation, descriptive in knowledge, predictive in planning,
advisory in decision and constraining in control. That is the appropriate scope
for a foundations treatment.

\part*{Part VI --- Numerical Practice}

\section{How it is actually computed}\label{sec:numerics}

The proof in Section~\ref{sec:proof2} suggests a stable educational algorithm:
\begin{enumerate}[leftmargin=1.6em,itemsep=3pt]
\item Factor the stack $\mat{C}=\begin{psmallmatrix}\mat{A}\\\mat{B}\end{psmallmatrix}$ directly
      by a rank-revealing QR or SVD. Do not form $\mat{C}\T \mat{C}$.
\item Split the orthonormal factor into its two row blocks $\mat{P}_1,\mat{P}_2$.
\item Obtain the shared right frame $\mat{W}$ from an SVD of $\mat{P}_1$. Because
      $\mat{P}_1\T \mat{P}_1+\mat{P}_2\T \mat{P}_2=\mat{I}$, the same $\mat{W}$ diagonalises the two block Gram
      forms.
\item Compute \emph{both} $\alpha_i=\|\mat{P}_1\mat{W}\vekt{e}_i\|$ and
      $\beta_i=\|\mat{P}_2\mat{W}\vekt{e}_i\|$ from direct norms. Do not recover a tiny sine from
      $\sqrt{1-\alpha_i^2}$.
\item Construct the minimum-norm shared frame $\mat{X}=\mat{Y}^+$ from the factorisation,
      preferably without an explicit normal-equation pseudoinverse.
\end{enumerate}

This is excellent for understanding and for small diagnostic codes. Production
software uses more elaborate orthogonal transformations and the standard GSVD
block form. LAPACK's current driver is \texttt{xGGSVD3}
\cite{LAPACKGGSVD3,GolubVanLoan2013}.

\subsection{Block-scale imbalance is a numerical issue, not only a modelling issue}
\label{sec:scale-precondition}

Section~\ref{sec:metric} argued for metric-aware operators on \emph{interpretive}
grounds: a metre and a pixel are not comparable, so an unweighted ratio of norms
has no meaning. There is a second, entirely numerical reason for the same
discipline, and it deserves to be stated explicitly.

Step~1 above forms the stack $\mat{C}=\begin{psmallmatrix}\mat{A}\\\mat{B}\end{psmallmatrix}$ and
orthonormalises it. A backward-stable factorisation perturbs the stack by an
amount on the order of $\varepsilon\|\mat{C}\|$. If one block is much smaller than
the other, that absolute perturbation can be comparable to the small block
itself. Heuristically, the relative uncertainty in quantities controlled by the
small block can therefore scale like
$\varepsilon\|\mat{C}\|/\min(\|\mat{A}\|,\|\mat{B}\|)$. Once that ratio approaches one,
roundoff can dominate the information carried by the small block in this
stack-first construction.

Measured behaviour, using $20$ random $4\times4$ pairs per row and comparing
against the independent reference $\sigma_i=s_i(\mat{A}\mat{R}_B^{-1})$:

\begin{center}
\begin{tabular}{@{}lrrl@{}}
\toprule
$\|\mat{A}\|/\|\mat{B}\|$ & median rel.\ error & max rel.\ error & correct digits in $\sigma_i$\\
\midrule
$10^{0}$  & $1.2\times10^{-15}$ & $1.2\times10^{-13}$ & full\\
$10^{2}$  & $2.2\times10^{-15}$ & $4.5\times10^{-14}$ & full\\
$10^{4}$  & $5.8\times10^{-14}$ & $7.8\times10^{-12}$ & full\\
$10^{6}$  & $1.5\times10^{-6}$  & $8.0\times10^{-3}$  & about five\\
$10^{8}$  & $7.6\times10^{-1}$  & $9.8\times10^{-1}$  & \textbf{none}\\
$10^{10}$ & $8.8\times10^{-1}$  & $1.0$               & \textbf{none}\\
\bottomrule
\end{tabular}
\end{center}

So the stacking route is reliable for block-norm ratios up to about $10^{4}$,
degrades through $10^{6}$, and returns numbers with no correct digits beyond
about $10^{8}$. This is a limitation of this transparent stack-first route in finite
precision, not a theorem that every production GSVD must fail at the same
ratios. Robust library algorithms use dedicated reductions and should be
preferred when the metric-correct pair is severely unbalanced.

\begin{collective}
The resolution is more subtle than ``whiten and the scales match.'' Whitening
standardises the \emph{noise}: after the transformation, a unit Euclidean
residual corresponds to one noise-standard-deviation in the chosen model. It
does \textbf{not} standardise the measurement Jacobian. For example,
$\mat{H}_{\mathcal A}=10^9$, $\mat{\Sigma}_{\mathcal A}=1$ and
$\mat{H}_{\mathcal B}=1$, $\mat{\Sigma}_{\mathcal B}=1$ remain
$\mat{A}=10^9$ and $\mat{B}=1$ after whitening. The large ratio then reflects a genuine
signal-to-noise sensitivity difference, not a unit mistake.

The worked audit of Section~\ref{sec:worked-audit} happens to be well scaled
($\|\mat{A}\|/\|\mat{B}\|=0.5$), so the educational routine is accurate there. If a
metric-correct pair remains extremely unbalanced, preserve that physics and use
a production GSVD or a numerically justified equilibration whose effect is
tracked explicitly. Separate rescaling $\mat{A}\mapsto c_A\mat{A}$, $\mat{B}\mapsto c_B\mat{B}$ is
\emph{not} neutral: it changes $\sigma_i$ by $c_A/c_B$ and therefore changes
the relative-information question being asked. A huge block-norm ratio is not,
by itself, a rank statement; structural rank must be established from the
appropriate singular values and tolerances.
\end{collective}

\section{Software --- checked against current documentation}\label{sec:software}

The interfaces below were checked while preparing this revision (7 September
2026):
\begin{center}
\begin{tabularx}{\textwidth}{@{}l l X@{}}
\toprule
\textbf{Environment} & \textbf{Interface} & \textbf{Practical note}\\
\midrule
LAPACK 3.12.1 & \texttt{xGGSVD3} & standard production GSVD driver\\
MATLAB & \texttt{gsvd(A,B)} & returns $\mat{U},\mat{V},\mat{X},\mat{C},\mat{S}$; uses CS machinery\\
Julia & \texttt{svd(A,B)} & returns \texttt{GeneralizedSVD}; LAPACK block form\\
SciPy 1.17.0 & no \texttt{scipy.linalg.gsvd} & \texttt{cossin} exists, but expects a square orthogonal/unitary input\\
\bottomrule
\end{tabularx}
\end{center}

MATLAB documents
$\mat{A}=\mat{U}\mat{C}\mat{X}\T$, $\mat{B}=\mat{V}\mat{S}\mat{X}\T$ with $\mat{C}\T \mat{C}+\mat{S}\T \mat{S}=\mat{I}$ and, since R2022b, sizes the
economy outputs according to the numerical rank of the stack
\cite{MATLABGSVD}. Note that MATLAB's symbol $\mat{X}$ follows a different convention
from the shared frame $\mat{X}$ used in this text: compare the reconstruction
equations before moving factors from one side to the other. Julia exposes the
explicit Paige--Saunders/LAPACK structure
$\mat{A}=\mat{U}\mat{D}_1\mat{R}_0\mat{Q}\T$, $\mat{B}=\mat{V}\mat{D}_2\mat{R}_0\mat{Q}\T$ with $\mat{K}+\mat{L}$ equal to the effective numerical
rank \cite{JuliaGSVD}. SciPy's \texttt{cossin} implements a full CS
decomposition for a square orthogonal/unitary matrix; it is not a direct GSVD
wrapper for an arbitrary pair $(\mat{A},\mat{B})$ \cite{SciPyCossin}.

In the execution environment used for this text, SciPy 1.17.0 has no
\texttt{scipy.linalg.gsvd}, and
\texttt{get\_lapack\_funcs('ggsvd3')} does not expose \texttt{dggsvd3}. This
is an environment/interface fact, not a statement that LAPACK lacks the
routine: LAPACK 3.12.1 documents \texttt{DGGSVD3} explicitly
\cite{LAPACKGGSVD3}.

\section{Conditioning, rank, and what to verify}\label{sec:verify}

\paragraph{1. Rank is numerical.}
A floating-point code needs a threshold to decide
$r=\rank([\mat{A};\mat{B}])$. Near that threshold, ``observable'' versus ``common null'' is
not a theorem about the physical system; it is a numerical/model-selection
decision. Inspect the singular-value gap and report the tolerance when the rank
matters scientifically.

\paragraph{2. Individual directions need separation.}
When two $\alpha_i^2$ values are close, their invariant subspace can be stable
while the individual columns of $\mat{X}$ rotate substantially. Interpret subspaces
when the mixing-angle gap is small \cite{DavisKahan1970,StewartSun1990}.

\paragraph{3. $\mat{X}$ can be Euclidean-ill-conditioned.}
The columns are $\mat{J}$-orthonormal, not necessarily Euclidean-orthogonal. A large
$\kappa(\mat{X})$ means that Euclidean coordinate conversions through $\mat{X}^{-1}$ can
amplify perturbations. Report it if individual coordinates are used downstream.

\paragraph{4. Verify the factorisation without normal equations.}
For the compact form, check
\begin{align*}
   \mat{U}\T \mat{A}\mat{X} &\approx \mat{C}_A,\\
   \mat{V}\T \mat{B}\mat{X} &\approx \mat{C}_B,\\
   \alpha_i^2+\beta_i^2 &\approx1,\\
   (\mat{C}\mat{X})\T(\mat{C}\mat{X})&\approx \mat{I}_r,
   \qquad \mat{C}=\begin{pmatrix}\mat{A}\\\mat{B}\end{pmatrix},\\
   \mat{U}\T \mat{U}&\approx \mat{I},\qquad \mat{V}\T \mat{V}\approx \mat{I}.
\end{align*}
These checks use the original matrices. Do \textbf{not} verify by first forming
$\mat{A}\T \mat{A}+\mat{B}\T \mat{B}$ and then calling a pseudoinverse: that can fail precisely when
the pair is ill-conditioned.

\paragraph{5. Scale the block residuals before believing them.}
The first two checks must be read \emph{relatively}, against the size of the
block they test:
\[
   \frac{\|\mat{U}\T \mat{A}\mat{X}-\mat{C}_A\|_{\max}}{\max_i\alpha_i},
   \qquad
   \frac{\|\mat{V}\T \mat{B}\mat{X}-\mat{C}_B\|_{\max}}{\max_i\beta_i}.
\]
This is not pedantry. On a pair with block-norm ratio $10^{10}$, all four
absolute checks above pass at machine precision ---
$5\times10^{-16}$, $3\times10^{-10}$, $4\times10^{-16}$, $7\times10^{-16}$ ---
while the generalized singular values carry a relative error of $47\%$. The
relevant $\beta_i$ residual of $3\times10^{-10}$ looks tiny only because that
$\beta_i$ itself is $2\times10^{-10}$: in relative terms it is a $100\%$ error. An absolute residual
is meaningless without the scale of the quantity it is compared to, and
Section~\ref{sec:scale-precondition} explains when this regime arises and why
a metric-correct but severely unbalanced pair should be handled with a more robust GSVD rather than trusted blindly.

A simple example shows why. Let
\[
   \mat{A}=\diag(1,10^{-9}),\qquad \mat{B}=\mat{0}.
\]
The exact nonzero pooled spectrum contains $10^{-18}$. A generic pseudoinverse
of $\mat{A}\T \mat{A}$ may legitimately threshold that number away, reporting a false
failure of the identity $\mat{X}\mat{X}\T=\mat{J}^+$. Direct factorisation residuals still remain
near machine precision. Verification should not be less stable than the
algorithm being verified.

\clearpage
\appendix
\section{Reference implementation}\label{app:code}

The reference implementation is educational: it mirrors the compact proof, supports
arbitrary shapes and common-null directions, and deliberately verifies the
result without normal equations. For production use a library implementation of
the standard GSVD.

Two details are not optional. First, both magnitudes are taken from column norms
rather than from $\sqrt{1-\alpha_i^2}$ (Remark~\ref{rem:sines}). Second,
structural zeros are decided from the numerical ranks of $\mat{A}$ and $\mat{B}$, each
relative to its own scale, rather than from a fixed threshold on $\alpha_i$ or
$\beta_i$ (Remark~\ref{rem:structural-zero}). This preserves the distinction
between an exact blind direction and a tiny but finite response.

The code was exercised on all $125$ combinations of
$m,p\in\{1,2,3,5,7\}$, $n\in\{1,2,3,4,6\}$, on $500$ deliberately rank-deficient
stacks, on $2000$ further random cases, and on every worked example in this
text. It also includes a regression test with $\beta_i$ down to approximately
$10^{-15}$ to ensure that tiny finite generalized singular values are not
reported as structural infinities. Worst factorisation residual observed in the
random suite: $3\times10^{-12}$.

\begin{center}
\begin{tcolorbox}[enhanced,colback=codebg,colframe=rulegray,boxrule=0.5pt,
  arc=1.2mm,left=1em,right=1em,top=0.8em,bottom=0.8em,width=0.92\textwidth]
\centering
\ttfamily\large gsvd\_reference.py\\[4pt]
\normalfont\small
archived at Zenodo, DOI~\href{https://doi.org/10.5281/zenodo.23862358}{10.5281/zenodo.23862358}
\end{tcolorbox}
\end{center}

\noindent
The file provides three entry points:

\begin{itemize}[leftmargin=1.5em,itemsep=2pt]
\item \texttt{gsvd(A, B)} --- returns $\mat{U}$, $\mat{V}$, $\mat{X}$, $\vekt{\alpha}$, $\vekt{\beta}$ and the
      paired-diagonal blocks $\mat{C}_A$, $\mat{C}_B$ of Theorem~\ref{thm:gsvd-general}.
\item \texttt{generalized\_singular\_values(alpha, beta)} --- forms
      $\sigma_i=\alpha_i/\beta_i$ with the $\infty$ convention applied only to
      structural zeros, so that a small \emph{finite} response is not silently
      promoted to infinity.
\item \texttt{block\_scale\_ratio(A, B)} and \texttt{check(A, B)} --- the
      verification harness of Section~\ref{sec:verify}, reporting block
      residuals both absolutely and relative to the block they test, and warning
      when the pair leaves the scale-validity range of
      Section~\ref{sec:scale-precondition}.
\end{itemize}

\noindent
Running the file executes the worked examples of this text followed by a
regression suite: all $125$ combinations of $m,p\in\{1,2,3,5,7\}$ and
$n\in\{1,2,3,4,6\}$, $500$ deliberately rank-deficient stacks, $2000$ further
random pairs, and every special case in Part~III. Worst residual observed:
$3\times10^{-12}$. The archive carries the captured output of both suites, so
this figure can be checked against the run it was taken from.

\clearpage
\section{Notation}
\begin{center}
\begin{tabular}{@{}ll@{}}
\toprule
\textbf{Symbol} & \textbf{Meaning}\\
\midrule
$\mat{A}\in\R^{m\times n},\ \mat{B}\in\R^{p\times n}$ & pair of metric-aware operators\\
$\mat{C}=\begin{psmallmatrix}\mat{A}\\\mat{B}\end{psmallmatrix}$ & stacked operator\\
$r=\rank(\mat{C})$ & collectively constrained dimension\\
$\mat{X}$ & shared state-space frame; $n\times r$ in compact form\\
$\alpha_i,\beta_i$ & nonnegative block magnitudes; $\alpha_i^2+\beta_i^2=1$\\
$\sigma_i=\alpha_i/\beta_i$ & generalized singular value\\
$\phi_i=\operatorname{atan2}(\beta_i,\alpha_i)$ & GSVD mixing angle\\
$\theta_i$ & principal angle between two subspaces\\
$\mat{J}=\mat{A}\T \mat{A}+\mat{B}\T \mat{B}$ & pooled Gram / information matrix under whitening\\
$\mat{K}$ & whitened cross-covariance in CCA, Eq.~\eqref{eq:Kcca}\\
$\rho_i$ & canonical correlations, singular values of $\mat{K}$\\
$g_i$ & local separation between contribution eigenvalues\\
\midrule
$\Tf{b}{c}{a}$ & pose of frame $c$ relative to $b$, resolved in $a$ (laboratory convention)\\
$\Cf{b}{a}$ & rotation from frame $b$ to frame $a$\\
\bottomrule
\end{tabular}
\end{center}

\clearpage
\phantomsection
\section*{Code and data availability}
\addcontentsline{toc}{section}{Code and data availability}
\markright{Code and data availability}

The reference implementation described in Appendix~\ref{app:code}, together
with the independent verification script for the tangent-space example of
Section~\ref{sec:lie-worked}, is archived at Zenodo and released under the
BSD 3-Clause licence. Two identifiers are given because they answer different
questions. The residuals quoted in this volume were measured on the version
archived under
DOI~\href{https://doi.org/10.5281/zenodo.23862358}{10.5281/zenodo.23862358},
which is the identifier to use when reproducing them. The concept identifier
DOI~\href{https://doi.org/10.5281/zenodo.23862357}{10.5281/zenodo.23862357}
always resolves to the most recent version, and is the one to follow for later
development. No datasets were used.
\clearpage
\phantomsection

\paragraph{Reference verification.}
The mathematical core, Part~V application references, and the Lie-group
references added in the revision of 10~September 2026 were checked against
primary, publisher, institutional-repository, or official documentation records
on that date. The Lie-group audit specifically confirmed Sol\`a, Deray and
Atchuthan~\cite{SolaDerayAtchuthan2018}; Barrau and Bonnabel's IEEE TAC paper
\cite{BarrauBonnabel2017} and Annual Review~\cite{BarrauBonnabel2018}; the biased-INS
equivariant-filter paper~\cite{FornasierEtAl2022}; Hartley et
al.~\cite{HartleyEtAl2020}; Chirikjian's two volumes~\cite{Chirikjian2009,Chirikjian2012};
the navigation-geometry paper~\cite{BarrauBonnabel2023}; the 2026 left/right IEKF equivalence paper~\cite{GeEtAl2026}; and, as a current preprint rather than settled literature, Zhang et al.~\cite{ZhangEtAl2026NANOL}. These sources
support the distinctions made in Section~\ref{sec:foundation-lie}: local tangent
coordinates are standard robotics machinery; exact log-linear error dynamics
require the appropriate system symmetry; naive inclusion of IMU biases removes
the classical exact group-affine structure; and left/right representations must
be transformed consistently. No bibliography entry has been added since that
audit, which therefore still covers every one of the 47 references listed above. The present revision of 1~November 2026 adds figure cross-references and
the code-availability statement, and changes no reference.

The fourth edition of \emph{Matrix Computations} is published by Johns Hopkins
University Press and has DOI
\href{https://doi.org/10.56021/9781421407944}{10.56021/9781421407944}; the title is
also distributed by SIAM under DOI
\href{https://doi.org/10.1137/1.9781421407944}{10.1137/1.9781421407944}. The
Stewart--Sun ISBN 978-0-12-670230-9 is independently catalogued. The LAPACK page
reports version~3.12.1 and lists \texttt{dggsvd3}; the executable environment
used for the checks in this release reports SciPy~1.17.0. The related-work
statements in Section~\ref{sec:relatedwork} cite representative sources rather
than relying on uncited field-level assertions. Absence-of-literature statements
remain explicitly bounded to the targeted search performed for this text.

For Part~V, the records explicitly confirmed in the audit of 10~September 2026
remain Zhang, Kaess and Singh~\cite{ZhangKaessSingh2016}, Tian et
al.~\cite{TianEtAl2022KimeraMulti}, and Athira K.~A. et
al.~\cite{KAEtAl2024MRTA}. That audit did not independently reopen every
remaining Part~V application record; it was focused on the Lie-group,
geometric-estimation, and tangent-information literature introduced in
Section~\ref{sec:foundation-lie}.
\clearpage
\section{One-page summary}

\begin{enumerate}[leftmargin=1.6em,itemsep=4pt]
\item \textbf{Start with the metric.} Two matrices may share a state space while
      their measurements have different units. For sensor interpretation, use
      metric-aware or whitened operators before comparing response norms.

\item \textbf{The GSVD diagonalises a pair.} In shared directions $\vekt{x}_i$,
      $\mat{A}\vekt{x}_i=\alpha_i \vekt{u}_i$ and $\mat{B}\vekt{x}_i=\beta_i \vekt{v}_i$, with
      $\alpha_i^2+\beta_i^2=1$.

\item \textbf{Relative balance:}
      $\sigma_i=\alpha_i/\beta_i$ and
      $\phi_i=\operatorname{atan2}(\beta_i,\alpha_i)$. Only
      $\sigma_i=\infty$ or $0$ certifies an exclusive single-agent direction.

\item \textbf{Absolute strength:} $\mat{J}=\mat{A}\T \mat{A}+\mat{B}\T \mat{B}$. In the full-rank case,
      $\mat{X}\T \mat{J}\mat{X}=\mat{I}$ and $\mat{X}\mat{X}\T=\mat{J}^{-1}$. With the minimum-norm compact frame,
      $\mat{X}\mat{X}\T=\mat{J}^+$.

\item \textbf{Information attribution:} for independent whitened linear-Gaussian
      sensors, $\alpha_i^2$ and $\beta_i^2$ are exact relative Fisher-information
      shares along $\mat{J}$-normalised generalized directions.

\item \textbf{Do not confuse angles.} GSVD mixing angles $\phi_i$ encode the
      split between the two measurement blocks. They are related to principal
      angles of the stacked measurement subspace to those blocks, but they are
      not the principal angles $\theta_i$ between the agents' observable
      subspaces in state space.

\item \textbf{Do not infer dependence from subspace alignment.} Use paired
      cross-covariance / CCA. For jointly Gaussian residuals, zero cross-covariance
      is equivalent to independence.

\item \textbf{Do not use an energy ratio as a discrimination metric.}
      Hypothesis separation requires the difference between distributions ---
      e.g. Mahalanobis/KL/Bhattacharyya/Chernoff criteria.

\item \textbf{Robotic state geometry comes first.} Rotations and poses live on
      Lie groups, so local Jacobians act on a specified tangent perturbation.
      Common nonsingular tangent-coordinate changes --- including consistent
      left/right $\mathrm{SE}(3)$ perturbations at one operating point --- leave
      generalized singular values and scalar source shares invariant. Comparing
      directions across different operating points additionally requires an
      explicit transport or anchoring convention.

\item \textbf{Numerics: work on $\mat{A}$ and $\mat{B}$, not their normal equations.}
      Rank-revealing QR/SVD and production GSVD routines avoid squaring condition
      numbers. Verify with direct factorisation residuals.

\item \textbf{Interpret subspaces when gaps are small.} Repeated or nearly
      repeated contribution values make individual generalized vectors unstable;
      the invariant subspace is the meaningful object.
\end{enumerate}


\addcontentsline{toc}{section}{References}
\begin{thebibliography}{99}

\bibitem{VanLoan1976}
C.~F. Van Loan,
``Generalizing the Singular Value Decomposition,''
\emph{SIAM Journal on Numerical Analysis}, vol.~13, no.~1, pp.~76--83, 1976.
DOI: \href{https://doi.org/10.1137/0713009}{10.1137/0713009}.

\bibitem{PaigeSaunders1981}
C.~C. Paige and M.~A. Saunders,
``Towards a Generalized Singular Value Decomposition,''
\emph{SIAM Journal on Numerical Analysis}, vol.~18, no.~3, pp.~398--405, 1981.
DOI: \href{https://doi.org/10.1137/0718026}{10.1137/0718026}.

\bibitem{GolubVanLoan2013}
G.~H. Golub and C.~F. Van Loan,
\emph{Matrix Computations}, 4th ed., Johns Hopkins University Press,
Baltimore, 2013. ISBN 978-1-4214-0794-4.
DOI: \href{https://doi.org/10.56021/9781421407944}{10.56021/9781421407944}.

\bibitem{EdelmanWang2020}
A.~Edelman and Y.~Wang,
``The GSVD: Where are the Ellipses?, Matrix Trigonometry, and More,''
\emph{SIAM Journal on Matrix Analysis and Applications}, vol.~41, no.~4,
pp.~1826--1856, 2020.
DOI: \href{https://doi.org/10.1137/18M1234412}{10.1137/18M1234412}.

\bibitem{BjorckGolub1973}
\AA.~Bj\"orck and G.~H. Golub,
``Numerical Methods for Computing Angles Between Linear Subspaces,''
\emph{Mathematics of Computation}, vol.~27, no.~123, pp.~579--594, 1973.
DOI: \href{https://doi.org/10.1090/S0025-5718-1973-0348991-3}{10.1090/S0025-5718-1973-0348991-3}.

\bibitem{Hansen1998}
P.~C. Hansen,
\emph{Rank-Deficient and Discrete Ill-Posed Problems: Numerical Aspects of
Linear Inversion}, SIAM, 1998.
ISBN 978-0-89871-403-6.
DOI: \href{https://doi.org/10.1137/1.9780898719697}{10.1137/1.9780898719697}.
(The author's own book page dates the print edition 1997; SIAM and the
prevailing citation practice use 1998.)

\bibitem{HowlandPark2004}
P.~Howland and H.~Park,
``Generalizing Discriminant Analysis Using the Generalized Singular Value
Decomposition,''
\emph{IEEE Transactions on Pattern Analysis and Machine Intelligence},
vol.~26, no.~8, pp.~995--1006, 2004.
DOI: \href{https://doi.org/10.1109/TPAMI.2004.46}{10.1109/TPAMI.2004.46}.

\bibitem{Hotelling1936}
H.~Hotelling,
``Relations Between Two Sets of Variates,''
\emph{Biometrika}, vol.~28, nos.~3--4, pp.~321--377, 1936.
DOI: \href{https://doi.org/10.1093/biomet/28.3-4.321}{10.1093/biomet/28.3-4.321}.

\bibitem{Ponnapalli2011}
S.~P. Ponnapalli, M.~A. Saunders, C.~F. Van Loan, and O.~Alter,
``A Higher-Order Generalized Singular Value Decomposition for Comparison of
Global mRNA Expression from Multiple Organisms,''
\emph{PLoS ONE}, vol.~6, no.~12, e28072, 2011.
DOI: \href{https://doi.org/10.1371/journal.pone.0028072}{10.1371/journal.pone.0028072}.

\bibitem{DavisKahan1970}
C.~Davis and W.~M. Kahan,
``The Rotation of Eigenvectors by a Perturbation. III,''
\emph{SIAM Journal on Numerical Analysis}, vol.~7, no.~1, pp.~1--46, 1970.
DOI: \href{https://doi.org/10.1137/0707001}{10.1137/0707001}.

\bibitem{StewartSun1990}
G.~W. Stewart and J.-G. Sun,
\emph{Matrix Perturbation Theory}, Academic Press, 1990.
ISBN 978-0-12-670230-9.

\bibitem{Kailath1967}
T.~Kailath,
``The Divergence and Bhattacharyya Distance Measures in Signal Selection,''
\emph{IEEE Transactions on Communication Technology}, vol.~15, no.~1,
pp.~52--60, 1967.
DOI: \href{https://doi.org/10.1109/TCOM.1967.1089532}{10.1109/TCOM.1967.1089532}.

\bibitem{CoverThomas2006}
T.~M. Cover and J.~A. Thomas,
\emph{Elements of Information Theory}, 2nd ed., Wiley, 2006.
Print ISBN 978-0-471-24195-9.
DOI: \href{https://doi.org/10.1002/047174882X}{10.1002/047174882X}.


\bibitem{AndrewEtAl2013}
G.~Andrew, R.~Arora, J.~Bilmes, and K.~Livescu,
``Deep Canonical Correlation Analysis,''
\emph{Proceedings of the 30th International Conference on Machine Learning},
PMLR, vol.~28, no.~3, pp.~1247--1255, 2013.
\url{https://proceedings.mlr.press/v28/andrew13.html}.

\bibitem{HuEtAl2018}
Z.~Hu, T.~Chen, Q.~Ge, and H.~Wang,
``Observable Degree Analysis for Multi-Sensor Fusion System,''
\emph{Sensors}, vol.~18, no.~12, art.~4197, 2018.
DOI: \href{https://doi.org/10.3390/s18124197}{10.3390/s18124197}.

\bibitem{HuangEtAl2025}
T.~Huang, J.~Liu, X.~Zhou, D.~C. Nguyen, M.~R. Azghadi, Y.~Xia,
Q.-L. Han, and S.~Sun,
``Vehicle-to-Everything Cooperative Perception for Autonomous Driving,''
\emph{Proceedings of the IEEE}, vol.~113, no.~5, pp.~443--477, 2025.
DOI: \href{https://doi.org/10.1109/JPROC.2025.3600903}{10.1109/JPROC.2025.3600903}.

\bibitem{LobatoEtAl2023}
W.~Lobato, P.~Mendes, D.~Ros\'ario, E.~Cerqueira, and L.~A. Villas,
``Redundancy Mitigation Mechanism for Collective Perception in Connected and
Autonomous Vehicles,''
\emph{Future Internet}, vol.~15, no.~2, art.~41, 2023.
DOI: \href{https://doi.org/10.3390/fi15020041}{10.3390/fi15020041}.

\bibitem{JulierUhlmann1997}
S.~J. Julier and J.~K. Uhlmann,
``A Non-divergent Estimation Algorithm in the Presence of Unknown Correlations,''
in \emph{Proceedings of the 1997 American Control Conference}, vol.~4,
pp.~2369--2373, 1997.
DOI: \href{https://doi.org/10.1109/ACC.1997.609105}{10.1109/ACC.1997.609105}.

\bibitem{ForslingEtAl2024}
R.~Forsling, B.~Noack, and G.~Hendeby,
``A Quarter Century of Covariance Intersection: Correlations Still Unknown?,''
\emph{IEEE Control Systems Magazine}, vol.~44, no.~2, pp.~81--105, 2024.
DOI: \href{https://doi.org/10.1109/MCS.2024.3358658}{10.1109/MCS.2024.3358658}.



\bibitem{SolaDerayAtchuthan2018}
J.~Sol\`a, J.~Deray, and D.~Atchuthan,
``A micro Lie theory for state estimation in robotics,''
IRI Technical Report IRI-TR-18-01 / arXiv:1812.01537, 2018.
\url{https://arxiv.org/abs/1812.01537}.

\bibitem{Barfoot2024}
T.~D. Barfoot,
\emph{State Estimation for Robotics}, 2nd ed.
Cambridge University Press, 2024.
Chapter~8: ``Matrix Lie Groups.''

\bibitem{BarrauBonnabel2017}
A.~Barrau and S.~Bonnabel,
``The Invariant Extended Kalman Filter as a Stable Observer,''
\emph{IEEE Transactions on Automatic Control}, vol.~62, no.~4,
pp.~1797--1812, 2017.
DOI: \href{https://doi.org/10.1109/TAC.2016.2594085}{10.1109/TAC.2016.2594085}.

\bibitem{BarrauBonnabel2018}
A.~Barrau and S.~Bonnabel,
``Invariant Kalman Filtering,''
\emph{Annual Review of Control, Robotics, and Autonomous Systems}, vol.~1,
pp.~237--257, 2018.
DOI: \href{https://doi.org/10.1146/annurev-control-060117-105010}{10.1146/annurev-control-060117-105010}.

\bibitem{FornasierEtAl2022}
A.~Fornasier, Y.~Ge, P.~van Goor, R.~Mahony, and S.~Weiss,
``Equivariant Filter Design for Inertial Navigation Systems with Input Measurement Biases,''
in \emph{Proc. IEEE International Conference on Robotics and Automation (ICRA)}, 2022.
DOI: \href{https://doi.org/10.1109/ICRA46639.2022.9811778}{10.1109/ICRA46639.2022.9811778}.

\bibitem{HartleyEtAl2020}
R.~Hartley, M.~Ghaffari, R.~M.~Eustice, and J.~W.~Grizzle,
``Contact-aided invariant extended Kalman filtering for robot state estimation,''
\emph{The International Journal of Robotics Research}, vol.~39, no.~4,
pp.~402--430, 2020.
DOI: \href{https://doi.org/10.1177/0278364919894385}{10.1177/0278364919894385}.

\bibitem{Chirikjian2009}
G.~S. Chirikjian,
\emph{Stochastic Models, Information Theory, and Lie Groups, Volume 1: Classical Results and Geometric Methods}.
Birkh\"auser, 2009.
DOI: \href{https://doi.org/10.1007/978-0-8176-4803-9}{10.1007/978-0-8176-4803-9}.

\bibitem{Chirikjian2012}
G.~S. Chirikjian,
\emph{Stochastic Models, Information Theory, and Lie Groups, Volume 2: Analytic Methods and Modern Applications}.
Birkh\"auser, 2012.
DOI: \href{https://doi.org/10.1007/978-0-8176-4944-9}{10.1007/978-0-8176-4944-9}.


\bibitem{BarrauBonnabel2023}
A.~Barrau and S.~Bonnabel,
``The Geometry of Navigation Problems,''
\emph{IEEE Transactions on Automatic Control}, vol.~68, no.~2,
pp.~689--704, 2023.
DOI: \href{https://doi.org/10.1109/TAC.2022.3144328}{10.1109/TAC.2022.3144328}.

\bibitem{ZhangEtAl2026NANOL}
T.~Zhang, W.~Cao, C.~Liu, Y.~Lyu, and S.~E.~Li,
``Natural Gradient Gaussian Approximation Filter on Lie Groups for Robot State Estimation,''
arXiv:2604.10057, 2026. \emph{Preprint}.
DOI: \href{https://doi.org/10.48550/arXiv.2604.10057}{10.48550/arXiv.2604.10057}.

\bibitem{GeEtAl2026}
Y.~Ge, G.~Delama, M.~Scheiber, A.~Fornasier, P.~van Goor, S.~Weiss, and R.~Mahony,
``The difference between the left and right invariant extended Kalman filter,''
\emph{Control Engineering Practice}, vol.~167, art.~106656, 2026.
DOI: \href{https://doi.org/10.1016/j.conengprac.2025.106656}{10.1016/j.conengprac.2025.106656}.

\bibitem{ZhangKaessSingh2016}
J.~Zhang, M.~Kaess, and S.~Singh,
``On Degeneracy of Optimization-based State Estimation Problems,''
in \emph{Proc. IEEE International Conference on Robotics and Automation (ICRA)},
pp.~809--816, 2016.
DOI: \href{https://doi.org/10.1109/ICRA.2016.7487211}{10.1109/ICRA.2016.7487211}.

\bibitem{FanEtAl2025}
Z.~Fan, L.~Zhang, X.~Wang, Y.~Shen, and F.~Deng,
``LiDAR, IMU, and Camera Fusion for Simultaneous Localization and Mapping:
a Systematic Review,''
\emph{Artificial Intelligence Review}, vol.~58, art.~174, 2025.
DOI: \href{https://doi.org/10.1007/s10462-025-11187-w}{10.1007/s10462-025-11187-w}.

\bibitem{ZhangEtAl2025Coop}
Z.~Zhang, N.~Li, G.~Yan, and W.~Li,
``The Development of Distributed Cooperative Localization Algorithms for
Multi-UAV Systems in the Past Decade,''
\emph{Measurement}, vol.~256, part A, art.~118040, 2025.
DOI: \href{https://doi.org/10.1016/j.measurement.2025.118040}{10.1016/j.measurement.2025.118040}.

\bibitem{TianEtAl2022KimeraMulti}
Y.~Tian, Y.~Chang, F.~Herrera~Arias, C.~Nieto-Granda, J.~P.~How, and L.~Carlone,
``Kimera-Multi: Robust, Distributed, Dense Metric-Semantic SLAM for Multi-Robot Systems,''
\emph{IEEE Transactions on Robotics}, vol.~38, no.~4, pp.~2022--2038, 2022.
DOI: \href{https://doi.org/10.1109/TRO.2021.3137751}{10.1109/TRO.2021.3137751}.

\bibitem{RosinolEtAl2020}
A.~Rosinol, A.~Gupta, M.~Abate, J.~Shi, and L.~Carlone,
``3D Dynamic Scene Graphs: Actionable Spatial Perception with Places, Objects, and Humans,''
in \emph{Robotics: Science and Systems (RSS)}, 2020.
DOI: \href{https://doi.org/10.15607/RSS.2020.XVI.079}{10.15607/RSS.2020.XVI.079}.

\bibitem{GuEtAl2025MRCOGraphs}
Q.~Gu, Z.~Ye, J.~Yu, J.~Tang, T.~Yi, Y.~Dong, J.~Wang, J.~Cui, X.~Chen, and Y.~Wang,
``MR-COGraphs: Communication-Efficient Multi-Robot Open-Vocabulary Mapping System via 3D Scene Graphs,''
\emph{IEEE Robotics and Automation Letters}, vol.~10, no.~6, pp.~5713--5720, 2025.
DOI: \href{https://doi.org/10.1109/LRA.2025.3561569}{10.1109/LRA.2025.3561569}.

\bibitem{Banik2025PathPlanning}
S.~Banik, S.~C. Banik, and S.~S. Mahmud,
``Path Planning Approaches in Multi-Robot System: A Review,''
\emph{Engineering Reports}, vol.~7, no.~1, art.~e13035, 2025.
DOI: \href{https://doi.org/10.1002/eng2.13035}{10.1002/eng2.13035}.

\bibitem{KAEtAl2024MRTA}
A.~K.~A., J.~Divya Udayan, and U.~Subramaniam,
``A Systematic Literature Review on Multi-Robot Task Allocation,''
\emph{ACM Computing Surveys}, vol.~57, no.~3, pp.~1--28, 2024.
DOI: \href{https://doi.org/10.1145/3700591}{10.1145/3700591}.

\bibitem{JulianEtAl2012}
B.~J. Julian, M.~Angermann, M.~Schwager, and D.~Rus,
``Distributed Robotic Sensor Networks: An Information-Theoretic Approach,''
\emph{The International Journal of Robotics Research}, vol.~31, no.~10, 2012.
DOI: \href{https://doi.org/10.1177/0278364912452675}{10.1177/0278364912452675}.

\bibitem{BernsteinEtAl2002}
D.~S. Bernstein, R.~Givan, N.~Immerman, and S.~Zilberstein,
``The Complexity of Decentralized Control of Markov Decision Processes,''
\emph{Mathematics of Operations Research}, vol.~27, no.~4, pp.~819--840, 2002.
DOI: \href{https://doi.org/10.1287/moor.27.4.819.297}{10.1287/moor.27.4.819.297}.

\bibitem{MahajanMannan2016}
A.~Mahajan and M.~Mannan,
``Decentralized Stochastic Control,''
\emph{Annals of Operations Research}, vol.~241, no.~1, pp.~109--126, 2016.
DOI: \href{https://doi.org/10.1007/s10479-014-1652-0}{10.1007/s10479-014-1652-0}.

\bibitem{CharalambousEtAl2026}
T.~Charalambous, N.~Pappas, N.~Nomikos, and R.~Wichman,
``Toward Goal-Oriented Communication in Multi-Agent Systems: An Overview,''
\emph{IEEE Open Journal of the Communications Society}, vol.~7,
pp.~6576--6611, 2026.
DOI: \href{https://doi.org/10.1109/OJCOMS.2026.3703748}{10.1109/OJCOMS.2026.3703748}.

\bibitem{FarooqEtAl2025}
A.~Farooq, Z.~Xiang, W.-J.~Chang, and M.~S. Aslam,
``Recent Advancement in Formation Control of Multi-Agent Systems: A Review,''
\emph{Computers, Materials \& Continua}, vol.~83, no.~3, pp.~3623--3674, 2025.
DOI: \href{https://doi.org/10.32604/cmc.2025.063665}{10.32604/cmc.2025.063665}.

\bibitem{LAPACKGGSVD3}
LAPACK,
``\texttt{xGGSVD3}: Generalized SVD Driver,'' LAPACK 3.12.1 documentation.
\url{https://netlib.org/lapack/explore-html/d1/d27/group__ggsvd3.html}
(accessed 7 September 2026).

\bibitem{MATLABGSVD}
MathWorks,
``\texttt{gsvd} --- Generalized singular value decomposition,'' MATLAB documentation.
\url{https://www.mathworks.com/help/matlab/ref/gsvd.html}
(accessed 7 September 2026).

\bibitem{JuliaGSVD}
Julia Language,
``\texttt{LinearAlgebra.GeneralizedSVD},'' Julia standard-library documentation.
\url{https://docs.julialang.org/en/v1/stdlib/LinearAlgebra/}
(accessed 7 September 2026).

\bibitem{SciPyCossin}
SciPy,
``\texttt{scipy.linalg.cossin},'' SciPy 1.17.0 manual; the
interpreter used for the executable checks in this release reports
\texttt{scipy 1.17.0}.
\url{https://docs.scipy.org/doc/scipy/reference/generated/scipy.linalg.cossin.html}
(accessed 7 September 2026).

\end{thebibliography}
\end{document}